\documentclass[11pt,letterpaper]{article}

\usepackage[T1]{fontenc}
\usepackage[utf8]{inputenc}
\usepackage{lmodern}
\usepackage{microtype}

\usepackage[margin=1in]{geometry}

\usepackage{amsmath,amssymb}

\usepackage{graphicx}
\usepackage{float}
\usepackage{subcaption}
\usepackage[figuresright]{rotating}
\usepackage{booktabs}
\usepackage{multirow}
\usepackage{array}
\usepackage{makecell}
\usepackage{threeparttable}
\usepackage{adjustbox}
\usepackage{overpic}
\usepackage{pdflscape}
\usepackage{placeins}

\usepackage{tikz}
\usepackage{pgfplots}
\pgfplotsset{compat=1.18}

\usepackage{enumitem}
\usepackage{xcolor}
\usepackage{soul}

\usepackage{changepage}
\newlength{\extralength}

\usepackage[colorlinks=true,linkcolor=blue,citecolor=blue,urlcolor=blue]{hyperref}
\usepackage{xurl}

\newcommand{\mdpibackmatter}[2]{\par\medskip\noindent\textbf{#1:} #2\par}
\newcommand{\supplementary}[1]{\mdpibackmatter{Supplementary Materials}{#1}}
\newcommand{\authorcontributions}[1]{\mdpibackmatter{Author Contributions}{#1}}
\newcommand{\funding}[1]{\mdpibackmatter{Funding}{#1}}

\newcommand{\dataavailability}[1]{\mdpibackmatter{Data Availability Statement}{#1}}
\newcommand{\acknowledgments}[1]{\mdpibackmatter{Acknowledgments}{#1}}
\newcommand{\conflictsofinterest}[1]{\mdpibackmatter{Conflicts of Interest}{#1}}

\newcommand{\PublishersNote}[1]{}  

\title{From Local Training to Large-Scale Mapping: A Comparative Assessment of Machine Learning and Deep Learning for Transferable Satellite-Derived Bathymetry}
\author{%
  Hsiao-Jou Hsu\thanks{\texttt{hsu.771@osu.edu}} \and
  Joachim Moortgat\thanks{Corresponding author: \texttt{moortgat.1@osu.edu}}\\[2pt]
  School of Earth Sciences, The Ohio State University,\\
  275 Mendenhall Laboratory, 125 South Oval Mall, Columbus, OH 43210, USA
}
\date{}

\begin{document}
\maketitle

\begin{abstract}
Satellite-derived bathymetry (SDB) provides a cost-effective means for mapping shallow-water depths, yet its scalability and cross-regional generalizability remain challenging in optically complex coastal environments. This study systematically evaluates machine learning (ML) and deep learning (DL) approaches for transferable SDB over the 0--20~m depth range using multispectral Sentinel-2 imagery. A Random Forest model and four deep learning architectures--ResNet-50, ResNet-101, EfficientNet-B4, and ConvNeXt-Large--are developed and trained using data from Pratas Island (South China Sea) and selected reef regions of the Great Barrier Reef (GBR), and subsequently evaluated on spatially independent intra-regional and cross-regional test areas to assess generalization performance.
Model sensitivity is investigated with respect to key training configurations, including loss-function design and data-splitting strategy. To enhance shallow-water learning, we introduce a Smooth Weight Function (SWF)-weighted RMSE loss that emphasizes near-surface depths and compare it with conventional RMSE and relative percentage error (RPE) objectives. In terms of training data, preserving spatial continuity during training substantially improves both numerical accuracy and structural consistency of predictions compared with random patch splitting. While the Random Forest model performs competitively in intra-regional tests, its accuracy degrades under cross-regional transfer (RMSE increasing from 1.53~m to 2.99--3.78~m). Deep learning models, although not always outperforming Random Forest in intra-regional settings, exhibit greater robustness to geographic shift. Using the spatially continuous training strategy, intra-regional RMSE ranges from 1.15 to 1.92~m over the full 0--20~m range, with shallow-water RMSE as low as 0.26~m for depths $\leq$3~m. Cross-regional transfer to geographically independent reefs yields moderate RMSE values of approximately 2.46--2.98~m (0--20~m range), indicating that geographic transfer remains challenging despite meaningful improvements over Random Forest.


We further benchmark the proposed architectures against a task-specific bathymetry network using the public MagicBathyNet dataset. Under a unified 0--16~m shallow-water configuration using aerial RGB imagery, the proposed models achieve RMSE values between 0.19 and 0.22~m, outperforming both the baseline U-Net and the transformer-based bathymetry architecture while using substantially fewer parameters.


In addition, we exploit multi-temporal repeat imagery for both training and inference, which increases training diversity and improves robustness to temporal variability arising from changing sun angles, atmospheric conditions, water properties, and tides. During inference, predictions from multiple repeat images are aggregated using the median to reduce noise and improve stability. Finally, we release optimized network architectures and pretrained weights to facilitate scalable application to new sites. This work demonstrates a practical pathway toward transferable, large-area SDB from multispectral satellite imagery using deep learning.
\end{abstract}

\noindent\textbf{Keywords:} satellite-derived bathymetry; machine learning; deep learning; model transferability; loss function weighting; shallow-water depth retrieval; large-scale mapping

\vspace{6pt}
\noindent 
\textbf{What are the main findings?}
 \begin{itemize}[labelsep=2.5mm,topsep=-3pt]

\item Deep learning models (ResNet and ConvNeXt variants) substantially outperform Random Forest in cross-regional satellite-derived bathymetry, with ConvNeXt-Large achieving cross-regional RMSE approaching $\sim$1\,m at intermediate depths (5--11\,m) at Cartier Reef and shallow depths (1--4\,m) at Ashmore Reef.
 \item Preserving spatial continuity in training data---keeping patches as contiguous reef blocks rather than random samples---is the most impactful training design choice, consistently improving both intra-regional accuracy (shallow-water RMSE as low as 0.26\,m for depths $\leq$3~m ) and cross-regional transferability.
 \end{itemize} \vspace{3pt}
\textbf{What are the implications of the main findings?}
 \begin{itemize}[labelsep=2.5mm,topsep=-3pt]
 \item A depth-weighted loss function (Smooth Weight Function), combined with multi-temporal median aggregation of repeat Sentinel-2 imagery, provides a practical and deployable strategy for accurate and stable shallow-water bathymetry retrieval at new, unmapped sites without additional field surveys.
 \item General-purpose convolutional backbones with fewer than 60\,M parameters match or surpass task-specific transformer architectures (e.g., Swin-BathyUNet with 395\,M parameters), with pretrained weights released openly to enable scalable, transferable mapping of shallow coastal and reef environments.
 \end{itemize}

\section{Introduction}
\label{sec:introduction}

Shallow-water bathymetry is essential for navigation, coastal engineering, ecosystem monitoring, and hazard mitigation, yet it remains among the least well-resolved parts of the global ocean. Conventional surveys using multibeam echosounders and airborne topo-bathymetric LiDAR provide high accuracy but are costly and logistically demanding, resulting in limited spatial coverage.

As of the 2025 Seabed~2030 update, only 27.3\% of the global seafloor has been mapped with modern techniques \cite{Seabed2030_2025}, and existing global products offer resolutions of 100--800~m that are inadequate for resolving reefs, lagoons, and navigation hazards \cite{Mayer2018}. Although shallow areas represent only about 7\% of the seafloor, they demand roughly 64\% of total survey effort \cite{Mayer2018}, motivating scalable approaches such as satellite-derived bathymetry.

Satellite-derived bathymetry (SDB) exploits the penetration of visible wavelengths into the water column. Classical approaches include linear regression in log-transformed reflectance \cite{Lyzenga1978, Lyzenga1985, Lyzenga2006} and log band-ratio methods \cite{Stumpf2003}, with numerous extensions proposed to improve robustness under varying optical conditions \cite{Chen2019, Mabula2023, Ye2024, Figliomeni2024}. Sentinel-2 has emerged as a particularly suitable sensor for shallow-water reef applications, providing the spatial resolution, spectral coverage, and revisit frequency required for operational bathymetry and benthic mapping over coral reef systems \cite{Hedley2018}. Despite these refinements, most traditional SDB models remain strongly site-dependent and reliably retrieve depths only within approximately 10~m, limiting their transferability.

To address the nonlinear relationship between spectral reflectance and water depth, machine learning (ML) methods have been increasingly adopted. Algorithms such as random forest, gradient boosting, support vector machines, and XGBoost have demonstrated improved accuracy over classical regression models \cite{Mabula2023, Wu2022}, with random forest often exhibiting superior robustness on heterogeneous datasets \cite{Kwon2024,Tonion2020, Sagawa2019}. Nevertheless, ML-based SDB remains sensitive to training data characteristics and environmental variability such as water turbidity, seabed composition, and atmospheric conditions \cite{Peng2022, Najar2022, Xie2024}, with contrasting findings suggesting persistent uncertainty regarding generalizability \cite{AbdulGafoor2022, Setiawan2019}.

Recent advances in deep learning (DL) provide new opportunities to overcome these limitations by learning complex, high-dimensional relationships directly from multispectral imagery. Convolutional neural networks have been increasingly explored for SDB \cite{Lv2025, AlNajar2022, Yang2023, Zhong2022}; however, many existing DL studies rely on shallow architectures or limited training datasets, yielding performance comparable to traditional ML and, in some cases, production of spatial artifacts in predictions \cite{Qian2025, LumbanGaol2021}. Systematic investigations into DL-based SDB transferability across geographically distinct regions remain limited.

In this study, we present an extensive comparison of classical machine learning and modern high-capacity deep learning approaches for scalable and transferable satellite-derived bathymetry under relatively clear-water reef conditions. A Random Forest model and four pretrained deep learning architectures---ResNet-50, ResNet-101, EfficientNet-B4, and ConvNeXt-Large---are evaluated using Sentinel-2 imagery as input and reference bathymetry from airborne LiDAR at Pratas Island and a 30-m DEM for the Great Barrier Reef. Training samples are jointly constructed from Pratas Island and selected outer-shelf and oceanic coral reef systems in the Great Barrier Reef, while other, geographically remote, holdout regions are reserved for testing to assess model generalizability.

The scope of this work is intentionally focused on what general-purpose, off-the-shelf machine learning and deep learning architectures can achieve for Sentinel-2 shallow-water bathymetry, rather than on custom physics-informed models. The training and evaluation setup is, to our knowledge, more rigorous than that adopted in most comparable SDB studies: training data are jointly drawn from Pratas Island and selected reef systems of the Great Barrier Reef, while Ashmore and Cartier reefs serve as truly spatially-independent hold-out regions during evaluation. Many prior DL-based SDB studies instead rely on within-region random splits between training, validation, and test pixels, which can substantially overestimate transferability by allowing models to learn site-specific spectral--depth relationships. We therefore expect---and observe---larger cross-regional errors than are typically reported, reflecting a more honest evaluation rather than reduced model quality. Within this scope, our results apply to outer-shelf and oceanic coral reef systems under relatively clear-water conditions, and performance in turbid, inshore, or fringing environments may differ substantially; we do not claim global transferability or operational survey-grade accuracy. Closing this residual gap is the focus of complementary work pursued in parallel.

We further examine the influence of key training configurations, including patch size to control spatial context, spatial continuity through either random patch sampling or contiguous blocks with 50\% stride, and loss-function design. To enhance shallow-water learning, we propose a Smooth Weight Function (SWF)-weighted RMSE loss and compare it with conventional loss formulations.

The best-performing configurations are identified to guide future large-scale SDB applications.

\section{Materials and Data}
\label{sec:data}

\subsection{Study Areas}
\label{subsec:study_areas}

This study uses multiple coral reef environments spanning the South China Sea and Australian waters to evaluate both intra-regional performance and cross-regional generalization of satellite-derived bathymetry models.

\subsubsection{Pratas Island}
\label{subsubsec:pratas}

\begin{figure}[H]
\centering

\begin{minipage}{1.0\textwidth}
    \centering
    \includegraphics[width=0.75\textwidth, height=0.35\textheight, keepaspectratio]{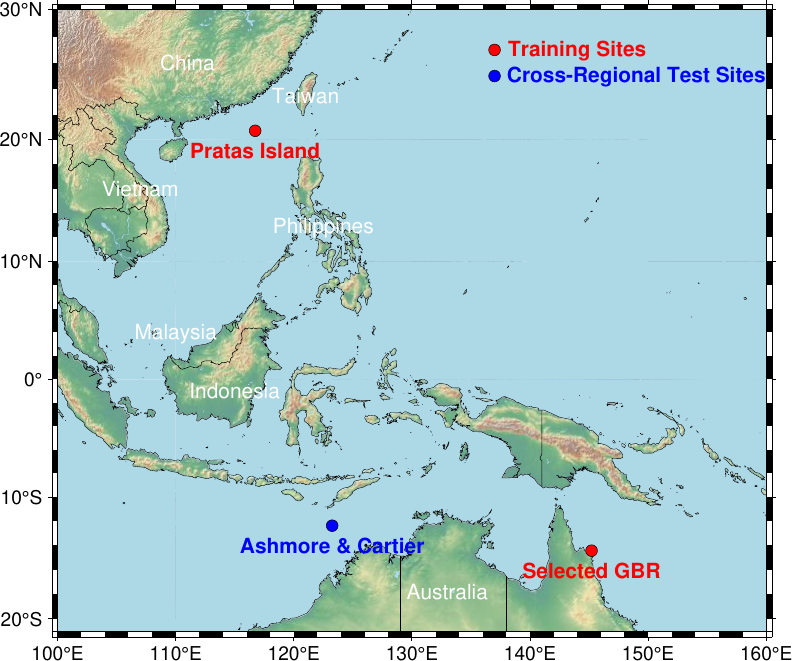}
    \caption*{\centering(a) Study area map}
\end{minipage}
\vspace{1em}

\begin{minipage}[t]{0.49\textwidth}
    \centering
    \includegraphics[height=0.22\textheight, width=\textwidth, keepaspectratio]{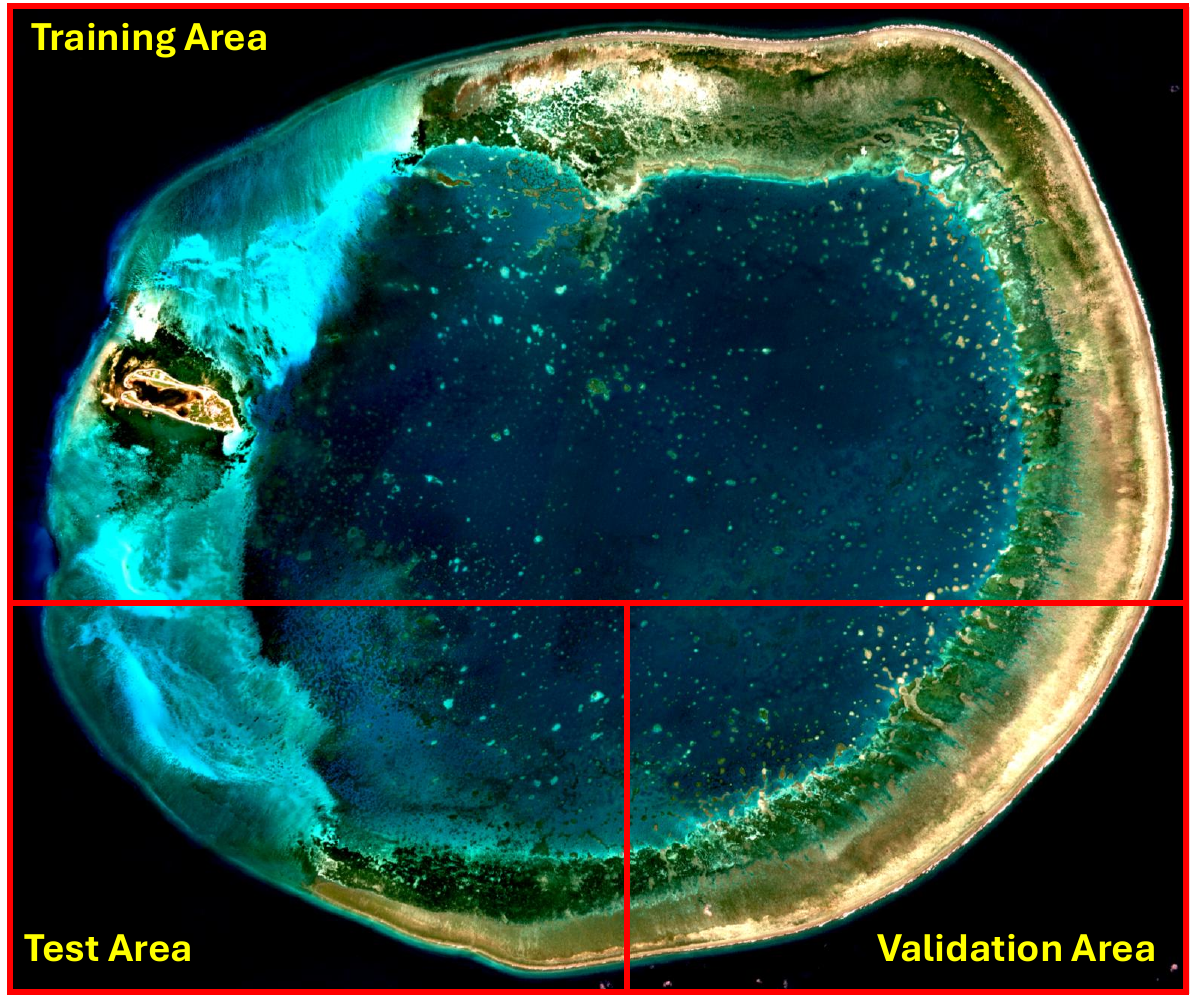}
    \caption*{\centering(b) Pratas Island (Dongsha Atoll)}
\end{minipage}
\hfill
\begin{minipage}[t]{0.49\textwidth}
    \centering
    \includegraphics[height=0.22\textheight, width=\textwidth, keepaspectratio]{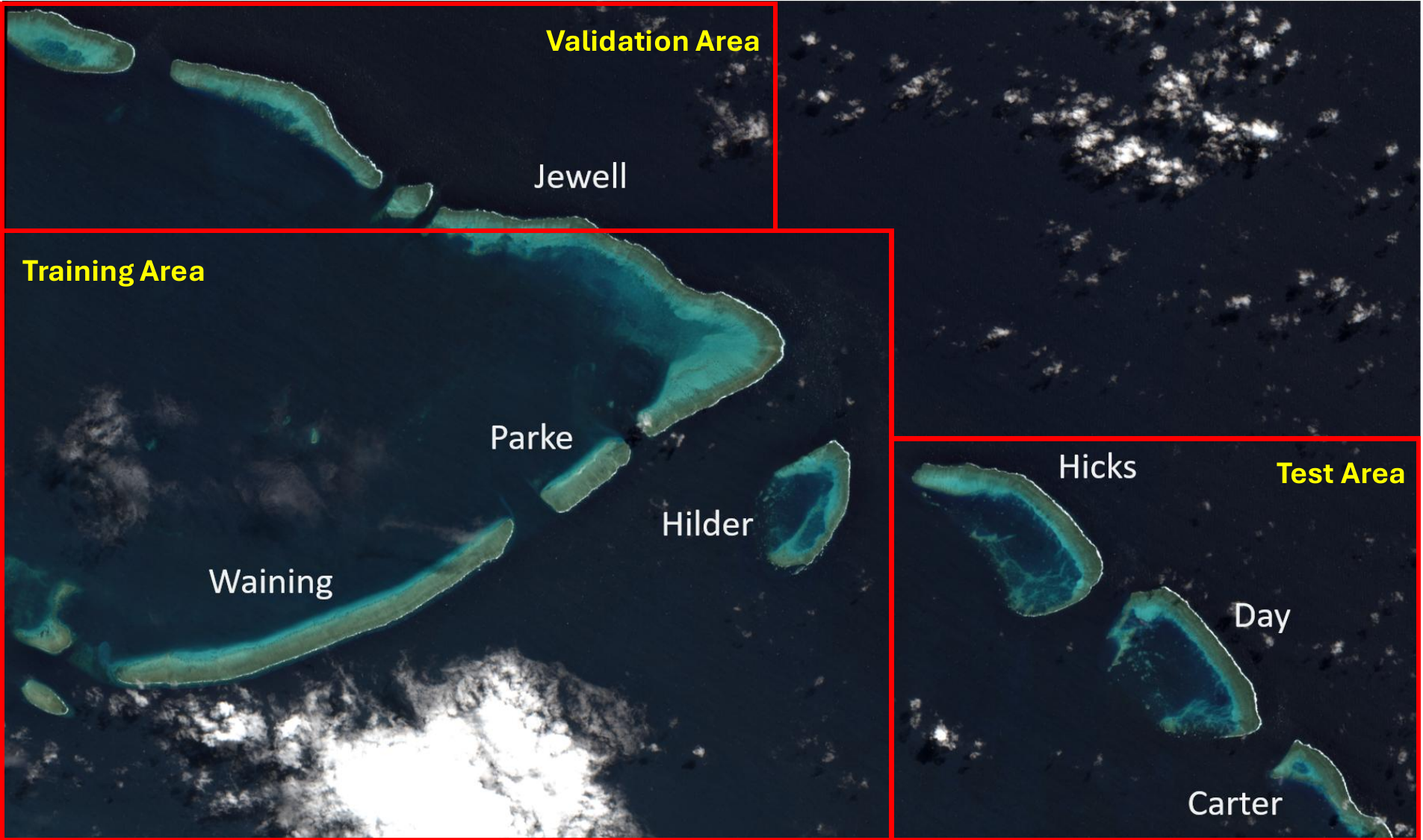}
    \caption*{\centering(c) Great Barrier Reef reef systems}
\end{minipage}
\vspace{1em}

\begin{minipage}[t]{0.49\textwidth}
    \centering
    \includegraphics[height=0.22\textheight, width=\textwidth, keepaspectratio]{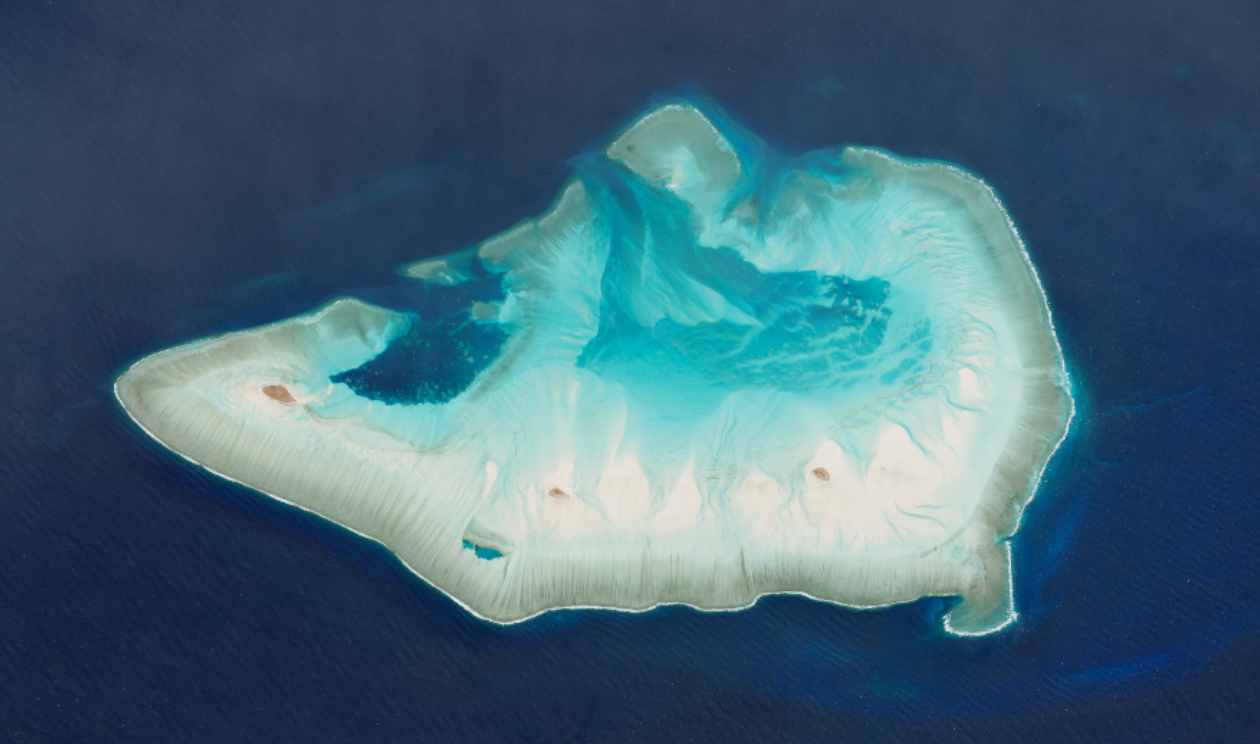}
    \caption*{\centering(d) Ashmore Reef}
\end{minipage}
\hfill
\begin{minipage}[t]{0.49\textwidth}
    \centering
    \includegraphics[height=0.24\textheight, width=\textwidth, keepaspectratio]{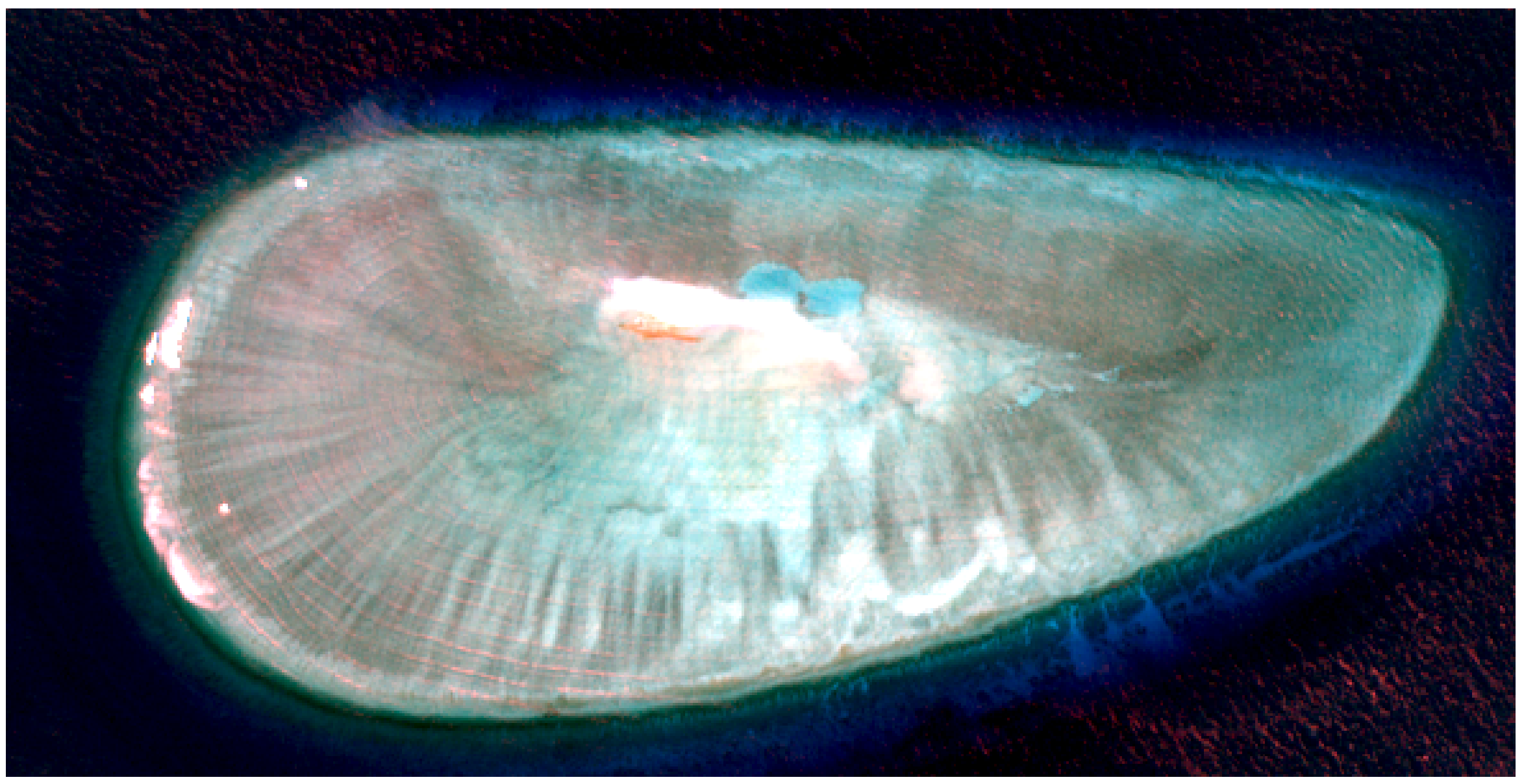}
    \caption*{\centering(e) Cartier Reef}
\end{minipage}
\vspace{6pt}
\caption{Study areas considered in this work. Panel (a) shows all site locations. Panels (b) and (c) show Pratas Island and the Great Barrier Reef, respectively, which are partitioned into training, validation, and intra-regional test regions. Panels (d) and (e) show the independent test sites at Ashmore Reef and Cartier Reef, which are used only to evaluate cross-regional generalization.}
\label{fig:study_areas}
\end{figure}

Pratas Island (Dongsha Island) is a ring-shaped coral atoll located in the northern South China Sea (Fig.~\ref{fig:study_areas}a). The atoll seafloor is complex and characterized by extensive seagrass meadows, sandy shorelines, and well-developed coral reef systems surrounding a shallow central lagoon. Water depths within the lagoon generally range from the surface to approximately 20~m, and the atoll has an overall diameter of about 25~km. The outer reef platform is predominantly very shallow, with large areas shallower than 2~m.

The waters around Pratas Island are typically clear due to its offshore location and limited terrestrial sediment input. Tidal variability is relatively small compared with typical coastal environments, with residual tidal corrections on the order of centimeters as estimated by the DTU23 global ocean tide model \cite{Andersen2023} (see Section~\ref{subsec:preprocessing}).
Spatial tidal differences across the atoll are (modeled as) less than 1~cm from west to east. Despite the generally favorable optical conditions, satellite imagery may still be affected by thin clouds, haze, surface waves, currents, and sun-glint effects \cite{Hedley2005}.

High-resolution airborne topo-bathymetric LiDAR data were acquired over Pratas Island in September 2010 by the Ministry of the Interior, Taiwan \cite{Shih2011}, providing reference depth measurements at 5~m~$\times$~5~m resolution used as ground-truth labels. The original survey reports an external vertical RMSE of approximately 0.3~m relative to sonar measurements, with increased error over complex reef morphology, which propagates to the interpolated gridded product used here.

In this study, Pratas Island is spatially partitioned into training and independent validation subsets to enable within-site generalization testing. Its representative shallow-water depth range and high-quality reference data make it a key component of the multi-regional training and evaluation framework.

\subsubsection{Great Barrier Reef}
\label{subsubsec:gbr}

Several coral reef systems within the Great Barrier Reef (GBR) were selected to support both model training and independent testing (Fig.~\ref{fig:study_areas}c). The study sites include Jewell, Parke, Waining, Hilder, Hicks, Day, and Carter Reefs, all located in the northeastern sector of the GBR and contained within a single Sentinel-2 tile. Human activities such as sightseeing and fishing at Jewell and Waining Reefs may influence local water conditions and optical properties, introducing additional variability for model learning.

To assess spatial generalization within the GBR, a subset of reefs (Jewell, Parke, Waining, and Hilder) is used jointly with Pratas Island for model training, while the remaining reefs (Hicks, Day, and Carter) are reserved exclusively for independent intra-regional testing. This split enables evaluation of cross-reef generalization within a geographically coherent but optically diverse environment.

These sites represent typical tropical reef geomorphologies, including shallow reef flats, fore-reef slopes, and lagoonal environments with variable substrate composition and water optical properties. Reference bathymetry for the GBR is obtained from the AusBathyTopo regional-scale depth model, compiled into a 30-m digital elevation model (DEM) \cite{Beaman2017}, and is treated as ground truth for model validation.

\subsubsection{Cross-Regional Test Sites}
\label{subsubsec:Cross_Regional_sites}

To further evaluate the scalability and geographic generalization capability of the proposed satellite-derived bathy\-metry models, two independent cross-regional test sites were selected in northwestern Australia: Ashmore Reef and Cartier Reef. These sites are geographically distant from both Pratas Island and the Great Barrier Reef and exhibit distinct geomorphological and optical characteristics, providing a rigorous assessment of cross-regional model generalization.

Ashmore Reef is located in the Timor Sea ($\sim$12$^\circ$11$'$S, 122$^\circ$59$'$E), approximately 840~km west of Darwin. It forms a shelf-edge reef complex of roughly 583~km$^2$ with a lagoon, reef flat, and three small islands, providing diverse shallow-water conditions for evaluation.

Cartier Reef lies approximately 70~km east of Ashmore ($\sim$12$^\circ$31$'$S, 123$^\circ$33$'$E). It is a compact platform reef rising steeply from deep surrounding waters, with sharp bathymetric gradients and clear-water conditions well suited for cross-regional generalization testing.

Neither Ashmore nor Cartier reefs were used during model training or validation; instead, they serve exclusively as unseen inference sites to assess true cross-regional generalization. The locations and corresponding Sentinel-2 imagery of the two test sites are shown in Fig.~\ref{fig:study_areas}d and Fig.~\ref{fig:study_areas}e.

Within the Pratas Island and the GBR, bathymetry-relevant regions are spatially partitioned into 60\% training, 20\% validation, and 20\% intra-regional test subsets, as illustrated by the boxed regions in Fig.~\ref{fig:study_areas}b and Fig.~\ref{fig:study_areas}c. This partitioning is applied only over areas containing valid shallow-water bathymetry rather than over the full Sentinel-2 tile.

\subsection{Satellite Imagery}
\label{subsec:satellite_data}

Satellite imagery used in this study was acquired from the Sentinel-2 constellation, which provides multispectral observations with a nominal revisit time of approximately 3--5 days at the equator. The MultiSpectral Instrument (MSI) provides 13 spectral bands from the coastal aerosol (443~nm) through the shortwave infrared (2190~nm).

In this study, Sentinel-2 Level-2A (L2A) products providing bottom-of-atmosphere (BOA) surface reflectance were accessed via Google Earth Engine (GEE) \cite{Gorelick2017}. The GEE archive excludes Band~10 (cirrus), which is primarily used for cloud detection and does not provide surface reflectance; its omission does not affect bathymetry model construction.

The (imperfect) Scene Classification Layer (SCL) associated with the L2A product was used to mask non-water pixels, including clouds, cloud shadows, vegetation, and land. All spectral bands were resampled to a uniform spatial resolution of 10~m via bilinear interpolation to ensure spatial consistency with the reference bathymetry datasets and model inputs.

\subsection{Reference (Ground Truth) Bathymetry Data}
\label{subsec:reference_data}

Multiple reference bathymetry datasets were collected from different regions to support model training and validation. These datasets were derived from various surveying techniques, including airborne LiDAR, multibeam sonar, and single-beam sonar, and were obtained from official national and regional data portals \cite{Shih2011, Beaman2017, Twiggs2023}.

The airborne LiDAR bathymetry data for Pratas Island are referenced to the WGS84 horizontal datum, with elevations provided as ellipsoidal heights relative to the WGS84 reference ellipsoid. Bathymetric data for the Great Barrier Reef (GBR) were obtained from the Geoscience Australia Portal and provided as a 30~m spatial resolution Digital Elevation Model (DEM), compiled from all available bathymetric source data within the GBR region. The GBR datasets are referenced to the WGS84 horizontal coordinate system, while the vertical datum is generally defined relative to local mean sea level.

Bathymetry data for Ashmore and Cartier reefs were also obtained from the Geoscience Australia Portal. These datasets are referenced to WGS84 / UTM coordinates, with vertical datums defined relative to EGM2008, and have a native spatial resolution of approximately 2~m.

Due to the heterogeneous nature of these datasets--originating from different surveying instruments, spatial resolutions, horizontal and vertical datums, and tidal correction standards--all bathymetric GeoTIFF files were reformatted to a unified spatial and vertical reference system. Specifically, all datasets were transformed to the WGS84 horizontal datum, converted to the EGM2008 vertical datum, and resampled to a consistent 10~m spatial resolution aligned with Sentinel-2 imagery. Water depth is defined as a positive quantity increasing downward from the water surface; therefore, larger depth values indicate deeper bathymetry.

\section{Methods}
\label{sec:methods}

\subsection{Data Preprocessing}
\label{subsec:preprocessing}

Satellite imagery preprocessing was performed using atmospherically corrected Sentinel-2 Level-2A (L2A) products. Only scenes with cloud coverage below approximately 10\% were selected. 

Prior studies have often combined multiple acquisitions into a single representative composite (e.g., median mosaics) to reduce temporal noise during training \cite{Mabula2023,Zhang2024}. In contrast, we retain \textit{repeat} imagery as independent samples during both training and inference to explicitly capture temporal variability associated with changing sun angles, atmospheric conditions, water optical properties, sun-glint \cite{Hedley2005}, and surface wave states.

Specifically, 6 repeat images were used for Pratas Island (Supplementary Fig.~S4), 4 images for the Great Barrier Reef (GBR; Supplementary Fig.~S5), and 10 images each for Ashmore and Cartier, all selected to have minimal cloud coverage. The Sentinel-2 Scene Classification Layer (SCL) was employed to mask non-water pixels, including clouds, cloud shadows, snow and ice, vegetation, bare soil, and built-up areas, retaining only water-covered pixels for subsequent analysis (acknowledging that this layer is imperfect).

High-quality reference bathymetry was used as the target for model training and evaluation. All reference bathymetry data were resampled to 10 m spatial resolution and co-registered to the Sentinel-2 image grid. 

To account for tidal effects at the time of satellite image acquisition, tidal corrections were applied using the DTU23 global tide model \cite{Andersen2023}. The DTU23 model has a spatial resolution of 1/16$^\circ$ ($\sim$6 km at the equator) and includes 10 primary tidal constituents. For improved spatial consistency, tidal heights were computed separately for each grid cell at the exact Sentinel-2 acquisition timestamps. The variation in modeled tidal height across the selected Sentinel-2 acquisition times was relatively small at Pratas Island (approximately $\pm$20 cm) and larger in the Great Barrier Reef (approximately $\pm$70 cm). These values refer to differences in tidal height between the specific image acquisition times used in this study, rather than the full local tidal range. Applying tidal correction therefore improves consistency between multi-temporal observations and supports potential global bathymetry applications. We note, however, that global tide models may exhibit reduced accuracy in shallow and morphologically complex coastal waters due to unresolved local hydrodynamics and bathymetric controls, which can introduce residual uncertainty into the corrected depths.

Training patches were sampled from spatially designated areas, including part of Pratas Island and the Jewell, Parke, Waining, and Hilder Reefs in the GBR. Independent testing patches were extracted from held-out regions, including the remaining portion of Pratas Island and the Hicks, Day, and Carter Reefs, to ensure spatial independence between training and evaluation. This spatial partitioning strategy avoids information leakage and enables robust assessment of model generalization across sites.

For model training and evaluation, the co-registered datasets were divided into image patches. For the Random Forest model, patches of size $64 \times 64$ pixels were extracted. Although Random Forest operates on individual pixels, this patch-based strategy improves spatial independence between training and validation samples.

For the deep learning models, we evaluated patch sizes of $128 \times 128$ and $512 \times 512$ pixels. Larger patches ($1024 \times 1024$) were considered to further increase spatial context, but were not feasible under available GPU memory constraints. The intermediate $512 \times 512$ size provided the best practical balance between contextual information, computational efficiency, and network receptive field coverage (see Section~\ref{sec:dl_experimental_design}). Adjacent patches overlapped by 50\% to reduce edge artifacts and improve training stability.

Depth filtering was applied to focus training targets on shallow-water regions with depths less than 20~m. Pixels corresponding to land, invalid values, or non-water areas were excluded. For deep learning models, pixels deeper than 20~m were retained in the input patches but masked from loss computation. While these deeper pixels did not contribute to the loss gradient, their presence provided important spatial context--such as reef edges and spectral contrasts-that helped the network resolve adjacent shallow-water features (see Section~\ref{sec:masked_loss}).

To enhance model generalization and increase the effective training sample size, data augmentation was applied to the deep learning datasets. Augmentation operations included rotations of $90^\circ$, $180^\circ$, and $270^\circ$, as well as horizontal flipping, ensuring that spatial patterns and depth relationships were preserved while exposing the models to varied orientations of the input imagery.

\subsection{Model Architectures}
\label{subsec:models}

This study compares a classical machine learning model (Random Forest) with several state-of-the-art deep learning models implemented within a unified DeepLabV3+ framework \cite{Chen2018}. Different encoder backbones (ResNet50, ResNet101, EfficientNet-B4, and ConvNeXt-Large) were used within DeepLabV3+ to evaluate performance, transferability, and scalability for satellite-derived bathymetry.

Random Forest was implemented using scikit-learn \cite{Pedregosa2011}, while all deep learning models were implemented in PyTorch \cite{Paszke2019} using the Segmentation Models PyTorch (SMP) library \cite{Iakubovskii2019}. ImageNet-pretrained \cite{Deng2009} encoder weights were used for initialization 
and subsequently fine-tuned on the bathymetry datasets.

\subsubsection{Random Forest}

Random Forest (RF) is an ensemble learning method that combines multiple decision trees, with the final prediction obtained by averaging individual tree outputs. By training each tree on a random subset of samples and features, RF reduces overfitting and improves robustness on large, heterogeneous geospatial datasets. In this study, the RF model consisted of 100 decision trees.

RF models have been frequently applied in SDB, but exhibit limited transferability because their decision boundaries are tied to local reflectance-depth relationships that depend on water clarity, bottom type, and illumination. As a result, models trained in one region often perform poorly in optically different environments. Deep learning models, with hierarchical feature representations, can capture more complex, nonlinear relationships across multiple spectral bands, potentially improving cross-regional generalization.

\subsubsection{ResNet Architectures}

ResNet50 and ResNet101 are deep convolutional neural networks based on residual learning \cite{He2016}, employing skip connections between encoder and decoder branches to facilitate gradient propagation. ResNet50 (50 layers) balances representational capacity and efficiency, while ResNet101 (101 layers) can model more complex spectral-depth relationships at higher computational cost. Both were initialized with ImageNet-pretrained weights, providing strong spatial feature extraction for multispectral imagery.

\subsubsection{EfficientNet-B4}

EfficientNet-B4 \cite{Tan2019} employs compound scaling of network depth, width, and input resolution, using mobile inverted bottleneck convolution (MBConv) blocks and the Swish activation function. These design choices achieve competitive performance with significantly fewer parameters than many conventional deep CNNs, making it an attractive candidate for large-scale bathymetry mapping.

\subsubsection{ConvNeXt-Large}

ConvNeXt-Large \cite{Liu2022} is a modern convolutional architecture inspired by Vision Transformers, incorporating a patchify stem, inverted bottleneck structures, and large $7\times7$ depthwise convolutional kernels to capture long-range spatial dependencies. With approximately 198 million parameters, it offers high representational capacity particularly suited to complex reef environments where depth patterns reflect both local seabed structure and broader geomorphological organization, though it requires substantial training data and compute resources.

All deep learning backbones were initialized with ImageNet-1k pretrained weights and fine-tuned for bathymetry estimation. Although ImageNet pretraining is based on natural RGB imagery, early convolutional layers learn generic spatial primitives (e.g., edges, gradients, and texture transitions) that remain transferable to multispectral remote sensing data. Leveraging pretrained models improves training stability and facilitates a systematic comparison of architectural design paradigms under a consistent optimization framework.

Table~\ref{tab:model_params} summarizes the architectures, pretraining status, and approximate parameter counts for all models evaluated in this study.

\begin{table}[H]
\caption{Overview of the ML and DL architectures evaluated in this study, including architecture type, pretraining status, and approximate numbers of trainable parameters.}
\label{tab:model_params}
\begin{tabular}{lccc}
\hline
\textbf{Model} & \textbf{Architecture Type} & \textbf{Pretrained} & \textbf{Parameters (Millions)} \\
\hline
Random Forest & Ensemble (Decision Trees) & No & -- \\
ResNet50 & CNN (Residual) & ImageNet-1k & $\sim$39.7 \\
ResNet101 & CNN (Residual) & ImageNet-1k & $\sim$58.7 \\
EfficientNet-B4 & CNN (Compound Scaling) & ImageNet-1k & $\sim$18.6 \\
ConvNeXt-Large & Modern CNN & ImageNet-1k & $\sim$198.8 \\
\hline
\end{tabular}
\end{table}

\subsubsection{Training Configuration and Input Transformation}
\label{sec:dl_Training_Configuration}

The Random Forest (RF) model used Sentinel-2 Level-2A BOA surface reflectance values directly as input features without logarithmic transformation. In contrast to deep learning models that are sensitive to feature scaling, RF is inherently invariant to monotonic transformations because its decision rules depend on feature ordering rather than absolute magnitude \cite{Breiman2001,Hastie2009}. Empirical studies using RF regression have also shown that feature normalization or standardization often leads to only minor differences in predictive performance \cite{Bonamutial2023}. As a result, RF was trained directly on raw reflectance values to maintain a simple and consistent baseline formulation.

All deep learning models were trained using the proposed SWF-weighted RMSE loss, with a default patch size of $512 \times 512$ pixels and all 12 Sentinel-2 bands as input. Training employed a batch size of 16 and the AdamW optimizer with an initial learning rate of $10^{-4}$. Early stopping was applied based on validation loss (patience of 10 epochs), and the learning rate was reduced after five consecutive plateau epochs. Models were trained for up to 100 epochs, although convergence typically occurred within 20--50 epochs.

For deep learning inputs, a logarithmic transformation was applied to reflectance values, following classical satellite-derived bathymetry formulations \cite{Lyzenga1985,Stumpf2003}, where depth is approximately linear in log-transformed reflectance. For each pixel location $x$ with reflectance $R(x)$:
\begin{equation}
\mathbf{X}(x) = \ln\left(\frac{R(x)}{10^{4}} + \epsilon \right),
\end{equation}
where $\epsilon = 10^{-6}$ prevents numerical instability and $10^{4}$ normalizes the integer-scaled reflectance values. This transformation reduces multiplicative illumination effects and improves numerical stability during training in optically shallow waters. We emphasize that the log-reflectance encoding acts as a lightweight physics-inspired inductive bias on the network input, rather than enforcing an explicit physical attenuation model on the learned mapping.

%

\subsubsection{Deep Learning Experimental Design}
\label{sec:dl_experimental_design}

Although deep learning is inherently data-driven, hyperparameter selection remains non-trivial, particularly for input patch size and loss function. Patch size determines the effective spatial receptive field of the network and therefore controls the extent to which predictions are informed by local radiometric information versus broader multi-scale spatial context. In shallow-water bathymetry, observed reflectance is influenced by illumination conditions, adjacency effects, and water-column optical properties, which introduce spatial dependencies that extend beyond individual pixels and motivate the need for sufficient spatial context.

The receptive field determines how much spatial context a network can integrate. The \emph{effective receptive field} is typically much smaller than the theoretical field due to Gaussian weighting and architectural constraints \cite{Luo2016}, meaning adequate patch size remains important even for deep architectures. To evaluate fine-scale representation, an initial diagnostic experiment used small $128 \times 128$ patches (Supplementary Section~S2), while a depth-capability diagnostic determined the practical retrieval range (Supplementary Section~S3), motivating the operational focus on 0--20~m.

For operational mapping, $512 \times 512$ patches were adopted as the default. The theoretical receptive fields of ResNet50, ResNet101, EfficientNet-B4, and ConvNeXt-Large are approximately $427\times427$, $971\times971$, $380\times380$, and $224\times224$ pixels, respectively \cite{Kim2023, Richter2022}; however, the effective receptive field is substantially smaller in practice, so patch size remains the primary constraint on usable spatial context. At Sentinel-2's 10~m resolution, $512\times512$ pixels corresponds to $\sim$5~km -- sufficient to capture complete reef units (lagoon, reef flat, fore-reef slope, and adjacent deeper water) within a single sample, consistent with the characteristic geomorphic dimensions of the Pratas and GBR study sites. Larger $1024\times1024$ patches were tested but consistently caused out-of-memory failures on the 48~GB NVIDIA A40/L40/L40S GPUs used for training the high-capacity architectures, and reducing the batch size to fit them led to degraded convergence in preliminary tests. The $512\times512$ patch size therefore represents the practical upper limit under the current hardware while preserving the spatial context required to capture complete reef geomorphic units.

To systematically evaluate model performance, several complementary experiments were conducted:

\begin{itemize}
     \item \textbf{Loss-function comparison:} RMSE, relative percentage error (RPE), and the proposed Smooth Weight Function (SWF)-weighted RMSE loss were compared to assess their impact on predictive accuracy, particularly for shallow-water depths, and on generalization across regions.
    \item \textbf{Spatial continuity strategy:} Two data-splitting strategies were tested: a conventional random patch split (Strategy~1) and a spatially continuous split (Strategy~2). Strategy~2 preserves local geomorphic and radiometric structure, enabling the network to learn coherent spectral--depth transitions, and was evaluated with and without augmentation to isolate effects of spatial continuity versus training volume.
    \item \textbf{Cross-regional generalization:} Models trained on Pratas Island and GBR regions were tested on spatially independent reefs (Ashmore and Cartier) to evaluate geographic transferability.
    \item \textbf{Benchmarking against existing networks:} The proposed architectures were compared with published U-Net-based SDB networks using public datasets to contextualize model complexity and accuracy.
\end{itemize}

All architectures were trained under identical preprocessing, augmentation, and optimization settings to ensure a fair comparison. Together, these experiments quantify the influence of patch size, receptive field, loss function, and spatial data organization on model performance and cross-regional generalization within the operational 0--20 m depth range.

\subsection{Accuracy Assessment}
\label{subsec:iho_eval}

Model performance was quantified using Root Mean Square Error (RMSE) between predicted and reference bathymetric depths. Depth-dependent performance was further analyzed using 1-m depth bins to evaluate error variation as a function of water depth.

As a familiar reference scale for hydrographic accuracy, results are also placed in the context of the International Hydrographic Organization (IHO) S-44 standard for hydrographic surveys \cite{IHO2020}, which was originally developed for acoustic hydrographic surveying. The S-44 standard defines Total Vertical Uncertainty (TVU) as:
\begin{equation}
TVU = \pm \sqrt{a^2 + (b \cdot d)^2}
\end{equation}
where \(a\) represents the depth-independent uncertainty component, \(b\) is the depth-dependent coefficient, and \(d\) is water depth, with values of 
 \(a = 0.50~\mathrm{m}\) and \(b = 0.013\) for Order 1a/1b, and \(a = 1.00~\mathrm{m}\) and \(b = 0.023\) for Order 2 \cite{IHO2020}.

Depth-binned RMSE values are reported alongside the IHO Order 1 and Order 2 TVU thresholds purely as a familiar accuracy reference, not as a target that Sentinel-2-based satellite-derived bathymetry is expected to meet without dedicated hydrographic-grade processing. As the IHO S-44 TVU is formally defined at the 95\% confidence level, figures additionally display the 95\% confidence interval (1.96\,$\times$\,RMSE) alongside the RMSE curves to enable a statistically consistent comparison.

\section{Masked and Depth-Aware Loss Functions}
\label{sec:loss_functions}

To improve training efficiency and predictive accuracy, loss computation was restricted to relevant coastal areas using a combined water and depth mask. This focuses training on pixels that contribute significant bathymetric signal-primarily shallow to moderate depths-while ignoring non-water or deep areas that provide inconsistent information for model optimization. Masking was applied consistently for all loss functions considered in this study, including RMSE, RPE, and our proposed SWF-weighted RMSE. Depths greater than 20~m were still included such that the network can learn context from deeper depths but not contribute to the loss function (i.e.~not optimizing for deeper waters).

\subsection{Water and Depth Masking}

First, a water mask $M_{\mathrm{water}}(x) \in \{0,1\}$ was derived from the Sentinel-2 Scene Classification Layer (SCL):
\begin{equation}
M_{\mathrm{water}}(x) =
\begin{cases}
1, & \text{if SCL indicates open or coastal water}, \\
0, & \text{otherwise.}
\end{cases}
\end{equation}

Next, a depth-dependent mask emphasizes shallow-water learning by excluding pixels with depths greater than 20~m from the loss function:
\begin{equation}
M_{\mathrm{depth}}(x) =
\begin{cases}
1, & \text{if } Z(x) \leq 20~\mathrm{m}, \\
0, & \text{if } Z(x) > 20~\mathrm{m}.
\end{cases}
\end{equation}

The final effective mask is obtained as the element-wise product:
\begin{equation}
M(x) = M_{\mathrm{water}}(x) \cdot M_{\mathrm{depth}}(x).
\end{equation}

During training, the model receives spectral input and target depths from all pixels, but only shallow-water pixels defined by $M(x)$ contribute to the loss and parameter updates. 

\subsection{Loss Functions}
\label{subsec:loss_functions}
\subsubsection{Masked RMSE and RPE}
\label{sec:masked_loss}

A standard loss function is the masked Root Mean Squared Error (RMSE), which accounts for invalid or masked pixels:
\begin{equation}
N_M = \sum_{i=1}^{N} M_i, \qquad
\mathrm{RMSE} = \sqrt{
\frac{ \sum_{i=1}^{N} M_i \left( \widehat{Z}_i - Z_i \right)^2 }{N_M}
},
\end{equation}
where $N$ is the total number of pixels in the patch or image, $M_i$ is a binary mask indicating valid water pixels (1 = valid, 0 = masked), $\widehat{Z}_i$ is the predicted depth, and $Z_i$ is the reference depth for pixel $i$. RMSE penalizes absolute errors equally across all depths, meaning that a 1~m error at 1~m depth contributes the same to the total RMSE as a 1~m error at 20~m depth. While simple and widely used, RMSE tends to emphasize deeper pixels at the expense of shallow ones when real errors increase with depth. 

To improve sensitivity to shallow-water depths, a Relative Percentage Error (RPE) loss is also used:
\begin{equation}
\mathrm{RPE} = \frac{1}{N_M} \sum_{i=1}^{N} M_i \frac{\left| \widehat{Z}_i - Z_i \right|}{\left| Z_i \right|}.
\end{equation}

By normalizing errors relative to depth, RPE gives more weight to shallow regions.
\subsubsection{Smooth Weight Function (SWF)-Weighted RMSE}
\label{subsec:swf}

In an effort to balance the loss function, and thus the performance of the final trained model, we define a depth-aware Smooth Weight Function (SWF)-weighted RMSE (Eq.~\eqref{eq:swf_loss}). The SWF loss emphasizes shallow-water pixels while retaining meaningful contributions from intermediate and deeper pixels, providing a physically interpretable and numerically robust objective function.

The SWF-weighted RMSE is defined as:
\begin{equation}
\mathcal{L}_{\mathrm{SWF}} = 
\sqrt{
\frac{ \sum_{i=1}^{N} M_i w_i \left( \widehat{Z}_i - Z_i \right)^2 }{N_M}
},
\label{eq:swf_loss}
\end{equation}
where $w_i$ is a depth-dependent weighting factor:
\begin{equation}
w_i = 1 + \beta \exp\left(-\frac{|Z_i|}{Z_0}\right).
\end{equation}

Here, $\beta > 0$ controls the amplitude of shallow-water emphasis, and $Z_0$ defines the characteristic decay depth at which the weighting decreases by a factor of $1/e$. The smooth exponential formulation ensures continuous gradients, stable optimization, and progressively reduced influence of deeper regions.

The decay depth $Z_0 = 10$~m was chosen to align with the operationally critical shallow-water zone: depths shallower than 10~m are particularly relevant for safe navigation, where even moderate bathymetric errors can pose risks to vessel transit over reef structures. This threshold also coincides with the depth range where Sentinel-2 optical retrievals are most reliable under clear-water conditions, making it a physically and operationally motivated boundary. The amplitude parameter $\beta = 5$ ensures that shallow-water pixels receive substantially higher loss weight while $w_i > 1$ is maintained across the full depth range, preserving gradient contributions from deeper pixels. Under the adopted configuration ($\beta = 5$, $Z_0 = 10$~m), the weighting function assigns a maximum weight of 6.0 at the surface ($Z = 0$~m), decreasing smoothly to 4.03 at 5~m, 2.84 at 10~m, and 1.68 at 20~m depth (Supplementary Fig.~S6).

The SWF loss function is not intended to minimize overall RMSE, but rather to allow flexible control over where prediction accuracy is emphasized across the depth range. By contrast, standard RMSE is by definition heavily influenced by deeper pixels, which dominate the training distribution. The SWF formulation allows the user to deliberately shift accuracy toward shallower or deeper regimes depending on the application of interest: a narrower decay depth $Z_0$ concentrates emphasis on very shallow water, while a larger $Z_0$ broadens the zone of emphasis toward intermediate depths. Like the log-reflectance encoding, the SWF loss acts as a lightweight physics-inspired inductive bias---shaping the loss surface so that shallow-water regions, where optical attenuation makes retrieval most challenging, receive proportionally greater optimization emphasis---without imposing an explicit water-column physical model. An empirical sensitivity analysis over a $3 \times 3$ grid of $(\beta, Z_0)$ combinations, demonstrating this depth-regime shifting behaviour, is reported in Section~\ref{subsubsec:swf_hyperparam_sensitivity}.

\section{Results}
\label{sec:results}

\subsection{Random Forest Model Performance}
\label{subsec:results_rf}

The Random Forest (RF) model was trained as a baseline for comparison with deep learning approaches. A grid search with five-fold cross-validation showed that RF performance was largely insensitive to hyperparameter choices, with validation RMSE (used as the cross-validation scoring metric) varying by less than 0.01~m across configurations. A configuration with 100 trees was selected for computational efficiency and model parsimony.

\begin{figure}[H] 
    \centering
    \includegraphics[width=0.95\textwidth]{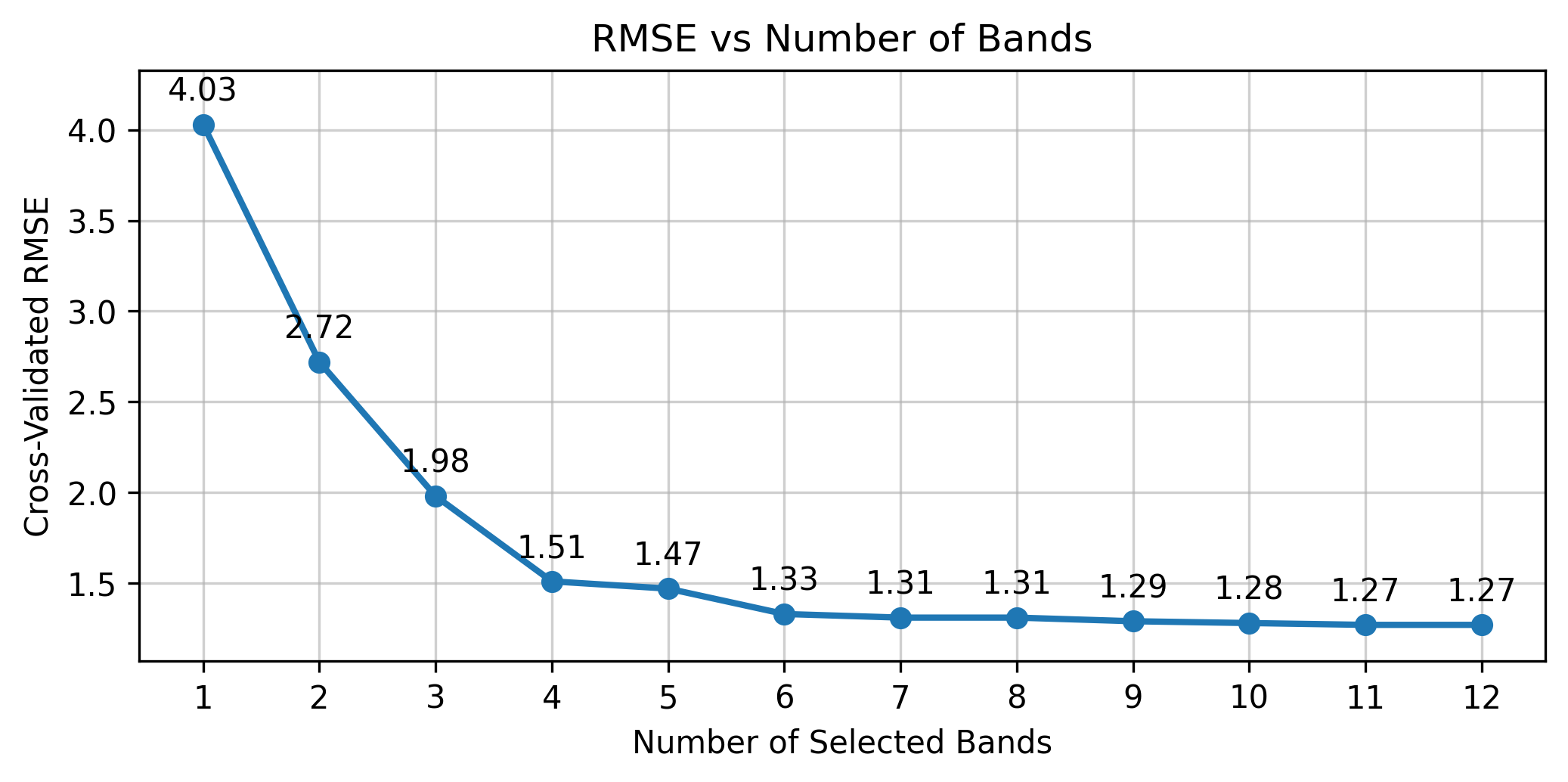}
    \caption{Relationship between RMSE and number of bands used for the RF model determined via recursive feature elimination (RFE). Each point represents performance for the top-k ranked feature subset (k = number of retained bands).}
    \label{fig:rf_rfecv_rmse_curve}
\end{figure}

Recursive Feature Elimination (RFE), which iteratively removes the least informative predictors based on mean decrease in impurity (MDI) feature importance scores, was used to identify the subset of Sentinel-2 bands yielding the lowest cross-validated RMSE. RFE indicated that performance stabilized when using five or more Sentinel-2 bands, with RMSE values ($\sim$ 1.27--1.33 m; Fig.~\ref{fig:rf_rfecv_rmse_curve}) showing minimal variation across this range. In this analysis, each point in Fig.~\ref{fig:rf_rfecv_rmse_curve} represents the cross-validated RMSE obtained from a Random Forest model trained on a specific subset of bands selected via recursive feature elimination, where bands are iteratively removed based on their contribution to predictive performance. This suggests limited benefit from including additional spectral bands beyond the most informative subset.

\vspace{-6pt}

Feature importance was quantified using permutation feature importance (PFI), in which each spectral band was randomly permuted and the resulting increase in root mean squared error (RMSE) was evaluated. The green band (B3) produced the largest degradation ($\Delta$RMSE = 6.20~m), followed by blue (B2; 1.89~m), red (B4; 1.00~m), red-edge 1 (B5; 0.91~m), and the aerosol band (B1; 0.47~m). All remaining bands increased RMSE by less than 0.15~m. This ranking is consistent with shallow-water optical physics \cite{Lyzenga1985,Stumpf2003}: visible and near-infrared bands carry water-leaving reflectance from the seabed, with longer wavelengths (red and red-edge) penetrating progressively shallower depths, while the shorter-wavelength infrared and SWIR bands are essentially absorbed within the first meter of the water column. Approximately five well-chosen bands therefore suffice for near-optimal predictive accuracy and twelve is overkill; in this study all twelve Sentinel-2 bands are nevertheless used as input in the main experiments, accepting a small efficiency cost to retain the lowest achievable RMSE.

The RF model achieved a training RMSE of 0.96~m, with performance decreasing to 1.18~m on the validation set and 1.53~m on the intra-regional test set ($R^{2} = 0.87$). In very shallow water ($\leq 3$~m), the RMSE was 0.89~m.

For cross-regional evaluation, the RF model was applied to the independent Ashmore and Cartier reefs. RMSE increased to 2.99--3.78~m, demonstrating reduced generalization to these geographically and optically distinct sites. These results establish the RF model as a robust baseline for locally trained bathymetry and provide a reference for comparison with deep learning models in Section~\ref{subsec:dl_results}.

\subsection{Deep Learning Performance}
\label{subsec:dl_results}

\subsubsection{Intra-Regional Performance} 


All DL models were trained using the SWF-weighted RMSE loss under identical settings (Section~\ref{sec:dl_Training_Configuration}). Table~\ref{tab:train_val_test_performance} summarizes model performance using $512 \times 512$ patches, including RMSE, mean absolute error (MAE), relative percent error (RPE), and coefficient of determination ($R^2$). ResNet50 and ResNet101 achieve the lowest test RMSE values (1.15~m and 1.26~m, respectively) across the full 0--20~m depth range, while also maintaining comparatively low MAE and RPE values and high $R^2$ scores (0.91 and 0.89, respectively). These models additionally achieve the best shallow-water performance, with within-3~m RMSE values below 0.3~m. EfficientNet-B4 and ConvNeXt-Large exhibit higher overall errors and lower $R^2$ values, particularly for shallow-water depths ($\leq 3$~m), whereas the Random Forest model yields competitive overall RMSE and relatively strong $R^2$ performance but substantially larger shallow-water error.

%
%
%
%

\begin{table}[H]
\caption{Intra-regional performance of deep learning and Random Forest models on cloud-free Sentinel-2 imagery using $512 \times 512$ patches. Metrics include RMSE, MAE, relative percent error (RPE), coefficient of determination ($R^2$), shallow-water RMSE for depths $\leq 3$~m, and training epochs.}
\label{tab:train_val_test_performance}

\resizebox{\textwidth}{!}{%
\begin{tabular}{lccc|ccc|ccccc|c}
\hline
\multirow{2}{*}{Model} &
\multicolumn{3}{c|}{Training} &
\multicolumn{3}{c|}{Validation} &
\multicolumn{5}{c|}{Test} &
\multirow{2}{*}{Epochs} \\
\cline{2-12}

& RMSE & MAE & RPE (\%)
& RMSE & MAE & RPE (\%)
& RMSE & MAE & RPE (\%)
& $R^2$
& $\leq$3 m RMSE
& \\
\hline

ResNet50
& 1.02 & 0.70 & 13.82
& 1.32 & 0.82 & 23.85
& 1.15 & 0.80 & 11.28
& 0.91
& 0.28
& 40 \\

ResNet101
& 0.95 & 0.63 & 10.69
& 1.29 & 0.77 & 18.64
& 1.26 & 0.90 & 12.17
& 0.89
& 0.26
& 33 \\

EfficientNet-B4
& 1.50 & 1.04 & 15.86
& 1.72 & 1.14 & 28.74
& 1.79 & 1.29 & 16.43
& 0.78
& 0.37
& 23 \\

ConvNeXt-Large
& 1.42 & 1.08 & 23.05
& 1.52 & 1.04 & 33.09
& 1.92 & 1.52 & 22.57
& 0.74
& 0.63
& 23 \\

Random Forest
& 0.96 & 0.60 & N/A
& 1.18 & 0.73 & N/A
& 1.53 & 1.02 & 13.63
& 0.87
& 0.89
& N/A \\

\hline
\end{tabular}
}
\end{table}

\begin{figure}[H]
   \centering
    \includegraphics[width=1\linewidth]{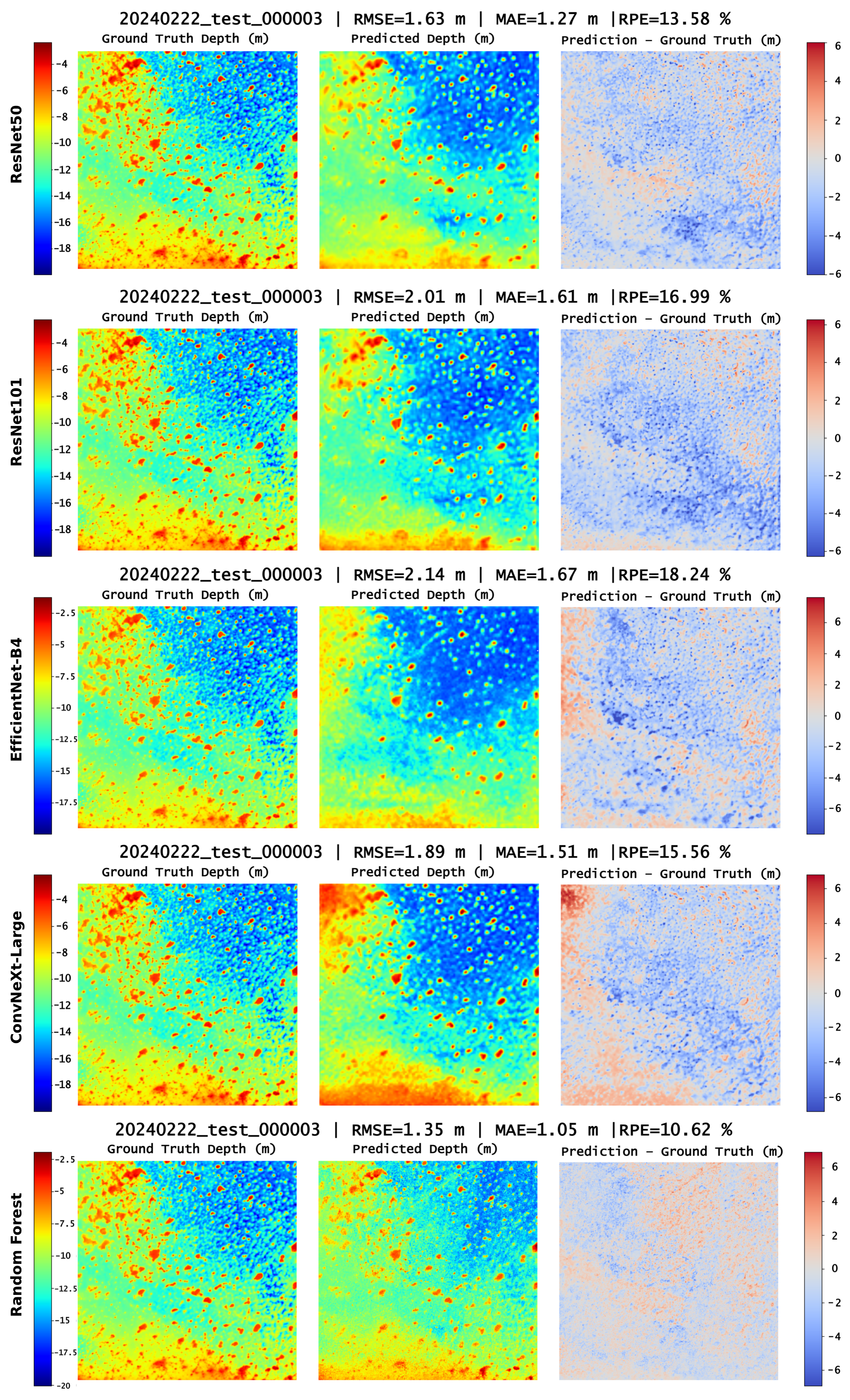}

    \caption{Spatial comparison of depth predictions ($0$--$20$~m). From top to bottom: ResNet50, ResNet101, EfficientNet-B4, ConvNeXt-Large, and Random Forest. Each row displays the LiDAR ground truth, the model prediction, and the corresponding residuals (m); the row labels report the test-set RMSE, MAE, and RPE for each architecture. The colorbars are consistent with each model's prediction range to facilitate direct visual comparison of model behavior.}
    \label{fig:512ps_comparisons}
\end{figure}

Fig.~\ref{fig:512ps_comparisons} shows that ResNet50 and ResNet101 preserve spatial detail and numerical accuracy. ConvNeXt-Large shows sharp predictions but a slight depth bias, EfficientNet-B4 produces smoother outputs, and Random Forest captures depth trends but lacks the fine geomorphic detail seen in DL models.

\subsubsection{Cross-Regional Generalization and Multi-Temporal Inference}
\label{subsubsec:cross_results}

\begin{figure}[H]
    \centering
    \includegraphics[width=1\textwidth]{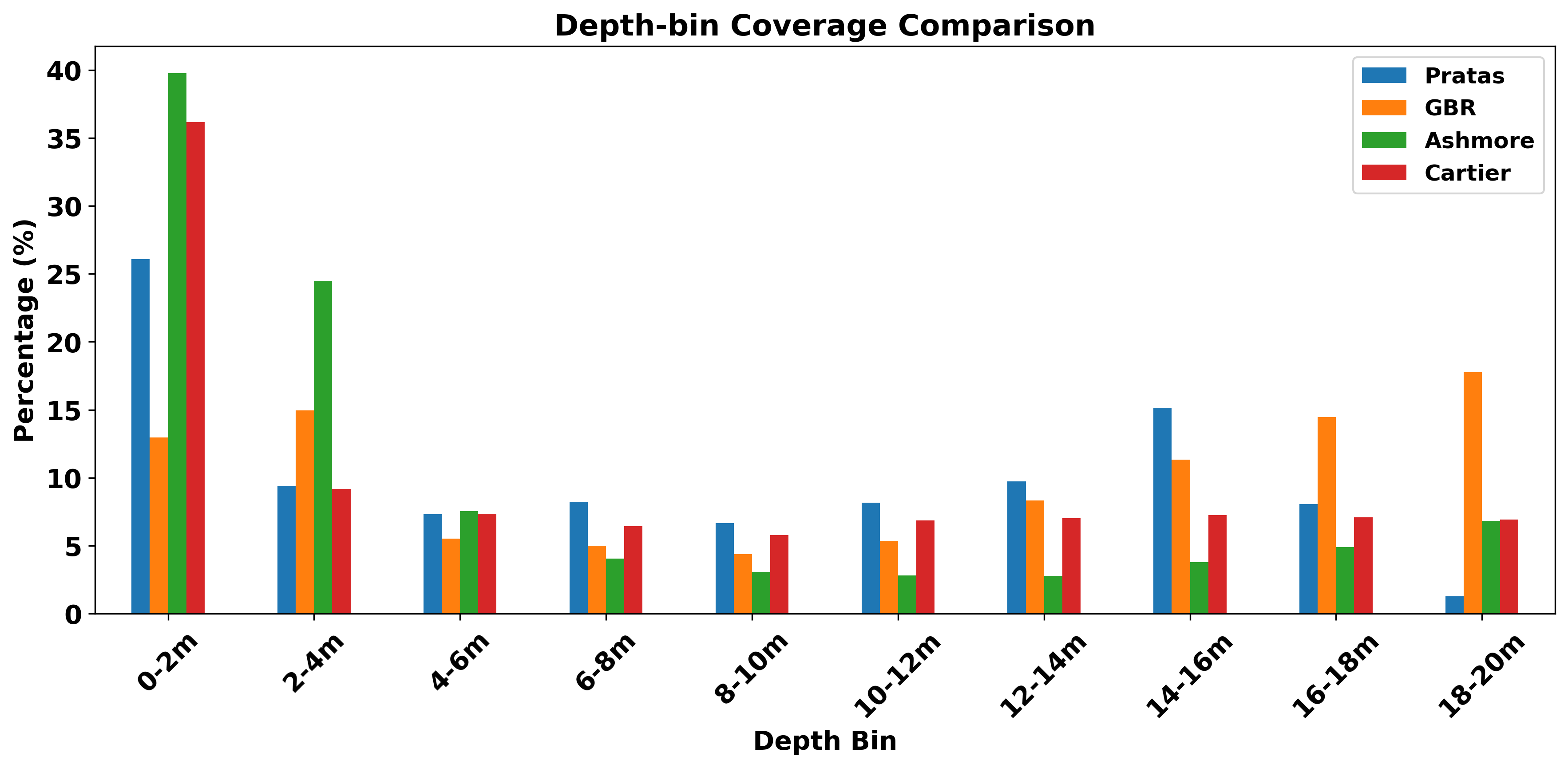}
    \caption{Depth-bin coverage comparison across the four study regions.}
    \label{fig:depth_bin_coverage_barplot}
\end{figure}

Before reporting cross-regional performance, we briefly characterize the depth distributions of the training and transfer regions, which provide important context for interpreting the results that follow. Table~\ref{tab:depth_distribution} summarizes depth composition statistics for the training and transfer regions. Ashmore is the most shallow-dominated region (79.0\%), followed by Cartier (64.6\%) and Pratas (61.3\%), while GBR contains the highest proportion of mid-depth pixels (53.7\%). Mean depth and standard deviation further indicate that GBR exhibits the deepest and most heterogeneous depth distribution, while Ashmore is strongly dominated by shallow waters, with both the lowest mean depth (5.38~m) and highest shallow-water proportion. Fig.~\ref{fig:depth_bin_coverage_barplot} illustrates the depth-bin distribution across the four regions, revealing distinct differences in shallow- and mid-depth coverage. These distributional differences mean that cross-regional accuracy reflects not only model transferability but also depth-distribution mismatch between training and inference domains.

\begin{table}[H]
\centering
\small
\setlength{\tabcolsep}{4pt}
\caption{Depth distribution statistics for training and transfer regions.}
\label{tab:depth_distribution}
\begin{tabular}{lccccc}
\hline
& \multicolumn{2}{c}{Depth Distribution} & \multicolumn{3}{c}{Statistics} \\
\cline{2-3} \cline{4-6}
Region & 0--10 m & 10--20 m & Mean Depth (m) & Std Depth (m) & Shallow Proportion (\%) \\
\hline
Pratas  & 2,421,537 & 1,528,479 & 8.12  & 5.86 & 61.3 \\
GBR     & 1,307,549 & 1,516,884 & 10.84 & 6.64 & 46.3 \\
Ashmore & 1,453,539 & 385,821   & 5.38  & 5.97 & 79.0 \\
Cartier & 47,908    & 26,231    & 7.13  & 6.48 & 64.6 \\
\hline
\end{tabular}
\end{table}

Table~\ref{tab:transfer_metrics} summarizes cross-regional prediction performance for Ashmore and Cartier reefs over the full 0--20~m depth range, aggregated across 10 Sentinel-2 acquisitions. These cross-regional RMSE values of 2.46--2.98~m for the best-performing DL models represent a moderate increase relative to intra-regional performance (1.15--1.92~m), with the degradation factor ranging from approximately 1.28$\times$ (ConvNeXt-Large at Cartier) to 2.50$\times$ (ResNet50 at Ashmore), reflecting partial transferability with a clear accuracy penalty under geographic shift. ConvNeXt-Large in particular shows the strongest relative robustness to cross-regional transfer of any model evaluated: its cross-regional RMSE values (2.46~m at Cartier and 2.98~m at Ashmore) are only 28\% and 55\% higher, respectively, than its intra-regional RMSE of 1.92~m, an average degradation of approximately 42\% across the two truly independent cross-regional reefs. In addition to RMSE and MAE, coefficient of determination ($R^2$) is reported to evaluate explained variance under cross-regional transfer. The best-performing models achieve median-aggregation $R^2$ values ranging from 0.69 to 0.85, indicating that substantial bathymetric variability remains recoverable despite regional differences in environmental conditions and seafloor characteristics.

RMSE is used here primarily to evaluate the effect of temporal aggregation strategies, while MAE provides a complementary measure less sensitive to extreme residuals. Median aggregation consistently reduces both RMSE and MAE relative to mean aggregation for all deep learning models, while also improving $R^2$ values, indicating improved robustness to outliers and temporally inconsistent predictions. ResNet50, ResNet101, and ConvNeXt-Large achieve the lowest median errors and highest $R^2$ values overall, whereas EfficientNet-B4 exhibits consistently larger transfer errors and reduced explained variance across both reef systems. In contrast to the deep learning models, Random Forest shows limited improvement from median aggregation, with little change in RMSE or $R^2$, suggesting reduced benefit from temporal compositing during cross-regional transfer.

\begin{figure}[H]
\centering
\begin{subfigure}{0.32\textwidth}
\centering
\includegraphics[width=\linewidth]{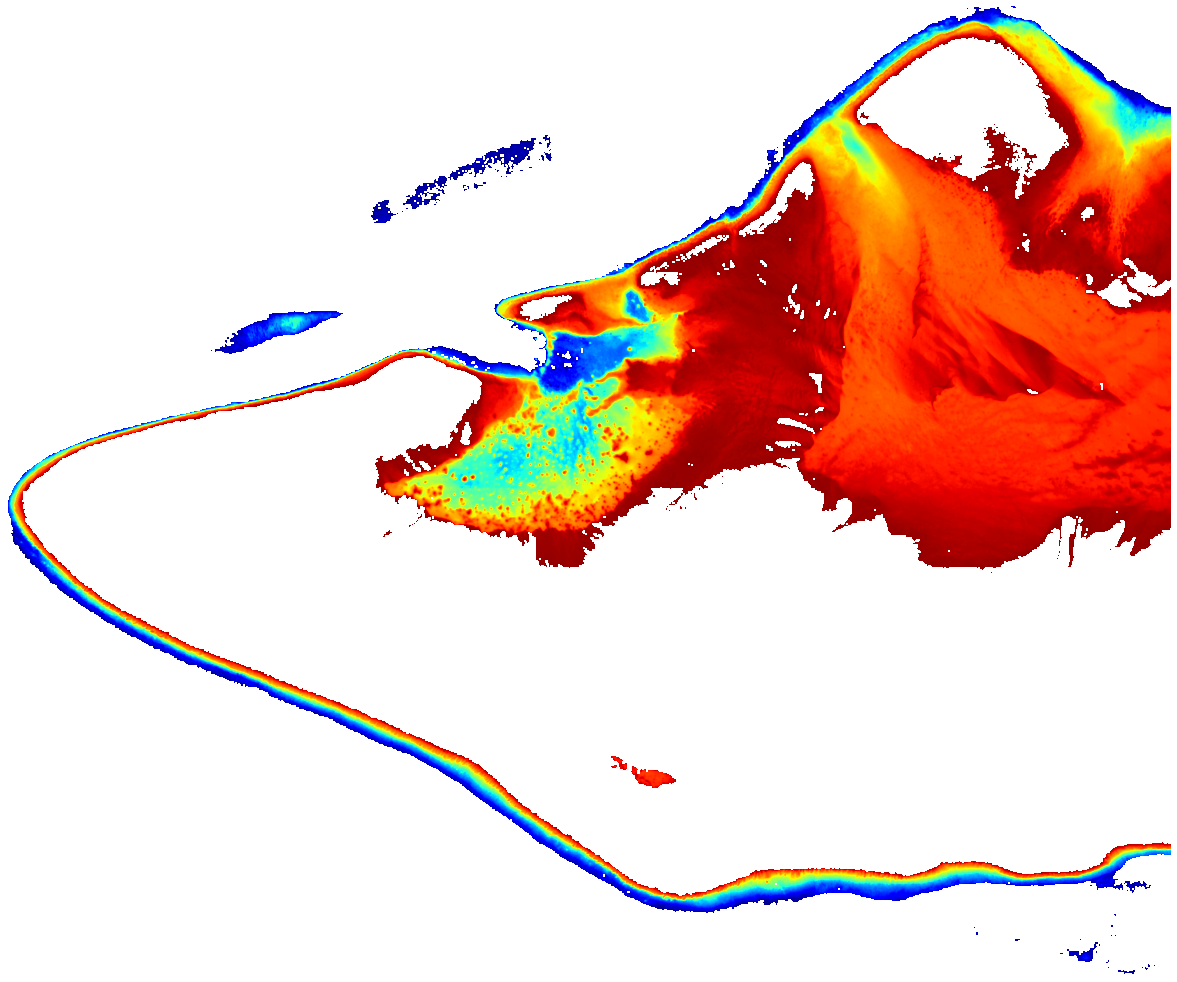}
\caption{\centering GT}
\vspace{6pt}
\end{subfigure}
\hfill
\begin{subfigure}{0.32\textwidth}
\centering
\includegraphics[width=\linewidth]{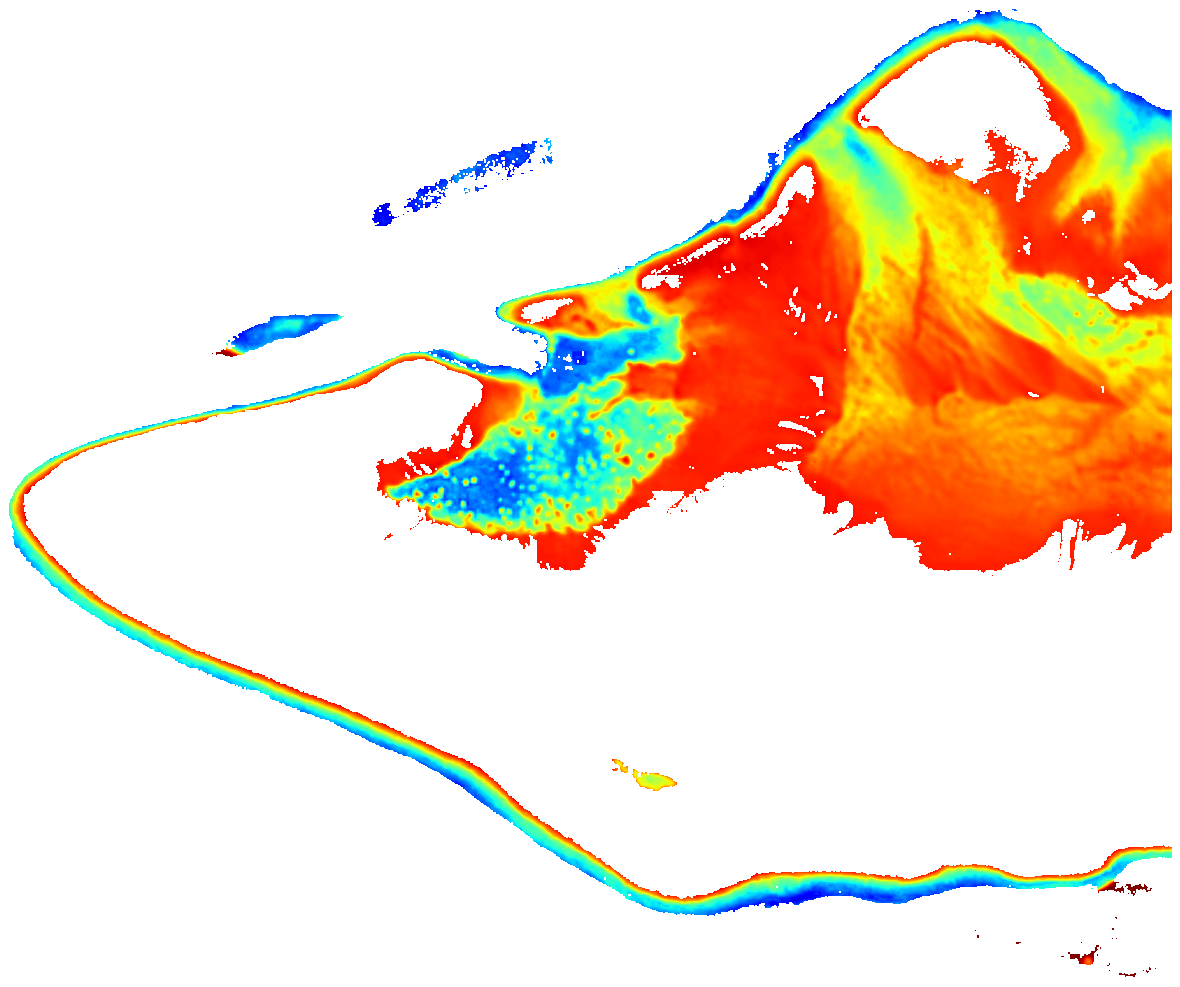}
\caption{\centering ResNet-50 \\ RMSE = 2.88 m}
\vspace{6pt}
\end{subfigure}
\hfill
\begin{subfigure}{0.32\textwidth}
\centering
\includegraphics[width=\linewidth]{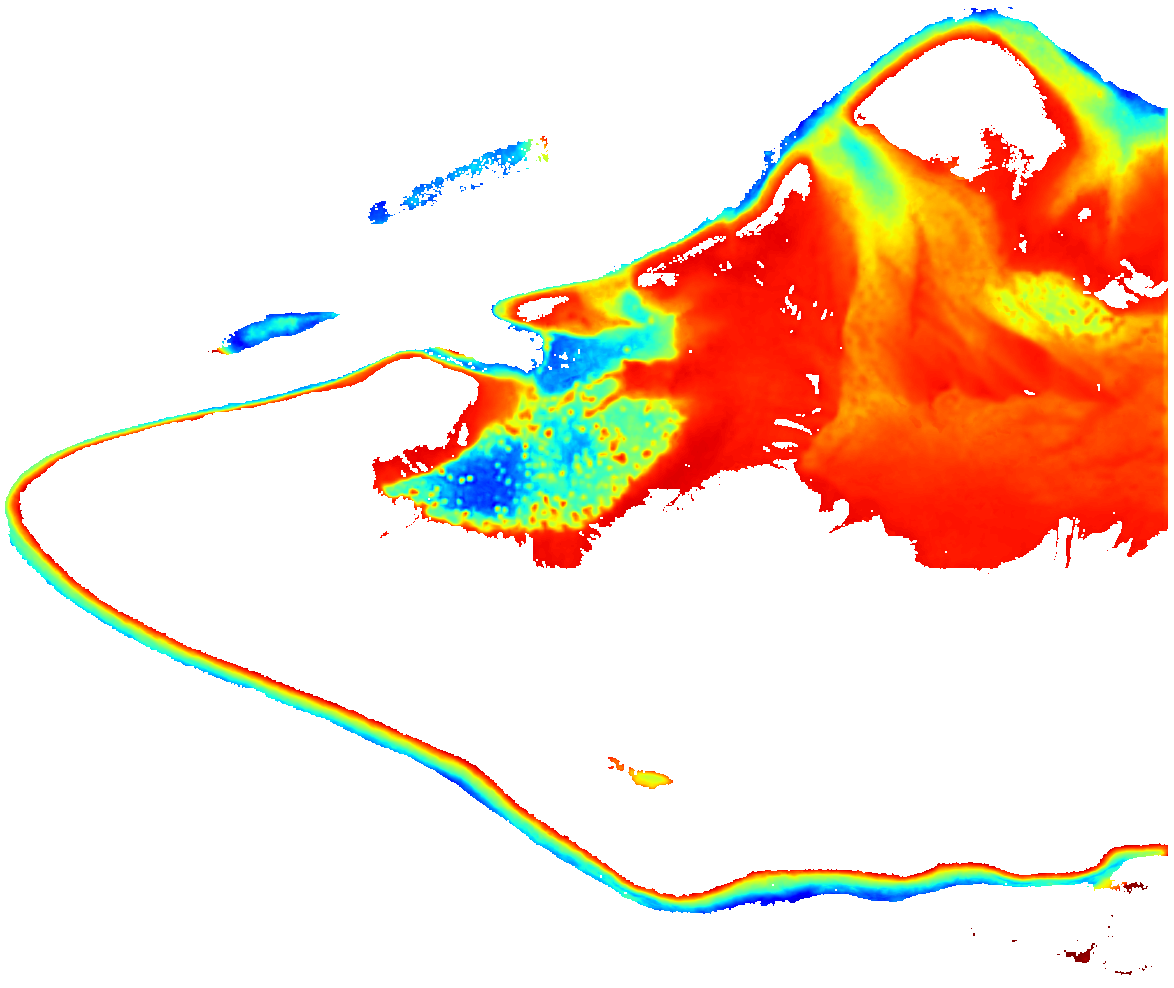}
\caption{\centering ResNet-101 \\ RMSE = 2.56 m}
\end{subfigure}
\vspace{0.3cm}
\begin{subfigure}{0.32\textwidth}
\centering
\includegraphics[width=\linewidth]{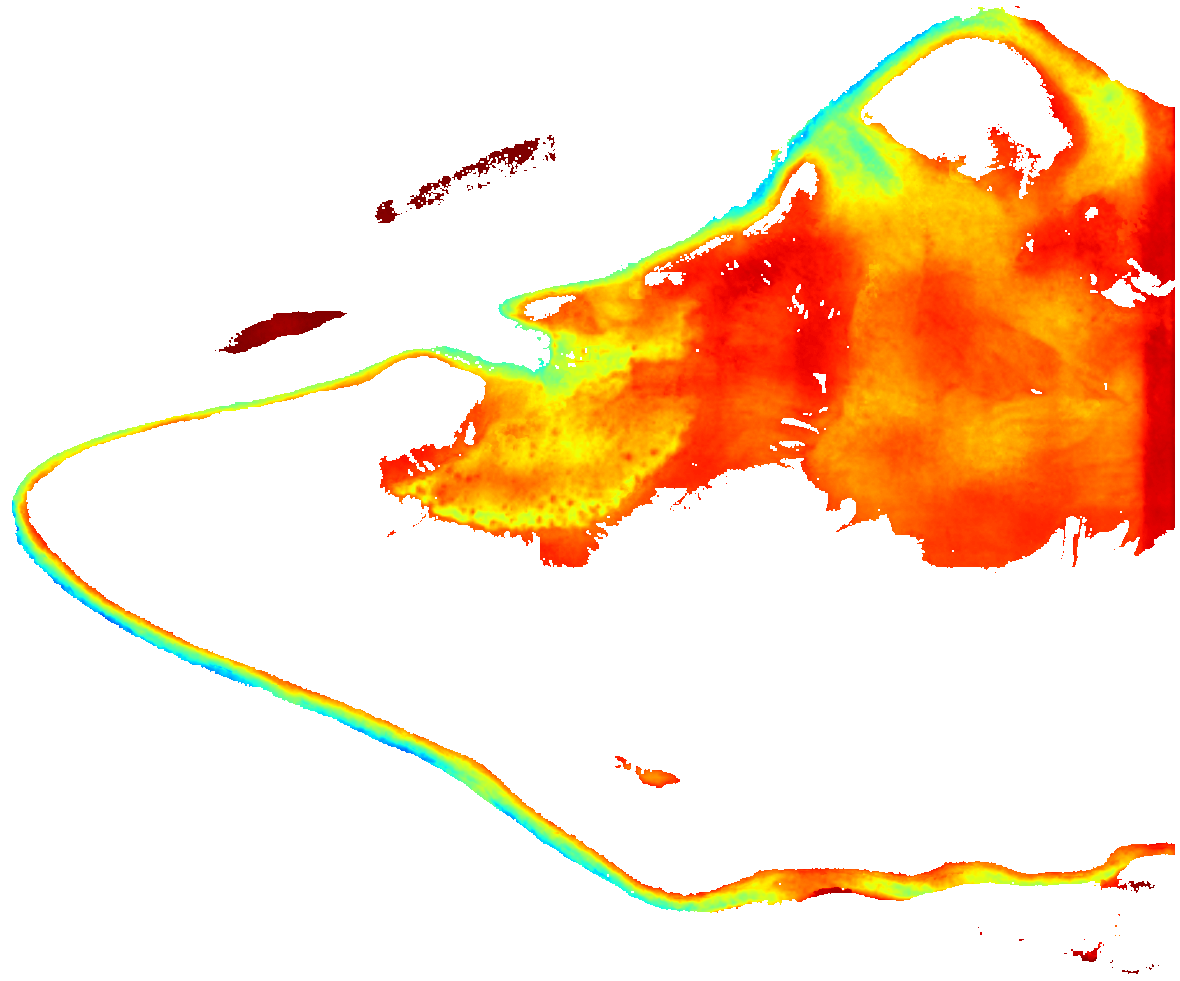}
\caption{\centering EfficientNet-B4 \\ RMSE = 4.49 m}
\end{subfigure}
\hfill
\begin{subfigure}{0.32\textwidth}
\centering
\includegraphics[width=\linewidth]{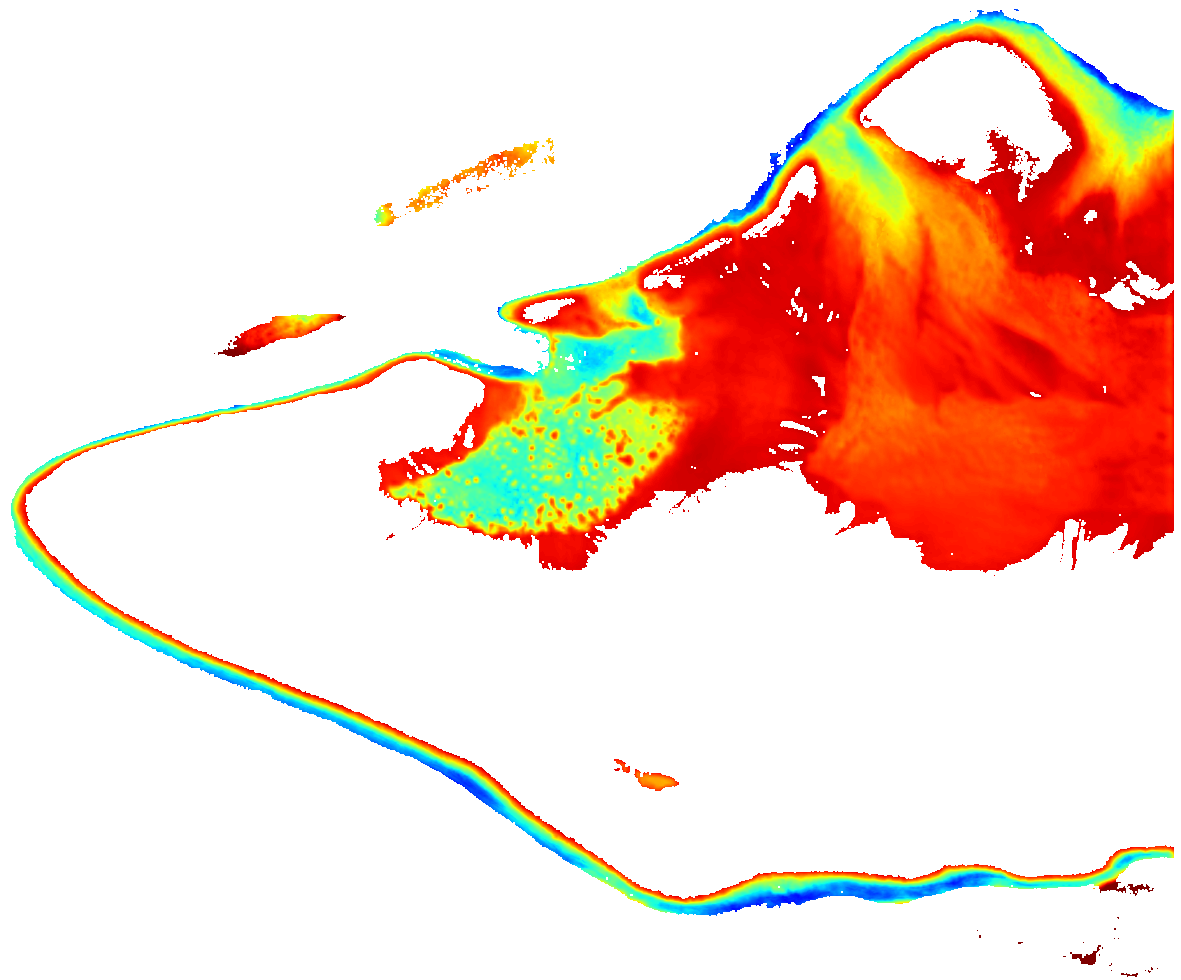}
\caption{\centering ConvNeXt-L \\ RMSE = 2.98 m}
\end{subfigure}
\hfill
\begin{subfigure}{0.32\textwidth}
\centering
\includegraphics[width=\linewidth]{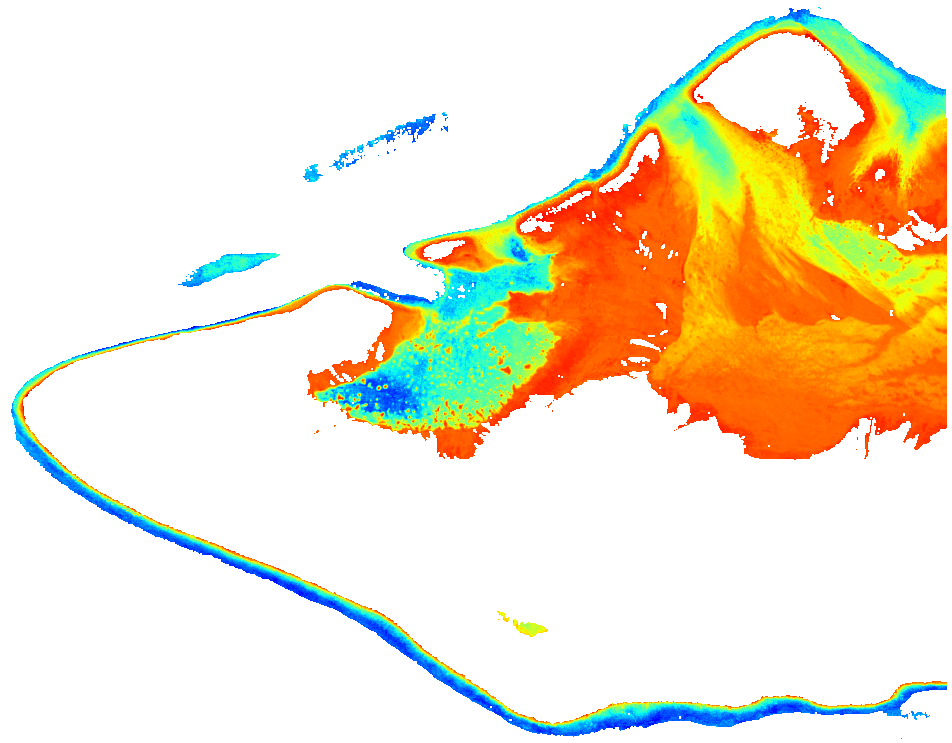}
\caption{\centering Random Forest \\ RMSE = 2.99 m}
\end{subfigure}
\vspace{0.2cm}
\begin{center}
    \includegraphics[width=0.45\linewidth]{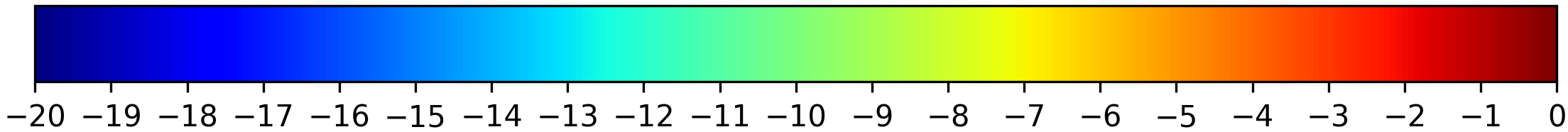}
    \vspace{0.1cm}
    {\small Depth (m)}
\end{center}
\caption{Cross-regional bathymetry predictions for Ashmore Reef. Top row: reference bathymetry and ResNet models. Bottom row: EfficientNet-B4, ConvNeXt-Large, and Random Forest. Reported RMSE values correspond to median aggregation across 10 Sentinel-2 acquisitions.}
\label{fig:ashmore_resultsfor5models}
\end{figure}

Fig.~\ref{fig:ashmore_resultsfor5models} shows cross-regional predictions for Ashmore Reef. ResNet-101 achieves the lowest median RMSE (2.56~m). EfficientNet-B4 exhibits larger errors in the deeper northwest sector; ConvNeXt-Large underestimates detached reef patches; Random Forest overestimates very shallow areas.

\begin{figure}[H]
\centering

\begin{subfigure}{0.32\textwidth}
\centering
\includegraphics[width=\linewidth]{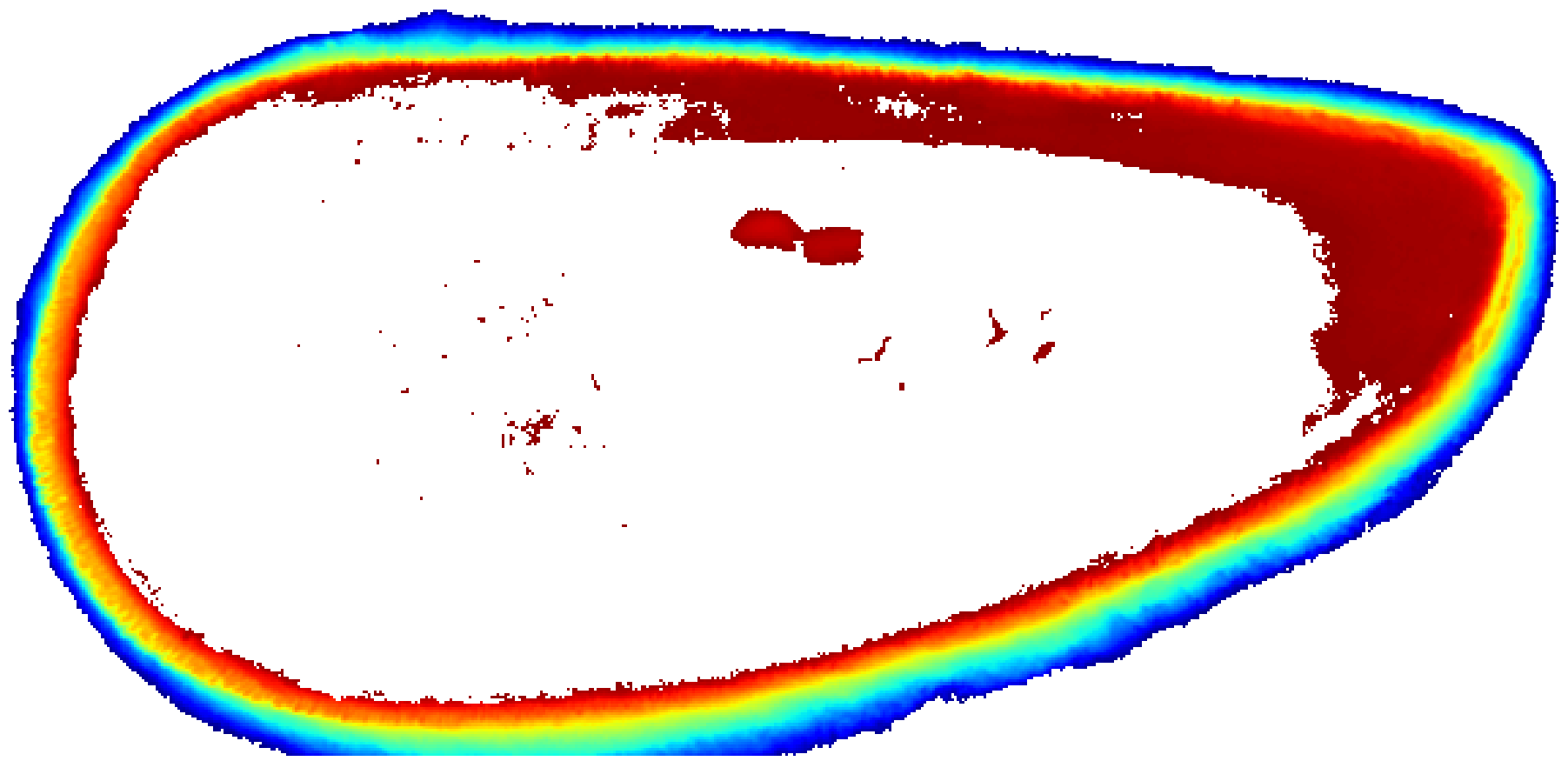}
\caption{\centering GT}
\end{subfigure}
\hfill
\begin{subfigure}{0.32\textwidth}
\centering
\includegraphics[width=\linewidth]{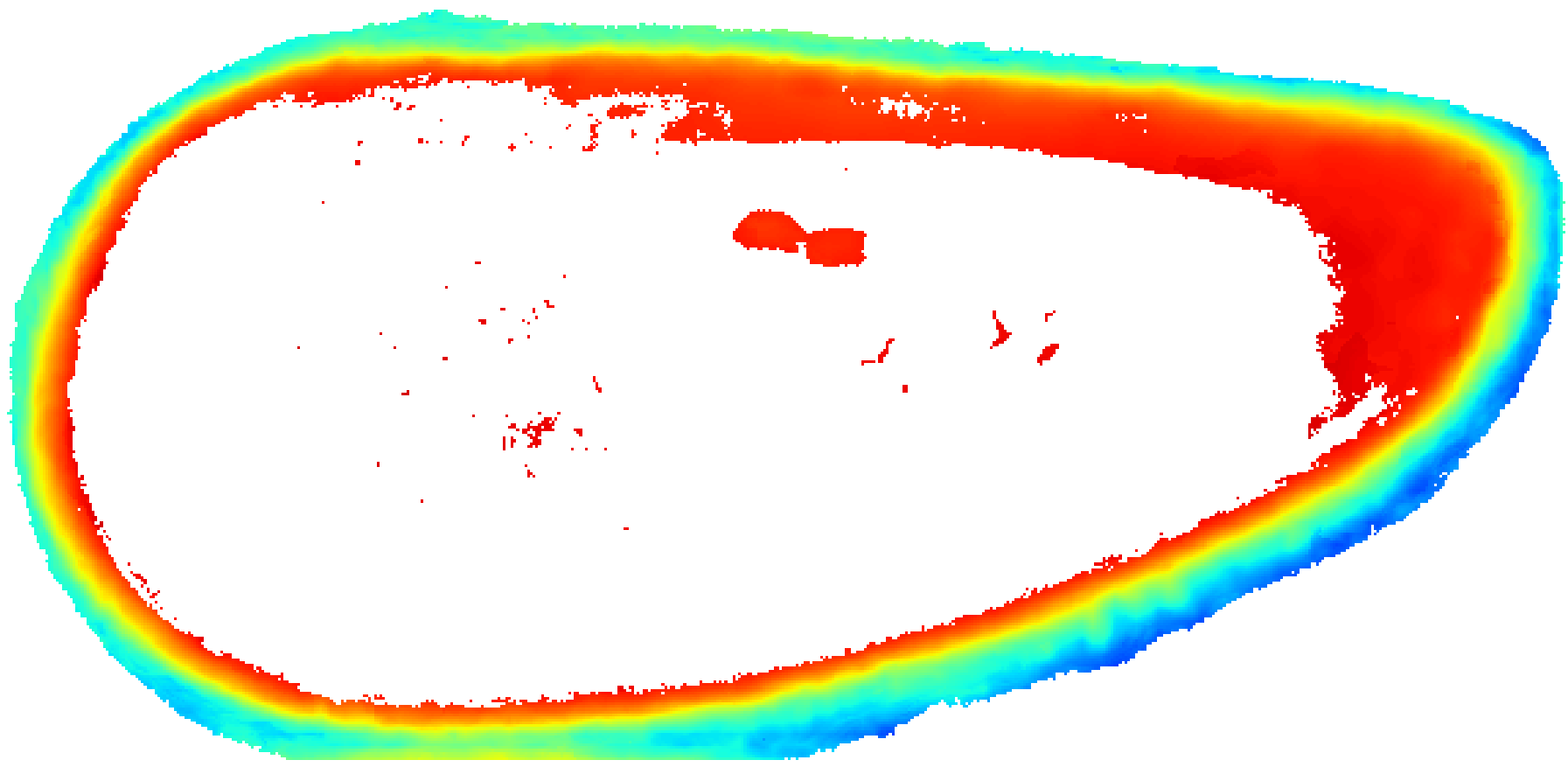}
\caption{\centering ResNet-50 \\ RMSE = 2.59 m}
\end{subfigure}
\hfill
\begin{subfigure}{0.32\textwidth}
\centering
\includegraphics[width=\linewidth]{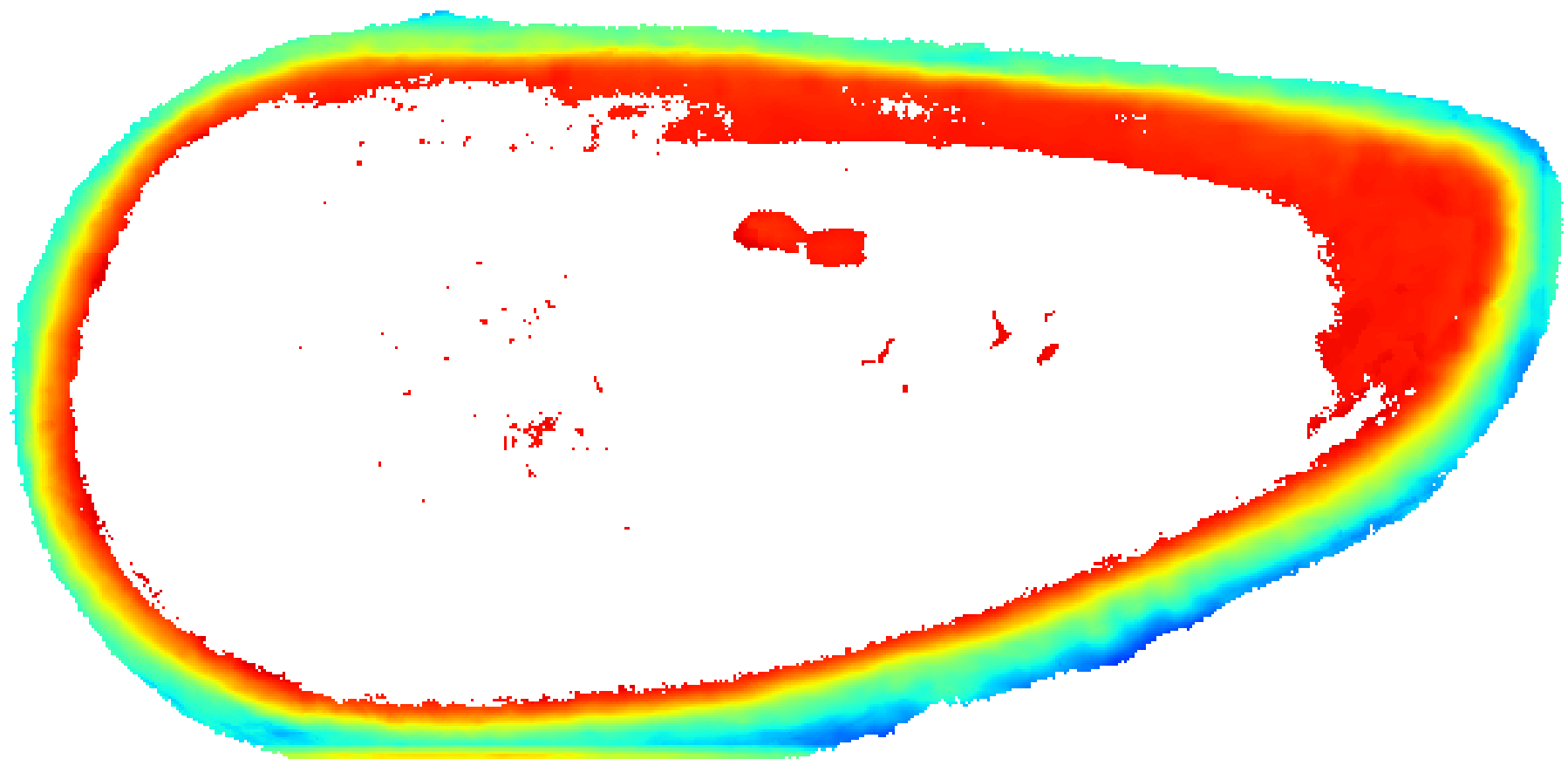}
\caption{\centering ResNet-101 \\ RMSE = 2.85 m}
\end{subfigure}
\vspace{0.3cm}

\begin{subfigure}{0.32\textwidth}
\centering
\includegraphics[width=\linewidth]{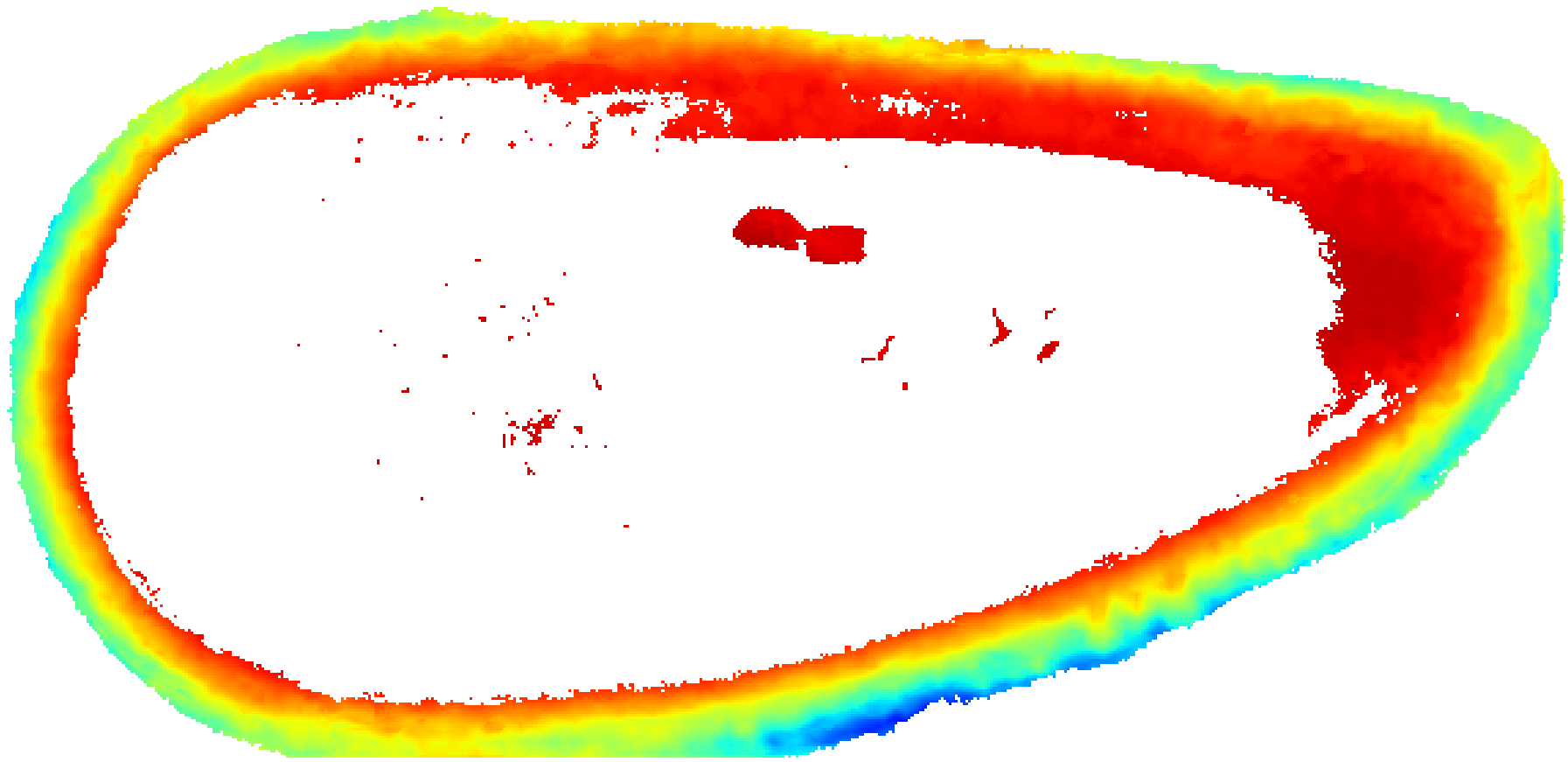}
\caption{\centering EfficientNet-B4 \\ RMSE = 3.78 m}
\end{subfigure}
\hfill
\begin{subfigure}{0.32\textwidth}
\centering
\includegraphics[width=\linewidth]{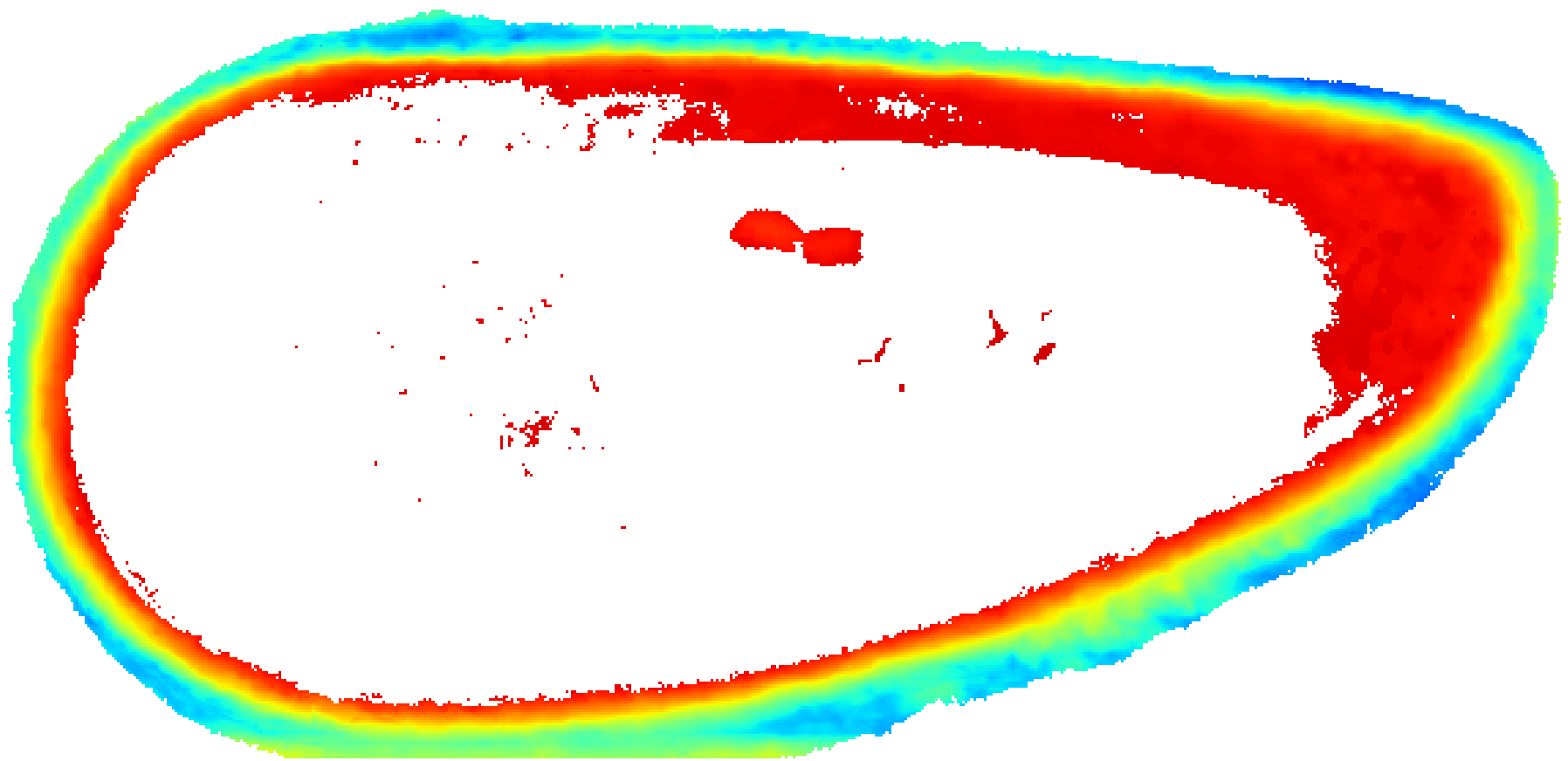}
\caption{\centering ConvNeXt-L \\ RMSE = 2.46 m}
\end{subfigure}
\hfill
\begin{subfigure}{0.32\textwidth}
\centering
\includegraphics[width=\linewidth]{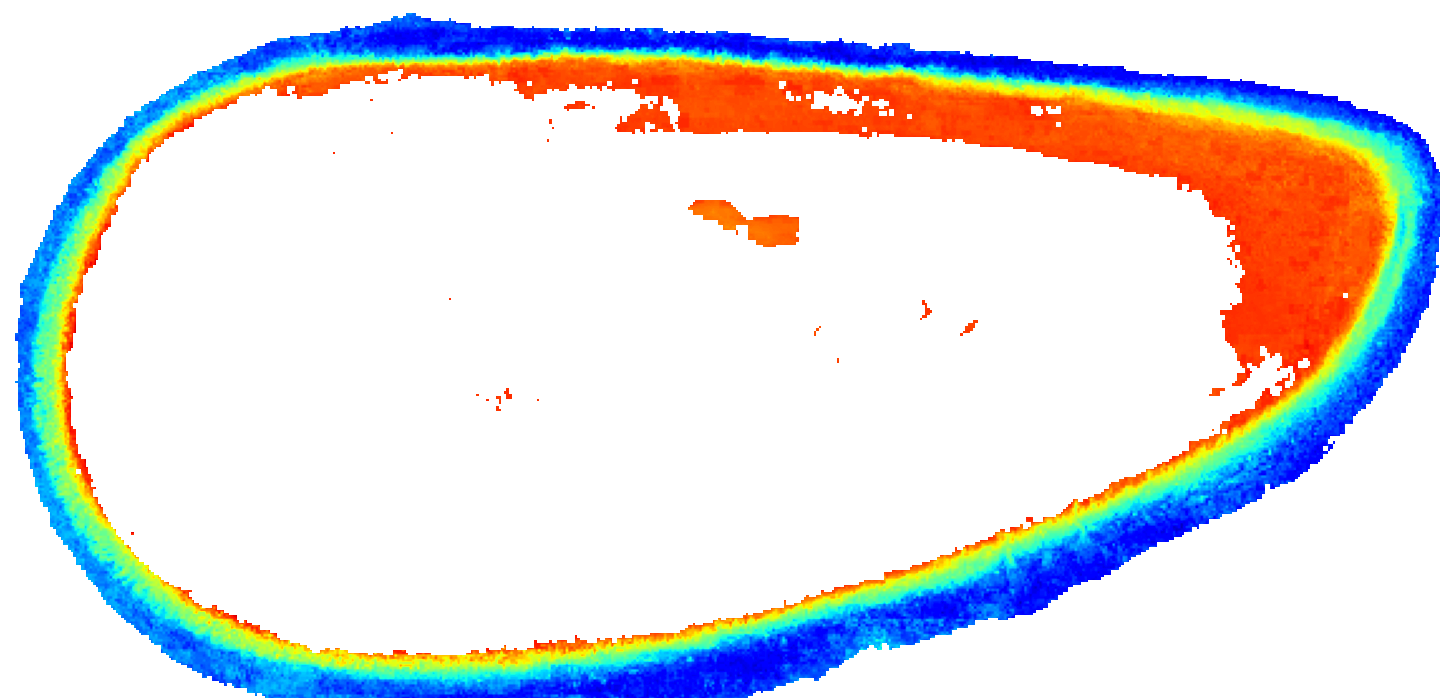}
\caption{\centering Random Forest \\ RMSE = 3.78 m}
\end{subfigure}
\vspace{0.2cm}

\begin{center}
    \includegraphics[width=0.45\linewidth]{Fig_5_colorbar.png}
    
    \vspace{0.1cm}
    {\small Depth (m)}
\end{center}
\caption{Cross-regional bathymetry predictions for Cartier Reef. Top row: reference bathymetry and ResNet models. Bottom row: EfficientNet-B4, ConvNeXt-Large, and Random Forest. Reported RMSE values correspond to median aggregation across 10 Sentinel-2 acquisitions.}
\label{fig:cartier_resultsfor5models}
\end{figure}

Fig.~\ref{fig:cartier_resultsfor5models} presents cross-regional predictions for Cartier Reef. ResNet-50, ResNet-101, and ConvNeXt-Large show similar spatial patterns; EfficientNet-B4 underestimates deeper outer reef areas; Random Forest overestimates both lagoon and outer reef depths. 

\begin{figure}[H]
    \centering
  
    
\begin{subfigure}{1\textwidth}
    \centering
    \includegraphics[width=\linewidth]{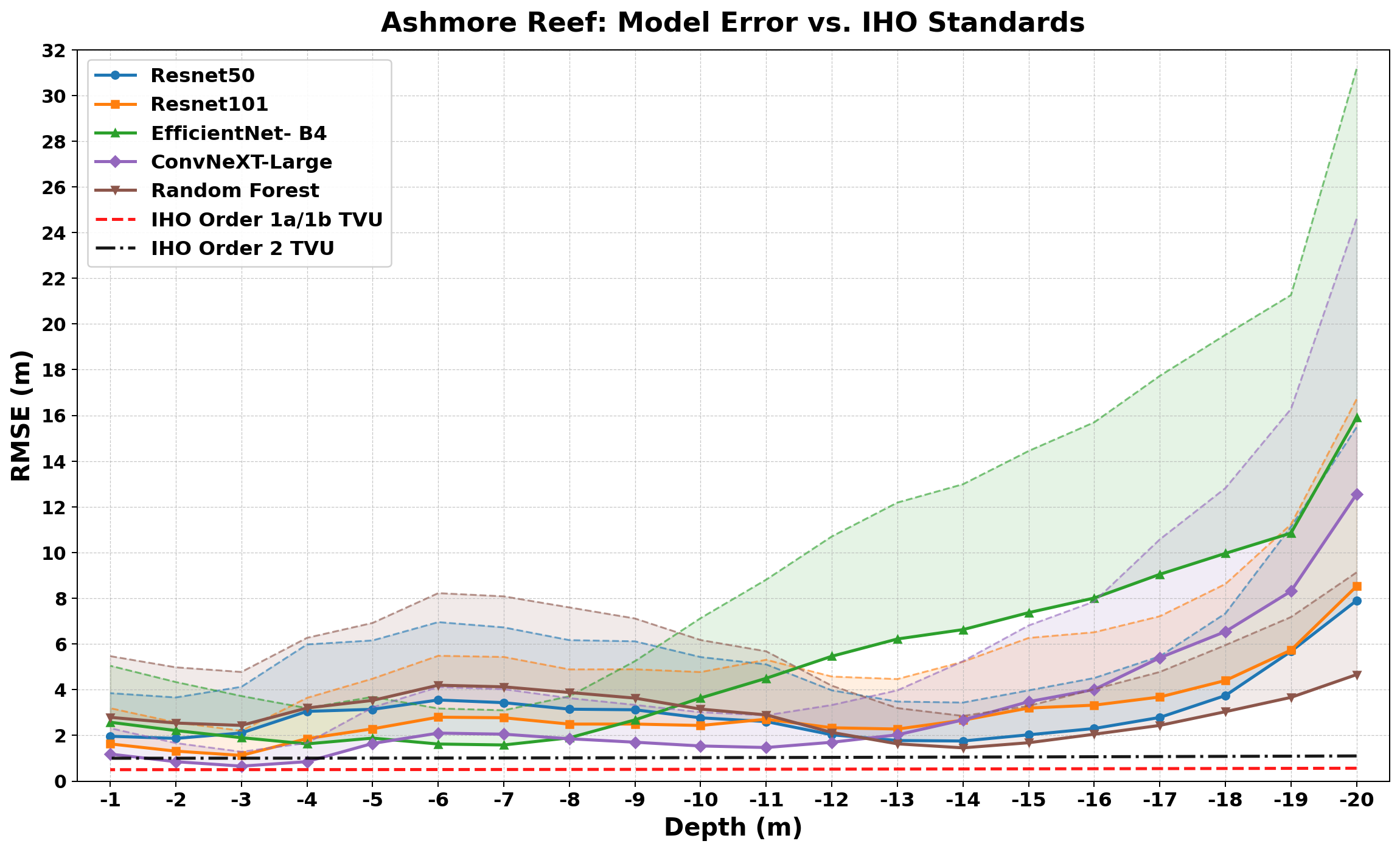}
    \caption{\centering Ashmore Reef}
    \label{fig:ashmore_medianRMSEfor5models_depthwise}
\end{subfigure}
\par\vspace{0.4cm}
\begin{subfigure}{1\textwidth}
    \centering
    \includegraphics[width=\linewidth]{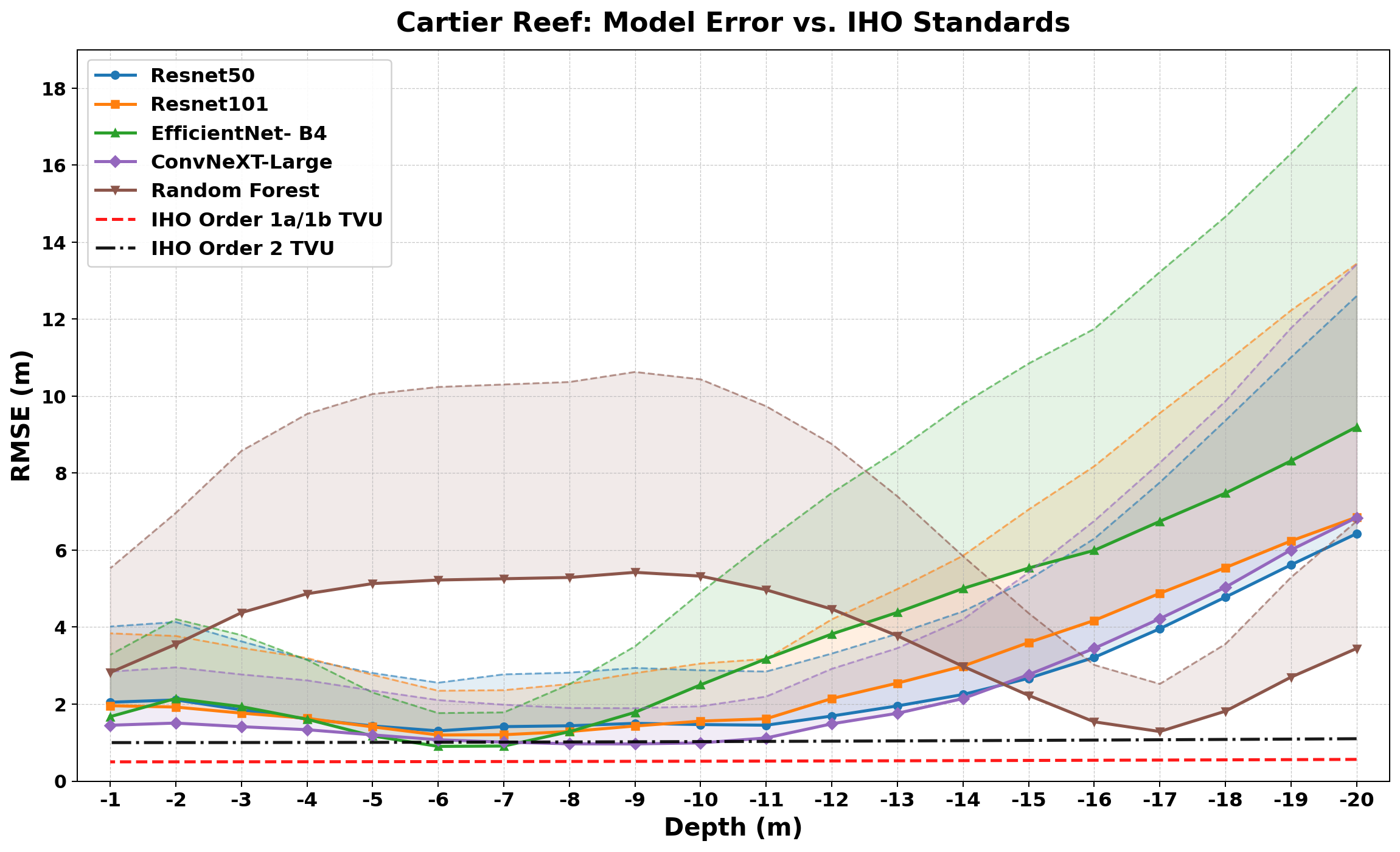}
    \caption{\centering Cartier Reef}
    \label{fig:cartier_medianRMSEfor5models_depthwise}
\end{subfigure} 
    \caption{Depth-dependent median RMSE (solid lines) and 95\% confidence intervals (1.96\,$\times$\,RMSE, dashed lines with shading) for five models across Ashmore and Cartier reefs. ConvNeXt-Large achieves the lowest cross-regional RMSE at intermediate depths. IHO S-44 Order~1 and Order~2 TVU thresholds (defined at 95\% confidence) are shown as a reference scale only, not as a target for Sentinel-2-based SDB.}
    \label{fig:medianRMSE_2reefs}
\end{figure}

Fig.~\ref{fig:medianRMSE_2reefs} shows depth-dependent median RMSE for each model across Ashmore and Cartier. Corresponding 95\% confidence intervals are overlaid, computed as $1.96 \times$RMSE under the assumption that residuals are approximately Gaussian and unbiased. RF produces the highest RMSE throughout, exceeding all other models across both sites. ResNet-50, ResNet-101, and ConvNeXt-Large consistently achieve the lowest cross-regional RMSE, particularly at intermediate depths of 5--11~m. ConvNeXt-Large in particular achieves cross-regional RMSE of approximately 0.97--1.01~m at 7--10~m depth at Cartier Reef and 0.65--0.85~m at 2--4~m depth at Ashmore Reef. The corresponding 95\% confidence intervals at these depths are approximately 1.3--2.0~m, compared with IHO S-44 Order~2 TVU thresholds of approximately 1.00--1.03~m at these depths. The best-performing model is therefore within a factor of 1.3--2 of the formal Order~2 threshold in its strongest depth ranges. As expected for Sentinel-2-based SDB without dedicated hydrographic-grade calibration, no model consistently meets the IHO Order~2 standard at the 95\% confidence level across the full 0--20~m depth range.

To support a more diagnostic interpretation of the cross-regional residual error, Figs.~S7 and~S8 in the Supplementary Information further decompose the depth-binned residuals into separate random and systematic components, $1.96\,\sigma(d)$ and $\text{Bias}(d)$, plotted separately for each cross-regional reef. The decomposition uses the standard identity $\text{RMSE}^2(d) = \text{Bias}^2(d) + \sigma^2(d)$. The 1.96 factor is the Gaussian conversion from one standard deviation to a 95\% confidence interval and applies to the random-error component only; the bias is a fixed depth-dependent offset and is plotted as-is. Three findings emerge.

First, at the shallowest 1~m depth bin, the random-error 95\% confidence interval ($1.96\,\sigma$) of the best-performing deep-learning models is at or below the IHO Order~2 TVU at both reefs: 0.44~m for ConvNeXt-Large and 0.65~m for ResNet-101 at Cartier, and 0.64~m for ResNet-101 and 0.73~m for ConvNeXt-Large at Ashmore. For ConvNeXt-Large at Cartier, this 0.44~m value is in fact also below the stricter IHO Order~1a/1b TVU of 0.50~m. This indicates that, at the shallowest depths, the random-error component of cross-regional Sentinel-2 SDB from off-the-shelf deep-learning models is in principle compatible with hydrographic-grade tolerances; the cross-regional RMSE at these depths is dominated by the systematic depth-dependent bias rather than by random scatter.

Second, the bias panels of Figs.~S7 and~S8 reveal a clear depth-dependent regression-to-mean signature at both reefs: predictions are systematically too deep at shallow depths (bias $\approx -1$ to $-2$~m) and too shallow at deeper depths (bias $\approx +5$ to $+12$~m at 20~m), with a zero-crossing near 10--12~m for ConvNeXt-Large. This pattern is consistent with a regression model that has no built-in functional prior on how reflectance attenuates with depth and is therefore biased toward the marginal depth distribution of its training set as the bottom-reflected signal becomes weaker with depth.

Third, when pooled across both cross-regional reefs (weighted by pixel count using the law of total variance), the depth-binned $1.96\,\sigma$ of ConvNeXt-Large is within a factor of approximately two of the IHO Order~2 TVU continuously from 1~m to 12~m of depth. Beyond approximately 12~m, both the bias and the random component grow more rapidly with depth: the pooled $1.96\,\sigma$ rises from $\approx 2$~m at 11--12~m depth to $\approx 3$~m at 15~m and exceeds 5~m at 19--20~m, and the bias becomes the dominant contributor to the total cross-regional RMSE.

We emphasise that the deeper-water deterioration is not a fundamental physical limit of Sentinel-2 retrievals --- in exceptionally clear waters, Sentinel-2 reflectance carries depth information well beyond 20~m \cite{Parrish2019} (Supplementary Section~S3), and the operational 0--20~m range adopted in this study is conservative rather than radiometric. The limitation observed here is instead specific to purely data-driven regression in this depth regime: as the bottom-reflected component of the at-sensor reflectance decreases with the exponential attenuation of visible light in the water column, the network --- which has no built-in functional form for this attenuation and must learn it from training pairs that are progressively sparser at depth --- falls back on the marginal depth distribution of its training set. The physics-informed neural network architecture outlined in the Future Work section is intended to address this limitation in several complementary ways, including by encoding a Beer-Lambert-style attenuation prior in the forward model, by introducing dedicated output heads for bottom-substrate albedo and water-column attenuation, and by ingesting multi-temporal Sentinel-2 stacks directly rather than via post-hoc median aggregation. The extent to which these elements can extract additional information from the lower signal-to-noise deep-water regime remains an empirical question that this parallel work is designed to address.

\begin{table}[H]
\small
\caption{Cross-regional prediction performance for Ashmore and Cartier reefs using five models. Results are aggregated over 10 Sentinel-2 acquisitions per reef. Metrics include RMSE (m), MAE (m), and R\textsuperscript{2} under both mean and median temporal aggregation.}
\label{tab:transfer_metrics}

\noindent\textbf{(a) Ashmore Reef}\\[2pt]
\resizebox{\textwidth}{!}{%
\begin{tabular}{lcccccc}
\hline
& \multicolumn{3}{c}{Mean Aggregation} & \multicolumn{3}{c}{Median Aggregation} \\
\cmidrule(lr){2-4} \cmidrule(lr){5-7}
Model
& RMSE (m) & MAE (m) & R\textsuperscript{2}
& RMSE (m) & MAE (m) & R\textsuperscript{2} \\
\hline
ResNet50        & 3.16 & 2.59 & 0.64   & 2.88 & 2.36 & 0.71   \\
ResNet101       & 2.86 & 2.06 & 0.71   & 2.56 & 1.85 & 0.77  \\
EfficientNet-B4 & 4.79 & 3.31 & 0.18   & 4.49 & 3.02 & 0.30   \\
ConvNeXt-Large  & 3.22 & 1.87 & 0.63   & 2.98 & 1.56 & 0.69   \\
Random Forest   & 2.93 & 2.67 & 0.70   & 2.99 & 2.77 & 0.69   \\
\hline
\end{tabular}
}

\vspace{8pt}

\noindent\textbf{(b) Cartier Reef}\\[2pt]
\resizebox{\textwidth}{!}{%
\begin{tabular}{lcccccc}
\hline
& \multicolumn{3}{c}{Mean Aggregation} & \multicolumn{3}{c}{Median Aggregation} \\
\cmidrule(lr){2-4} \cmidrule(lr){5-7}
Model
& RMSE (m) & MAE (m) & R\textsuperscript{2}
& RMSE (m) & MAE (m) & R\textsuperscript{2} \\
\hline
ResNet50        & 3.01 & 2.39 & 0.77   & 2.59 & 2.13 & 0.83   \\
ResNet101       & 3.00 & 2.36 & 0.77   & 2.85 & 2.29 & 0.79   \\
EfficientNet-B4 & 4.53 & 3.34 & 0.47   & 3.78 & 2.81 & 0.64   \\
ConvNeXt-Large  & 2.87 & 2.17 & 0.79   & 2.46 & 1.87 & 0.85  \\
Random Forest   & 3.60 & 3.47 & 0.68   & 3.78 & 3.44 & 0.65   \\
\hline
\end{tabular}
}

\end{table}

\vspace{-6pt}

\vspace{-6pt}

\subsubsection{Comparison with Reference Architectures}


The proposed models were benchmarked against Swin-BathyUNet and a baseline U-Net \cite{Agrafiotis2025} using the public MagicBathyNet dataset ($720 \times 720$ patches at 25~cm resolution). Agia Napa (22 patches, 0--16~m) and Puck Lagoon ($\sim$1575 patches, 0--6~m) were combined into a unified 0--16~m benchmark for training and evaluation, enabling more robust learning across heterogeneous shallow-water conditions. NaN and land pixels were masked during training and evaluation.

For the MagicBathyNet benchmark, all proposed models were trained within a DeepLabV3+ framework using the corresponding encoder backbones, Adam optimizer, learning rate of $1\times10^{-4}$ with ReduceLROnPlateau scheduling (factor = 0.5, patience = 5), batch size of 16, and a maximum of 100 epochs with early stopping (patience = 10). RMSE loss was adopted to align with Swin-BathyUNet, which also uses RMSE-based training. All models used RGB aerial imagery inputs consistent with the MagicBathyNet dataset modality. The Swin-BathyUNet and baseline U-Net results were taken directly from the original MagicBathyNet publication and therefore reflect the training configurations reported in that study rather than a fully reimplemented benchmark.


As shown in Table~\ref{tab:dl_comparison}, the proposed DeepLabV3+ models trained on the combined dataset under the MagicBathyNet benchmark achieve lower RMSE than the site-specific Swin-BathyUNet and baseline U-Net results reported in \cite{Agrafiotis2025}, while using substantially fewer parameters.

%

\begin{table}[H]
\caption{Comparison of RMSE, MAE, and model complexity on the MagicBathyNet shallow-water benchmark (0--16 m). 
Our models were trained on the combined Puck Lagoon and Agia Napa datasets. 
Reported Swin-BathyUNet and baseline U-Net results are site-specific models trained separately for each location \cite{Agrafiotis2025}.}
\label{tab:dl_comparison}
\resizebox{\textwidth}{!}{%
\begin{tabular}{lccc c}
\hline
\textbf{Model} & \textbf{Training Setup} & \textbf{RMSE (m)} & \textbf{MAE (m)} & \textbf{Parameters (M)} \\
\hline

ResNet-50        & Unified (0--16 m) & 0.21 & 0.12 & 39.7 \\
ResNet-101       & Unified (0--16 m) & \textbf{0.19} & \textbf{0.12} & 58.7 \\
EfficientNet-B4  & Unified (0--16 m) & 0.22 & 0.13 & 18.6 \\
ConvNeXt-Large   & Unified (0--16 m) & 0.21 & 0.13 & 198.8 \\

\hline
Swin-BathyUNet\textsuperscript{a}   & Puck (0--6 m)     & 0.24 & 0.17 & 395 \\
Swin-BathyUNet\textsuperscript{a}   & Agia Napa (0--16 m) & 0.53 & 0.40 & 395 \\
Baseline U-Net\textsuperscript{a}   & Puck (0--6 m)     & 0.22 & 0.15 & 31 \\
Baseline U-Net\textsuperscript{a}   & Agia Napa (0--16 m) & 0.67 & 0.53 & 31 \\
\hline
\end{tabular}
}
\begin{tablenotes}
\footnotesize
\item \textsuperscript{a}Reported from \cite{Agrafiotis2025}.
\end{tablenotes}
\end{table}

In contrast, Swin-BathyUNet reports RMSE of 0.24~m (Puck Lagoon) and 0.53~m (Agia Napa), while the baseline U-Net reports 0.22~m and 0.67~m, respectively \cite{Agrafiotis2025}. Notably, all proposed architectures achieve lower RMSE than Swin-BathyUNet while using substantially fewer parameters (e.g., 58.7~M vs.\ 395~M, approximately 85\% reduction).

These results establish a controlled shallow-water benchmark under consistent preprocessing and evaluation conditions, highlighting favorable accuracy--efficiency trade-offs for modern convolutional backbones relative to larger transformer-based architectures. Importantly, this benchmark is based on high-resolution aerial RGB imagery and should not be interpreted as representing operational Sentinel-2 SDB performance; the two experimental settings differ in sensor modality, spatial resolution, preprocessing, and geographic scope.

\subsection{Sensitivity Analyses}
\label{subsec:sensitivity}

We conduct sensitivity analyses focusing on two key aspects: loss function choice and spatial organization of training data, both of which directly influence shallow-water accuracy and mid-depth stability over the 0--20~m range. Unless stated otherwise, bathymetric depth is reported as a positive quantity (m) increasing downward.

\subsubsection{Loss Function Comparison}
\label{subsubsec:loss_func}

We compare three loss functions-standard RMSE, relative percentage error (RPE), and the proposed Smooth Weight Function (SWF)-weighted RMSE-using ResNet101 and ConvNeXt-Large as representative architectures. Models are trained on Pratas Island and selected Great Barrier Reef (GBR) sites and evaluated cross-regionally on Ashmore and Cartier reefs over the 0--20~m depth range (depth-dependent behavior within the Pratas intra-regional test set is discussed in Supplementary Section S3).

During initial experiments with the RPE loss, near-zero depth values produced excessively large gradients, biasing optimization toward shallow predictions. Such values may arise from interpolation, resampling, and tide-correction artifacts during preprocessing. To improve numerical stability and ensure fair comparison, pixels shallower than 1~cm were therefore excluded from the analysis.

Table~\ref{tab:sensitivity_loss} shows that SWF consistently achieves the lowest shallow-water RMSE ($\leq$3~m), while RPE often yields the lowest full-depth RMSE (0--20~m). ConvNeXt-Large is more sensitive to loss-function choice than ResNet101.

\begin{table}[H]
\caption{Sensitivity of cross-regional performance to different loss functions for ResNet101 and ConvNeXt-Large on Ashmore and Cartier reefs. RMSE is reported for the full 0--20~m range and shallow water ($\leq$3~m).}
\label{tab:sensitivity_loss}
\resizebox{0.95\linewidth}{!}{%
\begin{tabular}{llcccccc}
\hline
Model & Loss & \makecell{Ashmore \\ RMSE (0--20 m)} & \makecell{Ashmore \\ RMSE ($\leq$ 3 m)} & \makecell{Cartier \\ RMSE (0--20 m)} & \makecell{Cartier \\ RMSE ($\leq$ 3 m)} \\
\hline
ResNet101 & SWF  & 2.56 & 1.35 & 2.85 & 1.88 \\
ResNet101 & RMSE & 2.71 & 1.53 & 2.51 & 1.92 \\
ResNet101 & RPE (Z $>0.01$~m)  &  2.27 & 1.54 &  2.53 & 1.73 \\
\hline
ConvNeXt-L & SWF  & 2.98   & 0.90   & 2.46   & 1.46 \\
ConvNeXt-L & RMSE & 3.50   & 1.27   & 2.42   & 1.88   \\
ConvNeXt-L & RPE (Z $>0.01$~m)  & 2.00   &  0.96  & 2.35   & 1.92   \\
\hline
\end{tabular}}
\end{table}

\begin{figure}[H] 
    \centering
    \includegraphics[width=1\textwidth]{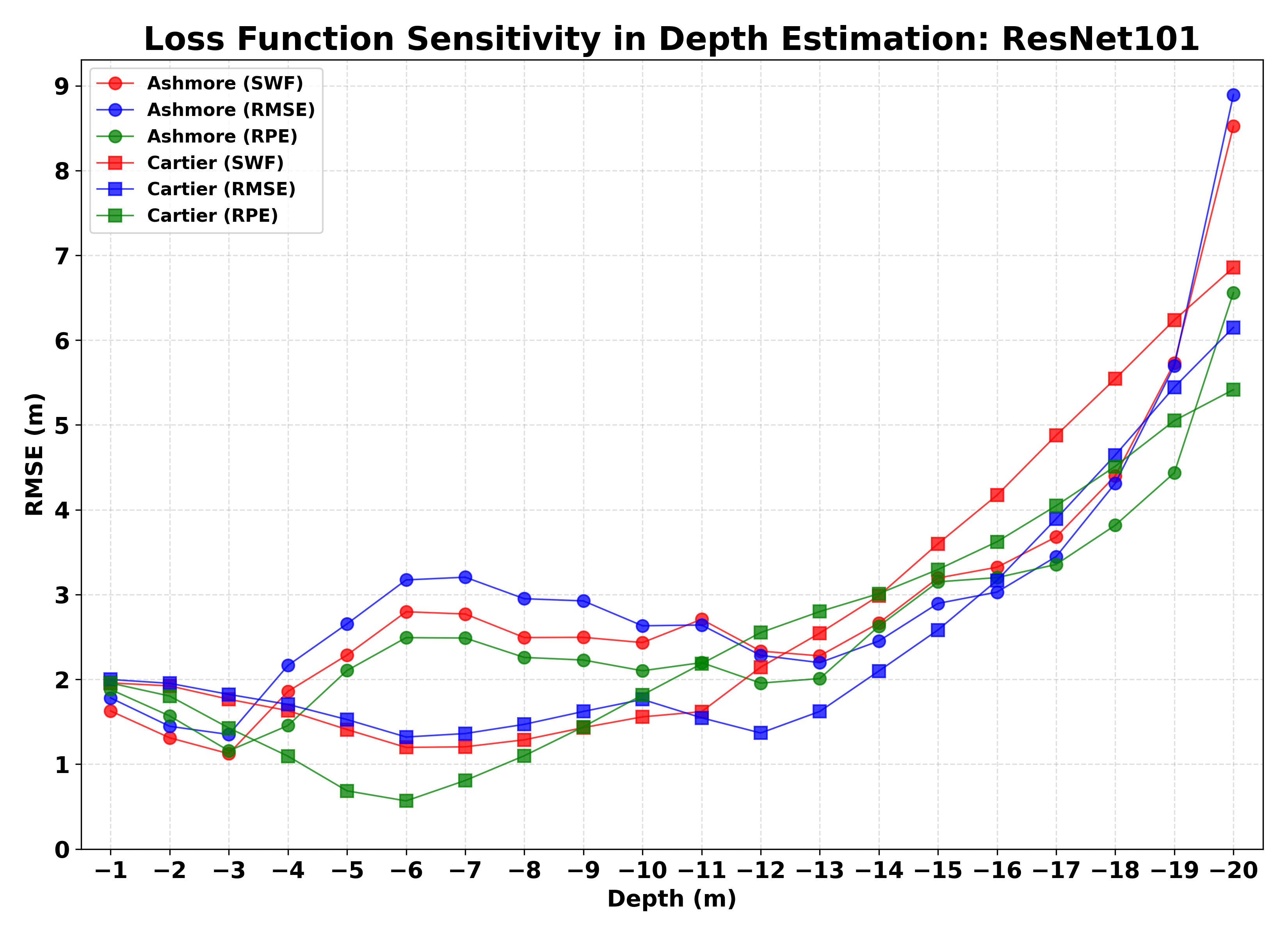}
    \caption{Depth-dependent RMSE for ResNet101 under SWF, RMSE, and RPE losses at Ashmore and Cartier reefs. SWF emphasizes shallow-water retrieval, while RPE normalizes errors by target depth.}
    \label{fig:RMSE_RPE_SWF_ResNet101}
\end{figure}

\begin{figure}[H] 
    \centering
    \includegraphics[width=1\textwidth]{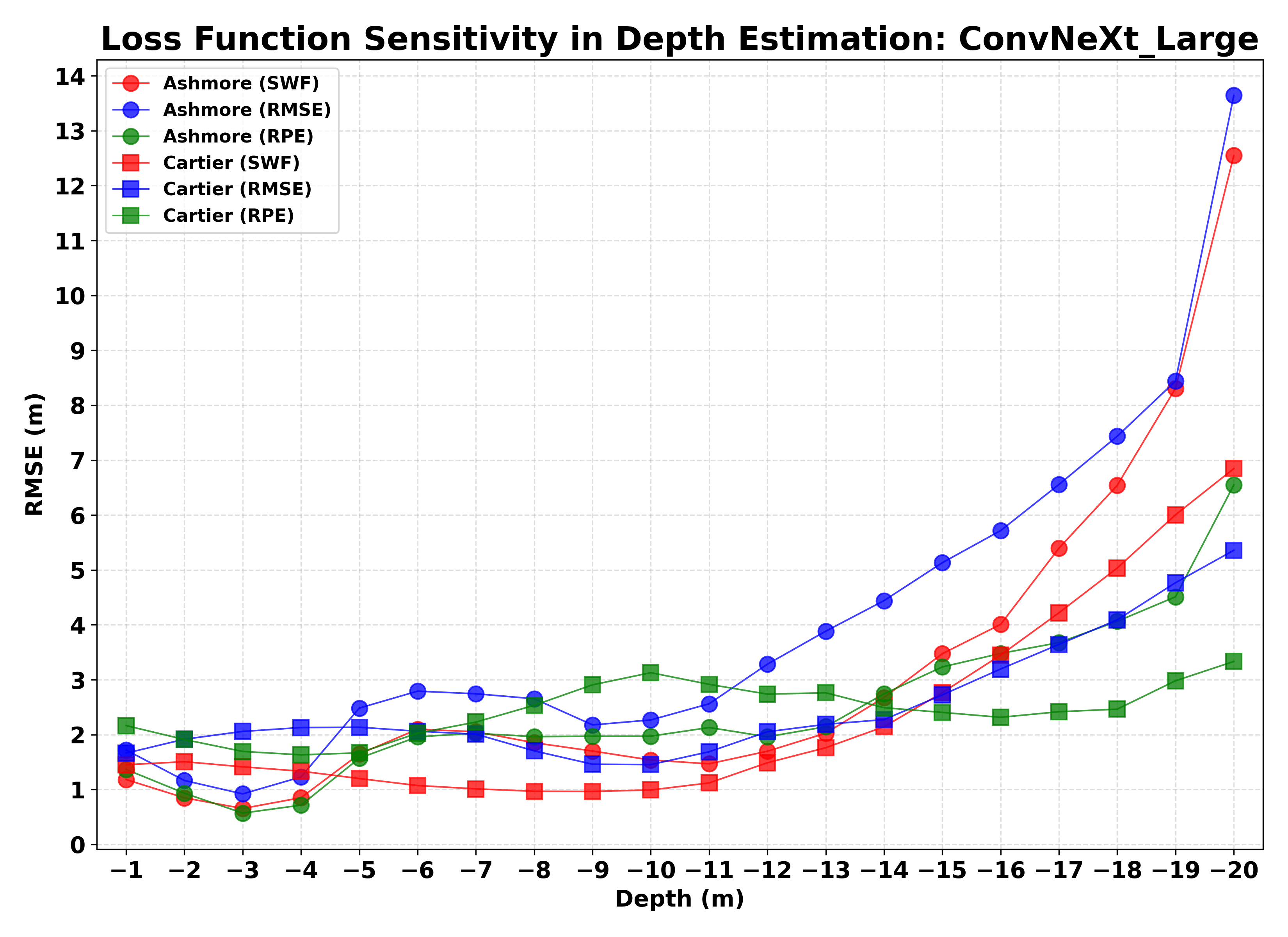}
    \caption{Depth-dependent RMSE for ConvNeXt-Large under SWF, RMSE, and RPE losses. SWF demonstrates sustained shallow-water gains up to $\sim$14~m, with RPE outperforming in deeper zones.}
    \label{fig:RMSE_RPE_SWF_ConvNeXt_Large}
\end{figure}

\begin{figure}[H]
    \centering
    \begin{subfigure}[t]{0.49\textwidth}
        \centering
        \includegraphics[width=\textwidth]{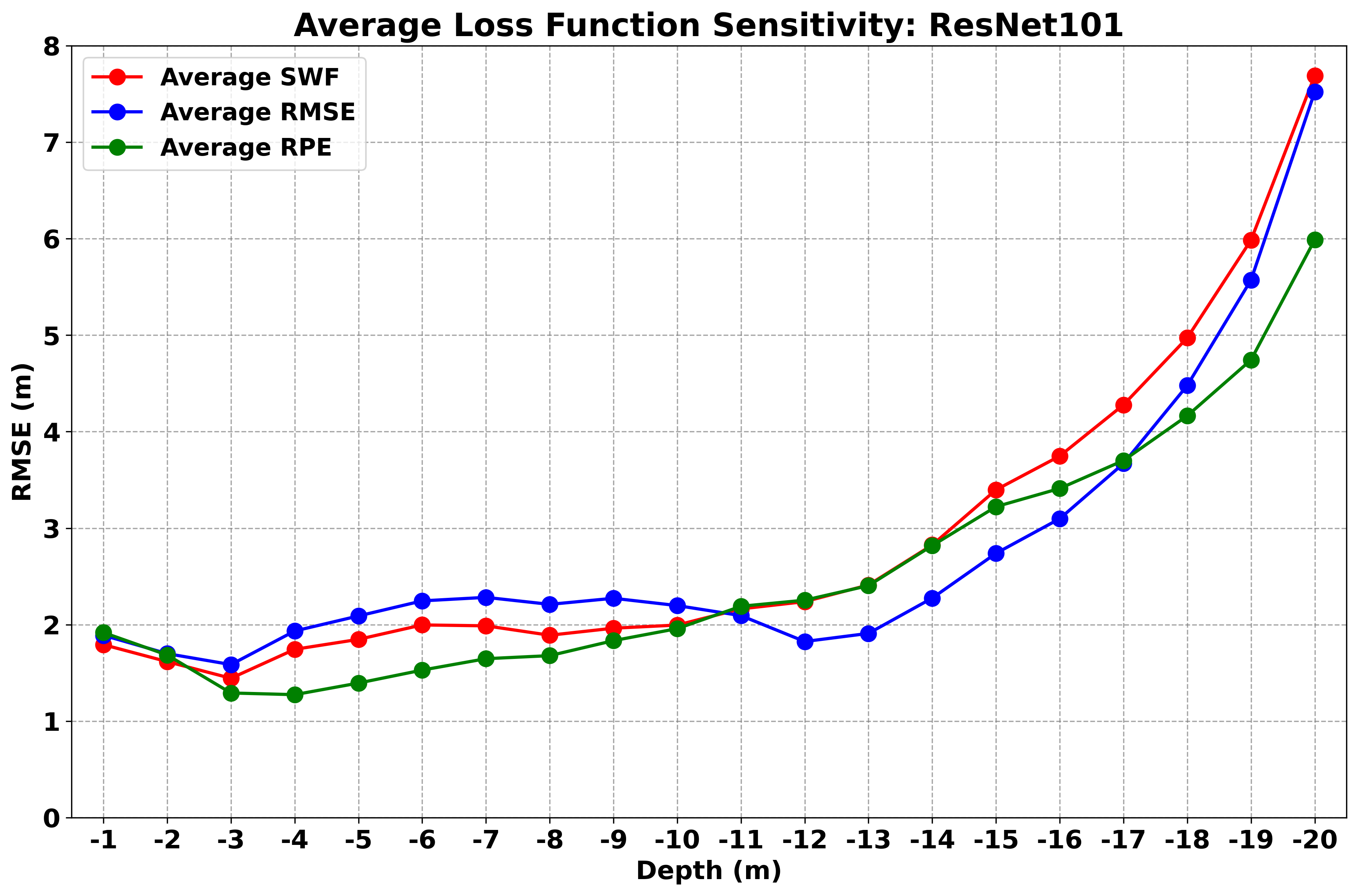}
        \caption{\centering ResNet101}
        \label{fig:Avg_ResNet101}
    \end{subfigure}
    \hfill
    \begin{subfigure}[t]{0.49\textwidth}
        \centering
        \includegraphics[width=\textwidth]{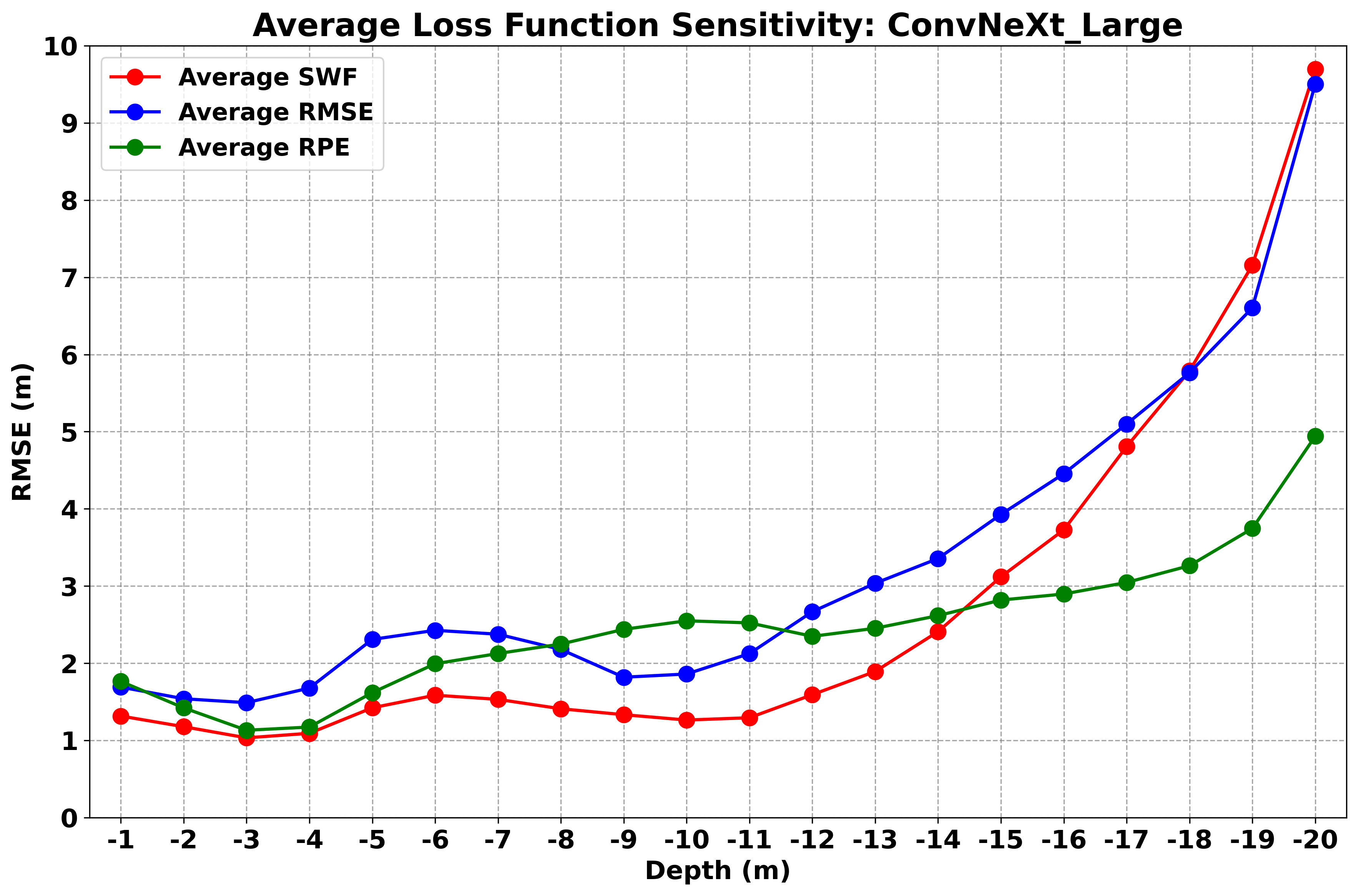}
        \caption{\centering ConvNeXt-Large}
        \label{fig:Avg_ConvNeXt}
    \end{subfigure}
    \caption{Depth-binned RMSE averaged across Ashmore and Cartier reefs under SWF, RMSE, and RPE losses for (a) ResNet101 and (b) ConvNeXt-Large.}
    \label{fig:Avg_models}
\end{figure}

Figs.~\ref{fig:RMSE_RPE_SWF_ResNet101} and \ref{fig:RMSE_RPE_SWF_ConvNeXt_Large} show depth-dependent RMSE for ResNet101 and ConvNeXt-Large. For ResNet101, differences between loss functions are small across both sites, with no single loss consistently dominant across the full depth range. For ConvNeXt-Large, SWF produces a clearer advantage: at Cartier, SWF achieves the lowest RMSE from the surface down to approximately 14~m, while RPE outperforms at greater depths (15--20~m); at Ashmore, SWF leads at shallow and intermediate ranges, with RPE performing better at deeper bins. Depth-binned RMSE averaged across reefs (Fig.~\ref{fig:Avg_models}) confirms that ResNet101 is relatively insensitive to loss choice, while ConvNeXt-Large benefits most from SWF across shallow-to-intermediate depths.

\vspace{-6pt}

\vspace{-6pt}

To identify which model-loss combination performs most consistently across regions, depth-binned RMSE values were averaged across Ashmore and Cartier reefs (Fig.~\ref{fig:Avg_models}).
\vspace{-6pt}

\subsubsection{SWF Hyperparameter Sensitivity}
\label{subsubsec:swf_hyperparam_sensitivity}

To examine how the SWF hyperparameters $\beta$ and $Z_0$ redistribute prediction accuracy across depth regimes, we conducted a sensitivity analysis over a $3 \times 3$ grid of combinations: $\beta \in \{2, 5, 10\}$ and $Z_0 \in \{5, 10, 20\}$~m. Each of these nine configurations corresponds to a separate full retraining of ResNet-50 using SWF with the specified hyperparameters as the optimization objective, resulting in nine additional ResNet-50 training runs beyond those reported in the main experiments. Results were evaluated on the intra-regional Pratas/GBR validation set.

\begin{figure}[H]
    \centering
    \includegraphics[width=1\textwidth]{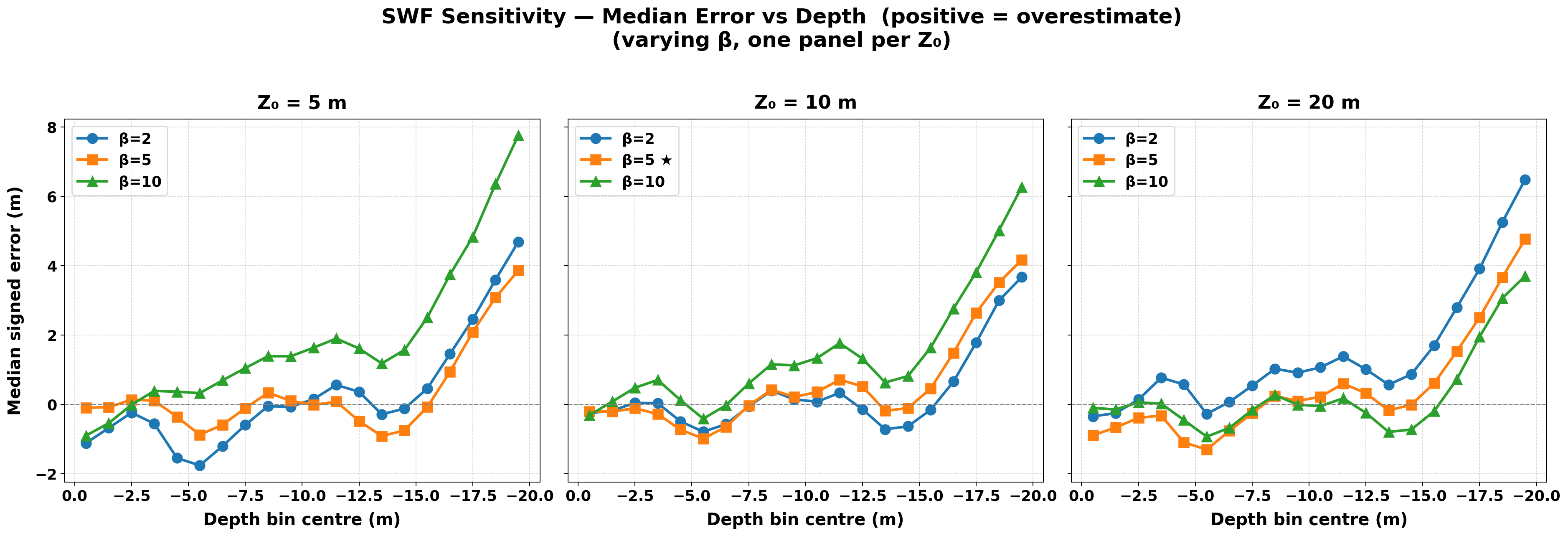}
    \caption{Median signed error as a function of depth bin centre for varying $\beta$ at fixed $Z_0$ values (positive = predicted shallower than true; negative = predicted deeper than true). Each panel corresponds to a fixed decay depth ($Z_0 \in \{5, 10, 20\}$~m), with curves showing how increasing $\beta$ shifts the depth at which prediction bias transitions from near-zero to positive. The selected configuration ($\beta=5$, $Z_0=10$~m) is marked with $\star$.}
    \label{fig:swf_depthcurves_medianErr_vary_beta}
\end{figure}

Fig.~\ref{fig:swf_depthcurves_medianErr_vary_beta} reports median signed error---defined as the median of $\widehat{Z}_i - Z_i$ within each 1~m depth bin, where positive values indicate predictions shallower than the true depth and negative values indicate predictions deeper than the true depth---as a function of depth-bin centre, for varying $\beta$ at each fixed $Z_0$. For a fixed $Z_0$, increasing $\beta$ shifts the depth at which the bias transitions from near-zero to positive: smaller $\beta$ values tend to predict deeper than the true depth in shallow bins, while larger $\beta$ values maintain near-zero bias in shallow water but shift toward predicting shallower than the true depth at intermediate-to-deep depths. This trade-off is most clearly visible in the $Z_0 = 10$~m panel, where the selected configuration ($\beta=5$, $Z_0=10$~m) maintains the most balanced bias profile across the operationally critical 0--15~m depth range before diverging at greater depths.

These results confirm that there is no universally optimal SWF parameterization: the choice of $\beta$ and $Z_0$ instead determines \emph{where} along the depth profile the model is most accurate. The appropriate selection should therefore reflect the depth regime of greatest interest for a given downstream application, with smaller $\beta$ and $Z_0$ favouring very shallow water and larger values broadening emphasis toward intermediate depths.

\vspace{-6pt}

\subsubsection{Image Continuity and Data Splitting Strategy}
\label{subsubsec:image_continuity}

Two training data organization strategies were evaluated. \textbf{Strategy~1 (Random Split)}: images are divided into non-overlapping patches and randomly assigned to 60/20/20\% train/validation/test splits, removing spatial relationships between neighboring samples. \textbf{Strategy~2 (Spatially Continuous Split)}: the study area is partitioned into three contiguous spatial regions, with overlapping patches extracted using a 50\% stride, preserving local geomorphic and radiometric transitions.

Both strategies are evaluated using ResNet50, ResNet101, EfficientNet-B4, and ConvNeXt-Large. Quantitative results and a sensitivity analysis on training volume are summarized in Table~\ref{tab:strategy_comprehensive}, with Strategy~2 (Full) representing the primary experimental configuration used throughout the paper.

\begin{table}[H]
\centering
\caption{Sensitivity analysis of data organization vs. training volume. Strategy~2 (No Augmentation) isolates spatial continuity effects without increasing training volume, while Strategy~2 (Full) represents the main experiment used in this study (repeated from Table~\ref{tab:train_val_test_performance}).}
\label{tab:strategy_comprehensive}
\small
\begin{tabular}{lcccccc}
\toprule
\multirow{2.5}{*}{Model} & \multicolumn{2}{c}{\textbf{Strategy 1 (Random)}} & \multicolumn{2}{c}{\textbf{Strategy 2 (No Aug)}} & \multicolumn{2}{c}{\textbf{Strategy 2 (Full)}} \\
& \multicolumn{2}{c}{$N_{train}=1922$} & \multicolumn{2}{c}{$N_{train}=518$} & \multicolumn{2}{c}{$N_{train}=2590$} \\
\cmidrule(lr){2-3} \cmidrule(lr){4-5} \cmidrule(lr){6-7}
 & RMSE (0--20 m)& $\leq$3m & RMSE (0--20 m) & $\leq$3m & RMSE (0--20 m)& $\leq$3m \\
\midrule
ResNet50        & 3.10 & 0.95 & 2.13 & 0.56 & 1.15 & 0.28 \\
ResNet101       & 2.17 & 1.11 & 1.52 & 0.60 & 1.26 & 0.26 \\
EfficientNet-B4 & 3.22 & 0.76 & 2.99 & 0.92 & 1.79 & 0.37 \\
ConvNeXt-Large  & 1.78 & 0.64 & 2.01 & 0.54 & 1.92 & 0.63 \\
\bottomrule
\end{tabular}
\end{table}

Because Strategy~2 naturally produces more samples through overlapping patches, improved performance could arise simply from increased training volume rather than preserved spatial structure. To decouple these effects, we introduce a control experiment, Strategy~2 (No Aug), which preserves spatial continuity while removing data augmentation. This configuration reduces the training set size to $N=518$ patches, approximately $27\%$ of the Strategy~1 baseline. This setting isolates the effect of spatial organization independent of dataset size.

Under Strategy~1, models exhibit moderate errors. When spatial continuity is preserved and full augmentation is applied (Strategy~2 Full), performance improves consistently across most architectures. Relative to Strategy~1, overall RMSE is reduced by approximately $63\%$ (ResNet50), $42\%$ (ResNet101), and $44\%$ (EfficientNet-B4). In contrast, ConvNeXt-Large exhibits a slight RMSE increase of approximately $8\%$ under Strategy~2 (Full) (Table~\ref{tab:strategy_comprehensive}). Improvements are even more pronounced in shallow waters ($\leq 3$~m), where RMSE decreases by approximately $71\%$, $77\%$, and $51\%$ for ResNet50, ResNet101, and EfficientNet-B4, respectively, while ConvNeXt-Large shows only marginal change ($\sim2\%$ reduction).

The control setting, Strategy~2 (No Aug), also improves shallow-water accuracy compared to Strategy~1 despite using substantially fewer training samples ($N=518$), indicating that performance gains are not solely attributable to increased training volume. 

\begin{figure}[H]
\centering
\setlength{\tabcolsep}{2pt}

\begin{tabular}{cc}

    \includegraphics[width=0.43\textwidth]{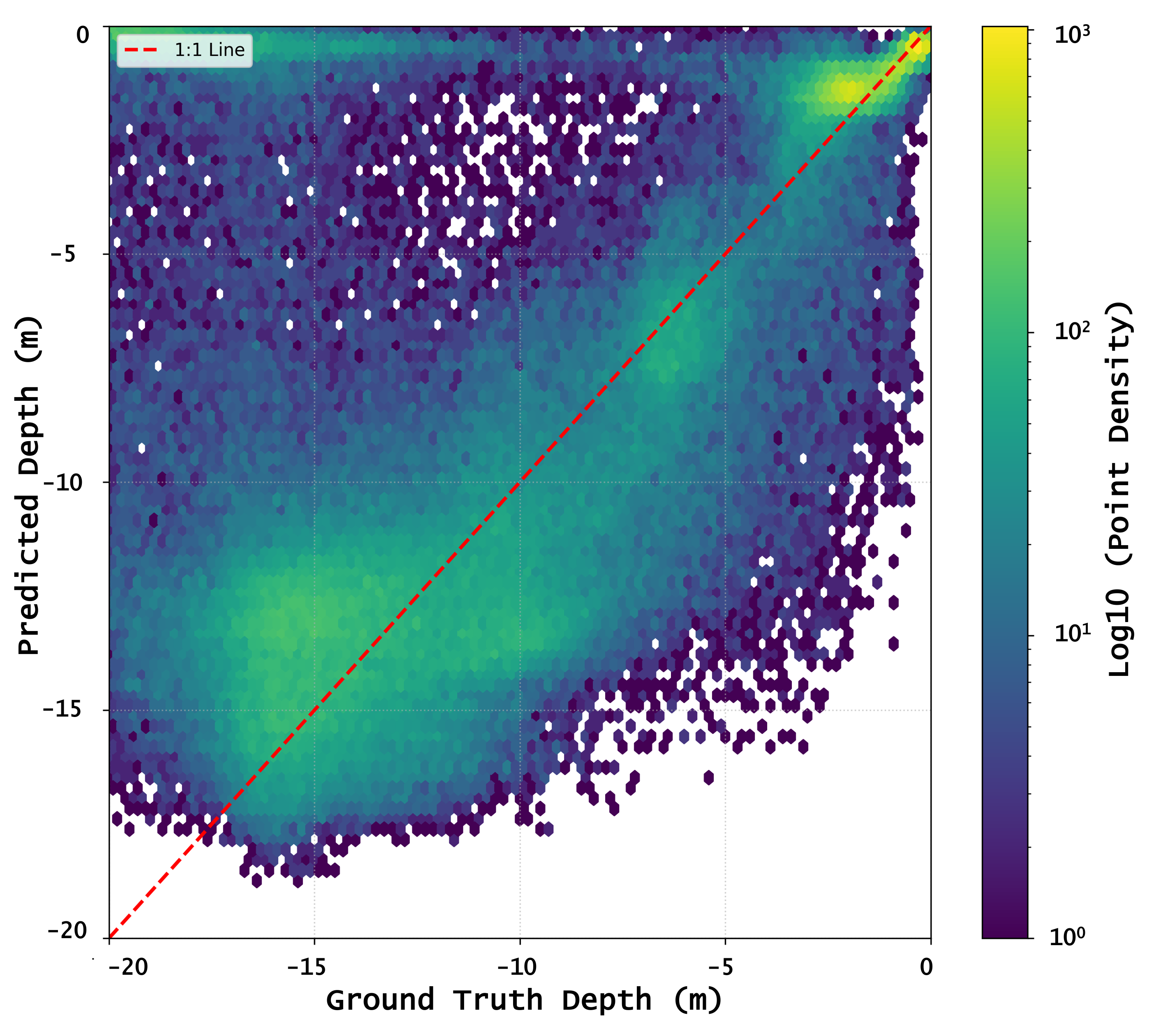} &
    \includegraphics[width=0.43\textwidth]{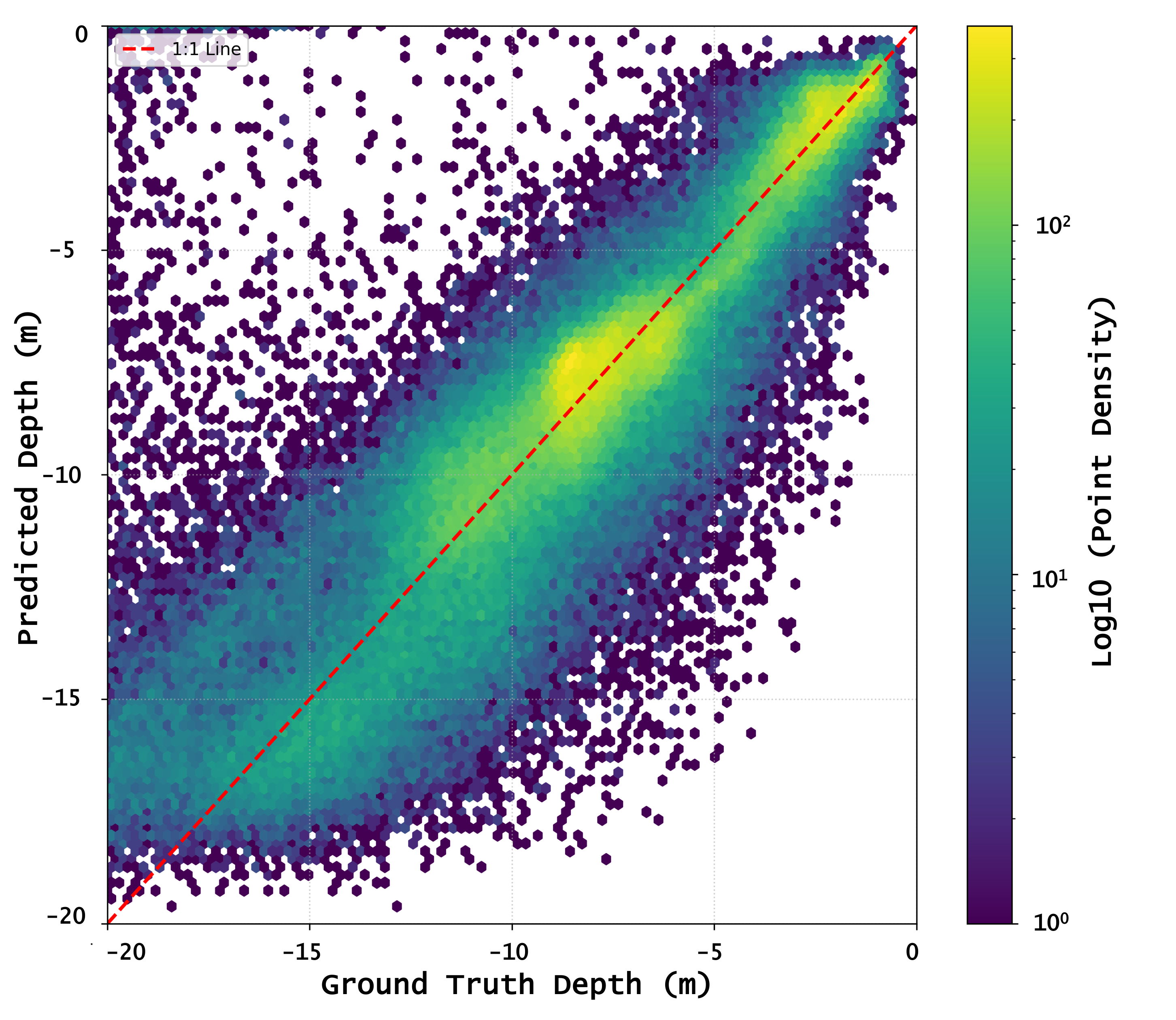} \\
    \footnotesize (a1) ResNet50 (S1) &
    \footnotesize (a2) ResNet50 (S2) \\[6pt]

    \includegraphics[width=0.43\textwidth]{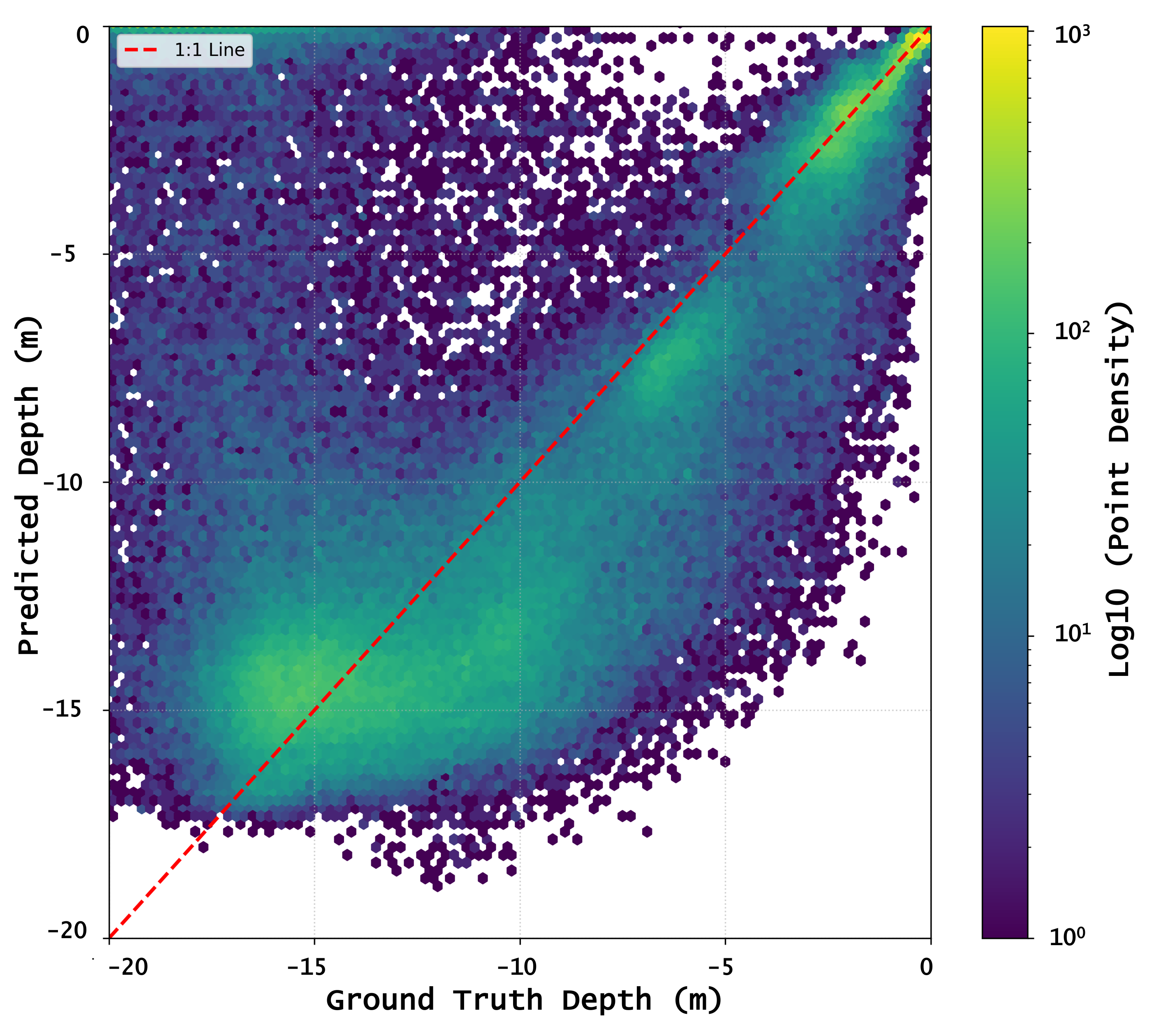} &
    \includegraphics[width=0.43\textwidth]{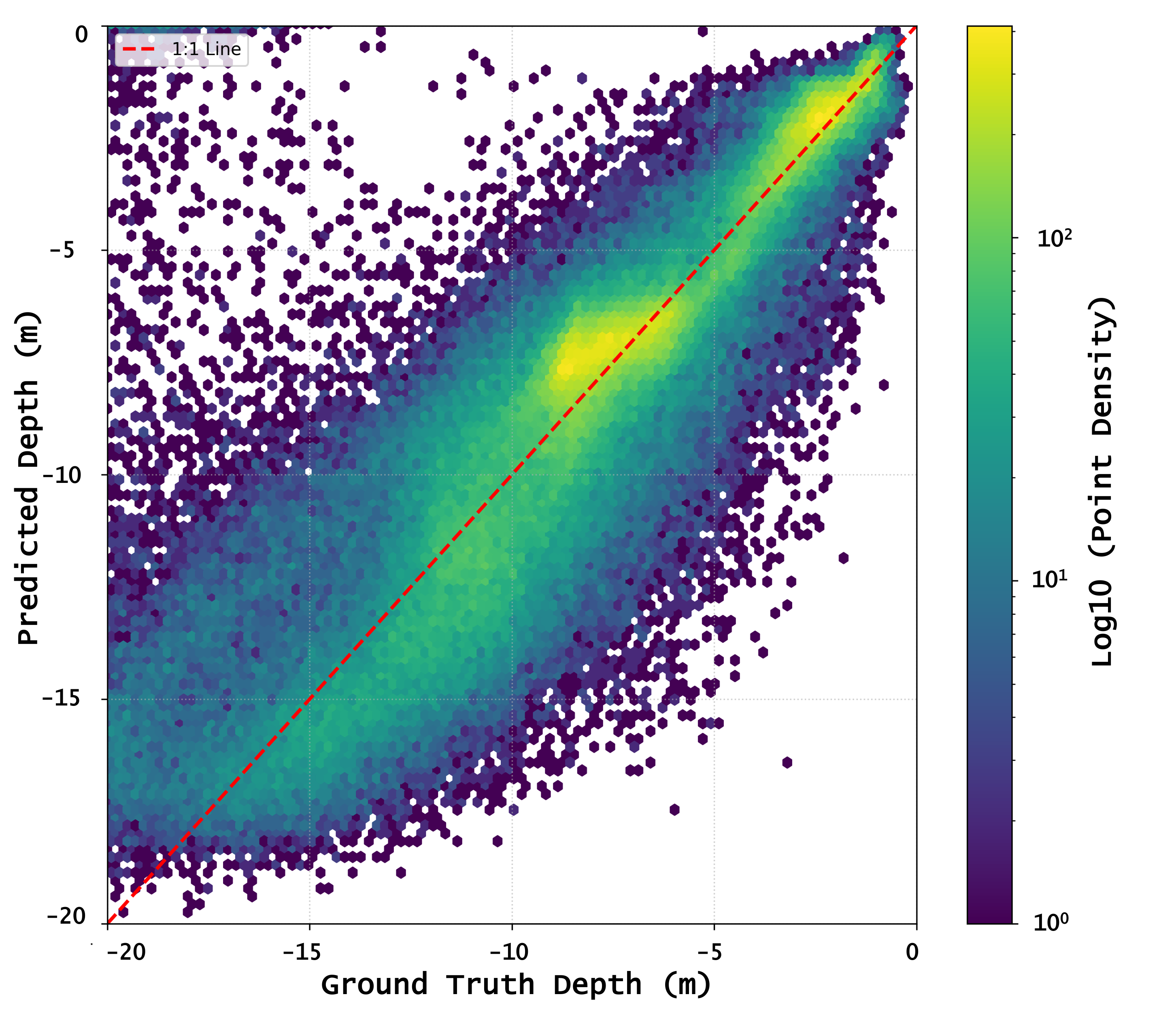} \\
    \footnotesize (b1) ResNet101 (S1) &
    \footnotesize (b2) ResNet101 (S2) \\[6pt]

    \includegraphics[width=0.43\textwidth]{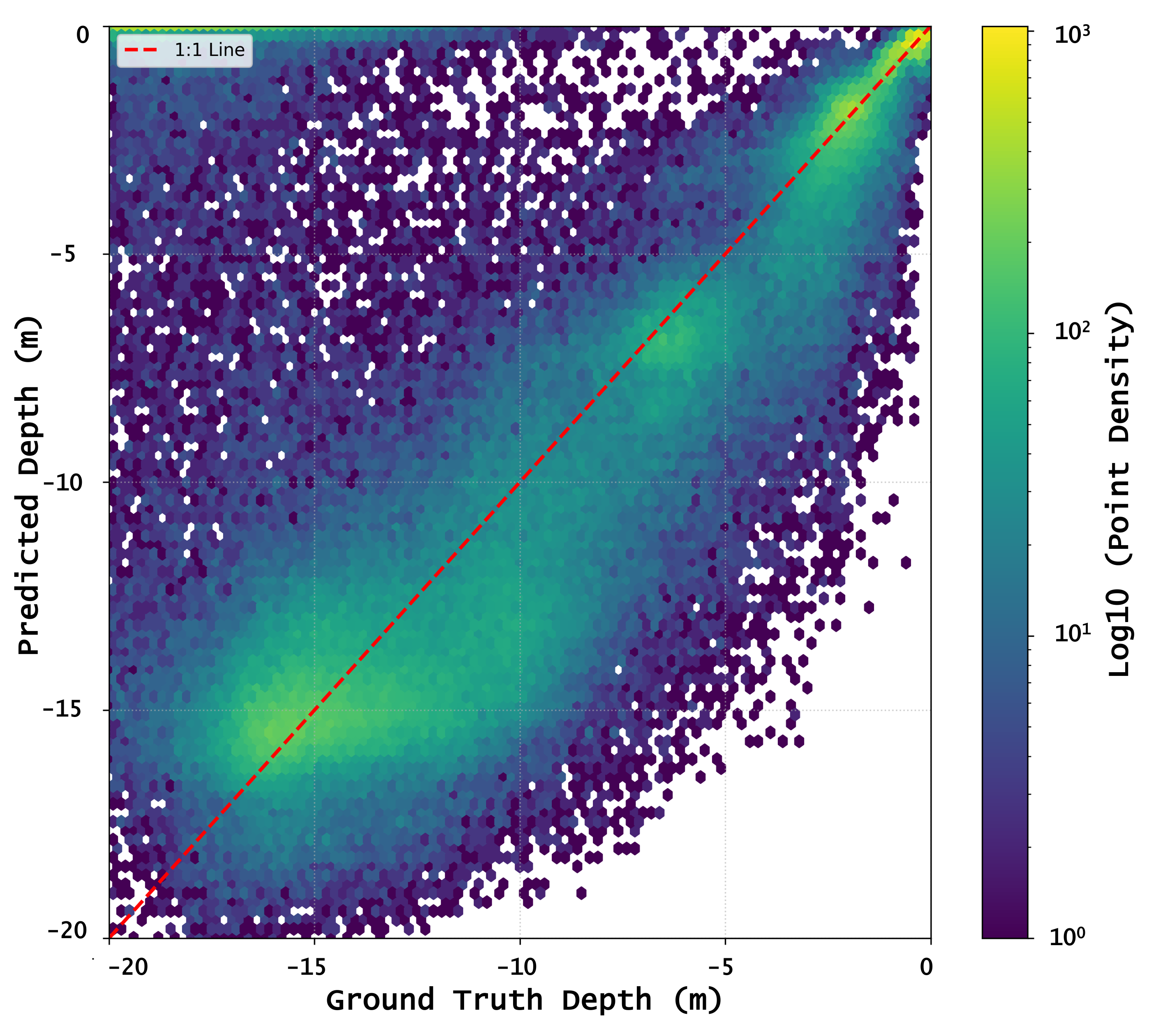} &
    \includegraphics[width=0.43\textwidth]{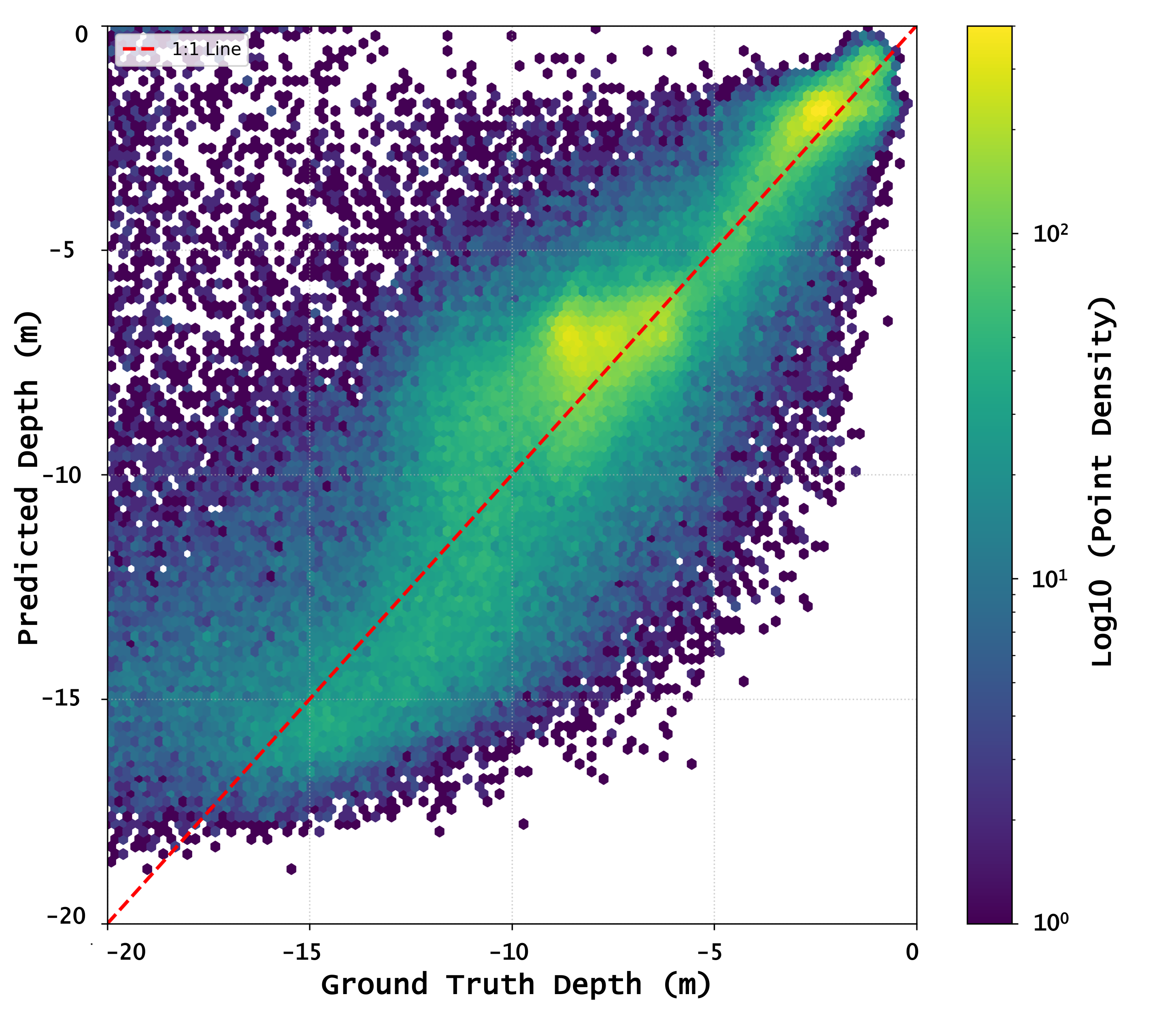} \\
    \footnotesize (c1) Eff-B4 (S1) &
    \footnotesize (c2) Eff-B4 (S2) \\[6pt]

    \includegraphics[width=0.43\textwidth]{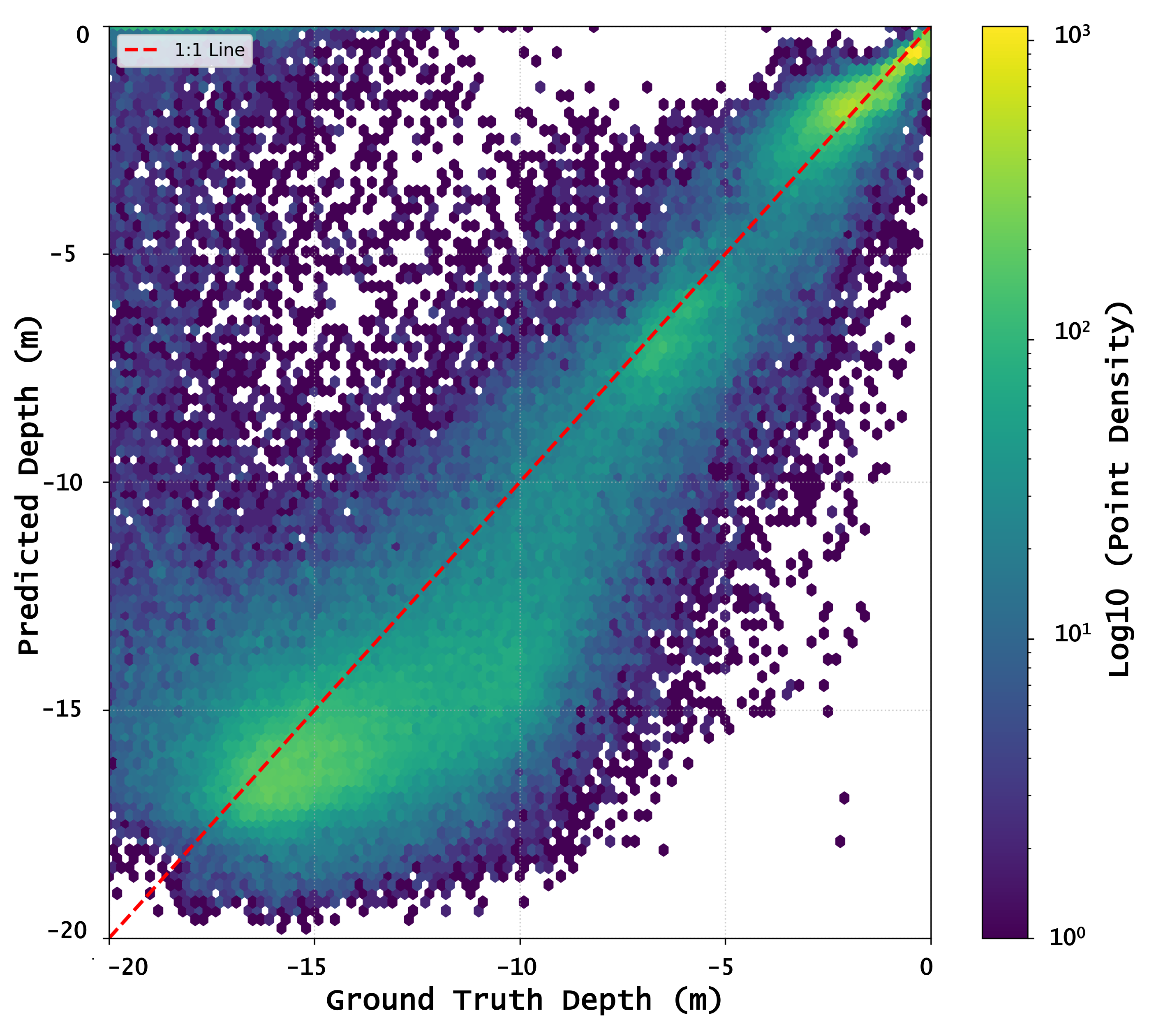} &
    \includegraphics[width=0.43\textwidth]{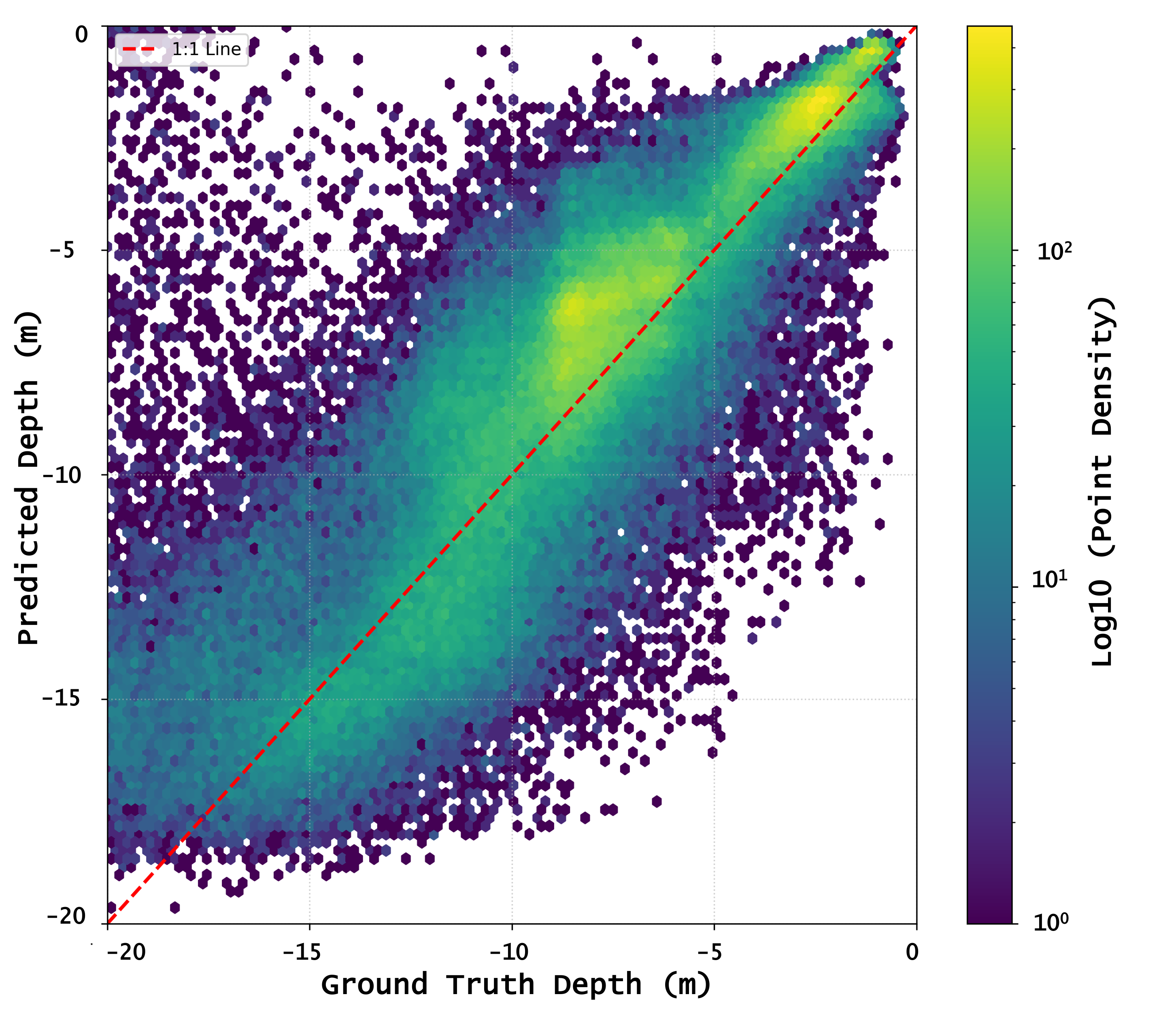} \\
    \footnotesize (d1) ConvNeXt-L (S1) &
    \footnotesize (d2) ConvNeXt-L (S2) \\

\end{tabular}

\caption{Hexbin density plots of predicted versus ground truth bathymetric depth for four models under two data-splitting strategies. Panels (a1--d1) correspond to Strategy~1 (random patch split), while panels (a2--d2) correspond to Strategy~2 (spatially continuous split).}
\label{fig:Pred_vs_GT_Hexbin}

\end{figure}

An exception is observed for ConvNeXt-Large, which achieves slightly lower overall RMSE under Strategy~1 (1.78~m) than under Strategy~2 (Full) (1.92~m). However, visual inspection of the prediction-ground truth relationship (Fig.~\ref{fig:Pred_vs_GT_Hexbin}) reveals increased dispersion under Strategy~1, suggesting differences in prediction structure despite comparable numerical accuracy.

To further evaluate robustness, cross-regional generalization experiments were conducted on geographically distinct reef systems (Ashmore and Cartier). As shown in Table~\ref{tab:cross_regional_strategy}, Strategy~2 (Full) yields the lowest RMSE across both regions for all evaluated architectures, whereas Strategy~1 produces substantially higher transfer errors. Strategy~2 (No Aug) demonstrates intermediate performance, generally outperforming Strategy~1 despite its reduced training volume.

Beyond numerical metrics, spatial continuity also influences prediction structure. Fig.~\ref{fig:Pred_vs_GT_Hexbin} shows that models trained with Strategy~2 exhibit tighter alignment with the 1:1 depth relation, particularly at shallow and intermediate depths. In contrast, Strategy~1 models display greater dispersion and a tendency toward underestimation at larger depths ($>12$~m).

\vspace{-6pt}

%
%

\vspace{-6pt}

\begin{table}[H]
\centering
\caption{Cross-regional generalization RMSE (m) for models trained under different data-splitting strategies and evaluated on independent reef systems (depth range 0--20~m). Lower values indicate better transferability.}
\label{tab:cross_regional_strategy}
\small
\begin{tabular}{llcccc}
\toprule
\textbf{Strategy} & \textbf{Region} & \textbf{ResNet50} & \textbf{ResNet101} & \textbf{Eff-B4} & \textbf{ConvNeXt-L} \\
\midrule
Strategy~1 (Random) 
& Ashmore & 4.36 & 3.94 & 6.24 & 3.55 \\
& Cartier & 5.66 & 3.74 & 5.34 & 3.30 \\
\midrule
Strategy~2 (No Aug) 
& Ashmore & 3.43 & 3.29 & 5.50 & 4.74 \\
& Cartier & 3.89 & 3.61 & 5.93 & 5.52 \\
\midrule
Strategy~2 (Full) 
& Ashmore & 2.88 & 2.56 & 4.49 & 2.98 \\
& Cartier & 2.59 & 2.85 & 3.78 & 2.46 \\
\bottomrule
\end{tabular}
\end{table}

\vspace{-6pt}
\begin{figure}[H]
\centering

\textbf{(a) Ashmore Reef (Strategy 1)}\\
\vspace{2pt}

\begin{tabular}{cccc}
\includegraphics[width=0.23\textwidth]{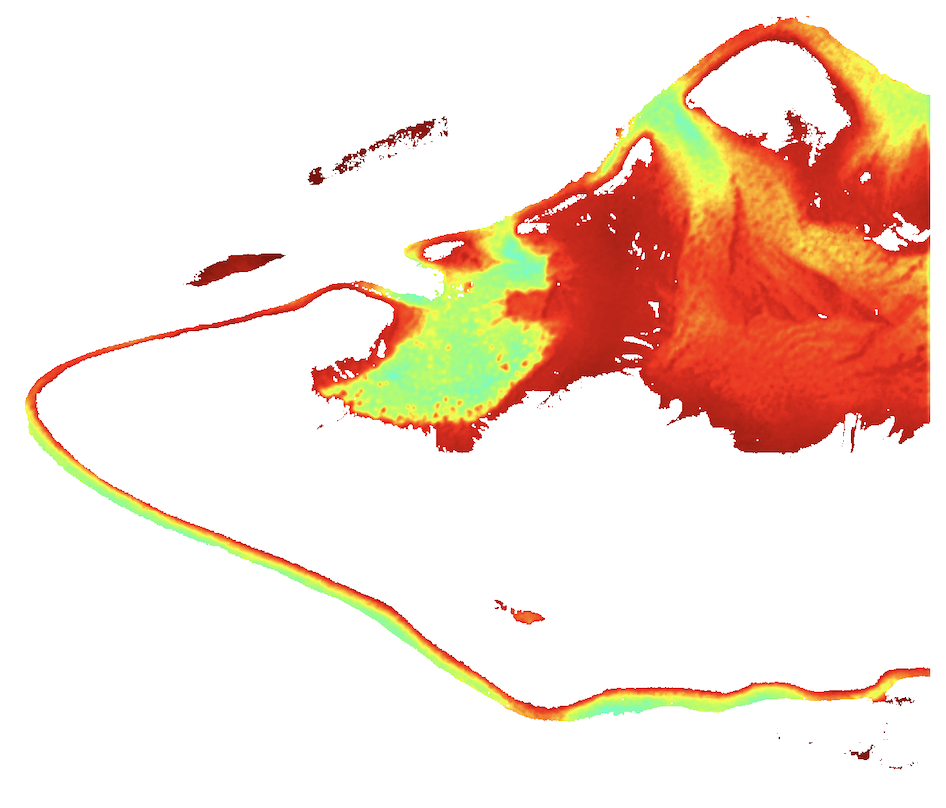} &
\includegraphics[width=0.23\textwidth]{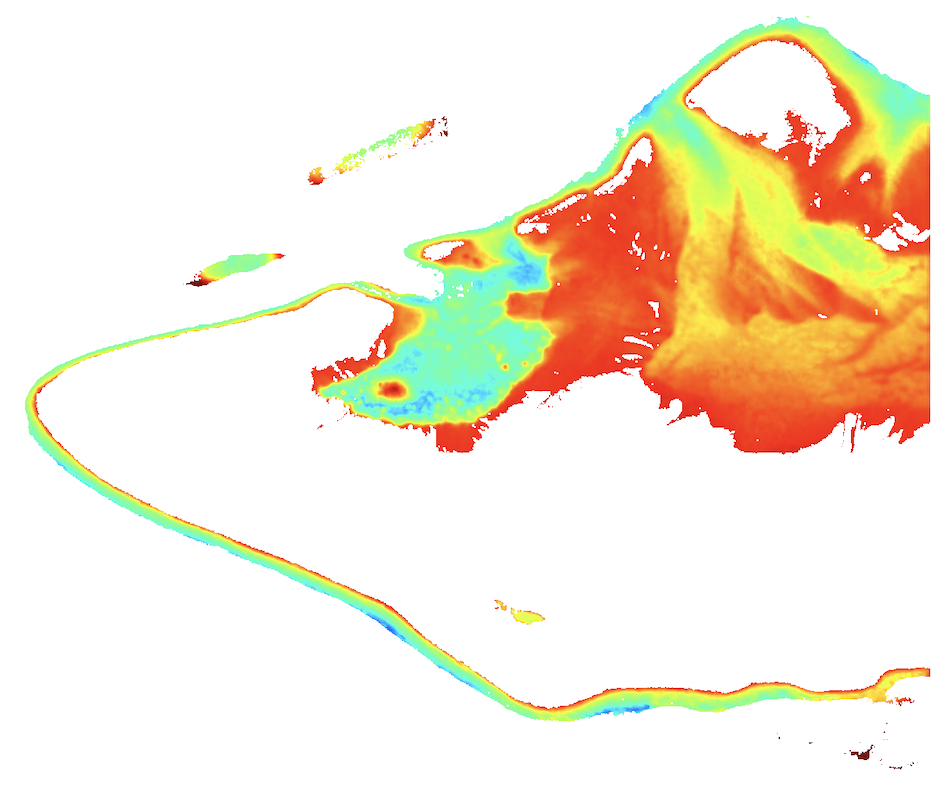} &
\includegraphics[width=0.23\textwidth]{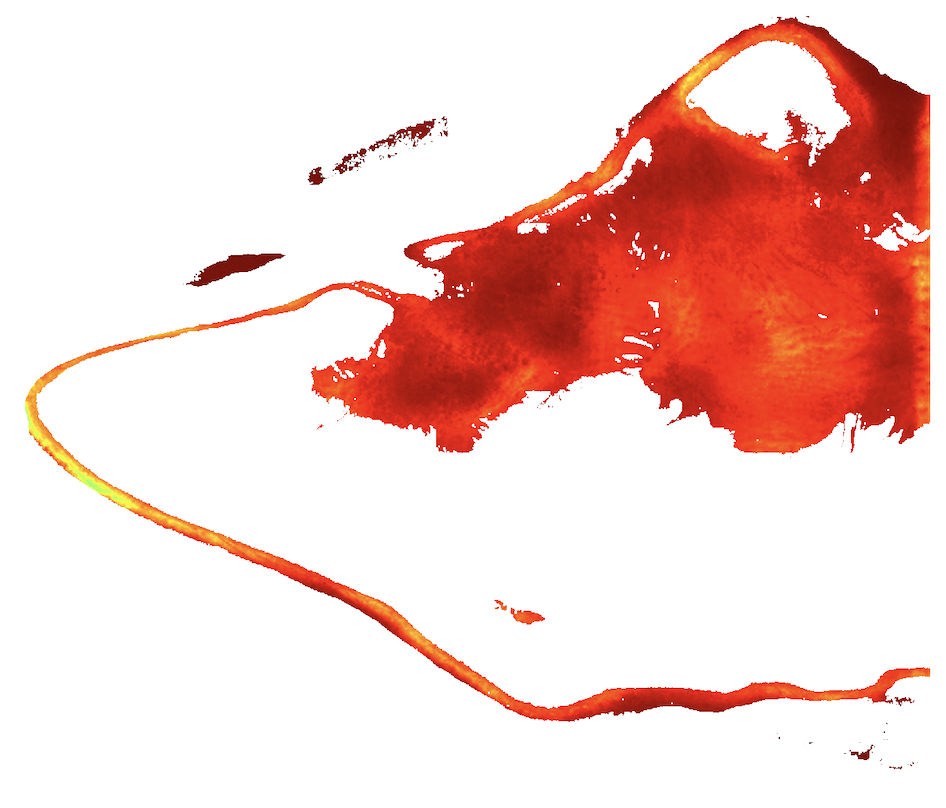} &
\includegraphics[width=0.23\textwidth]{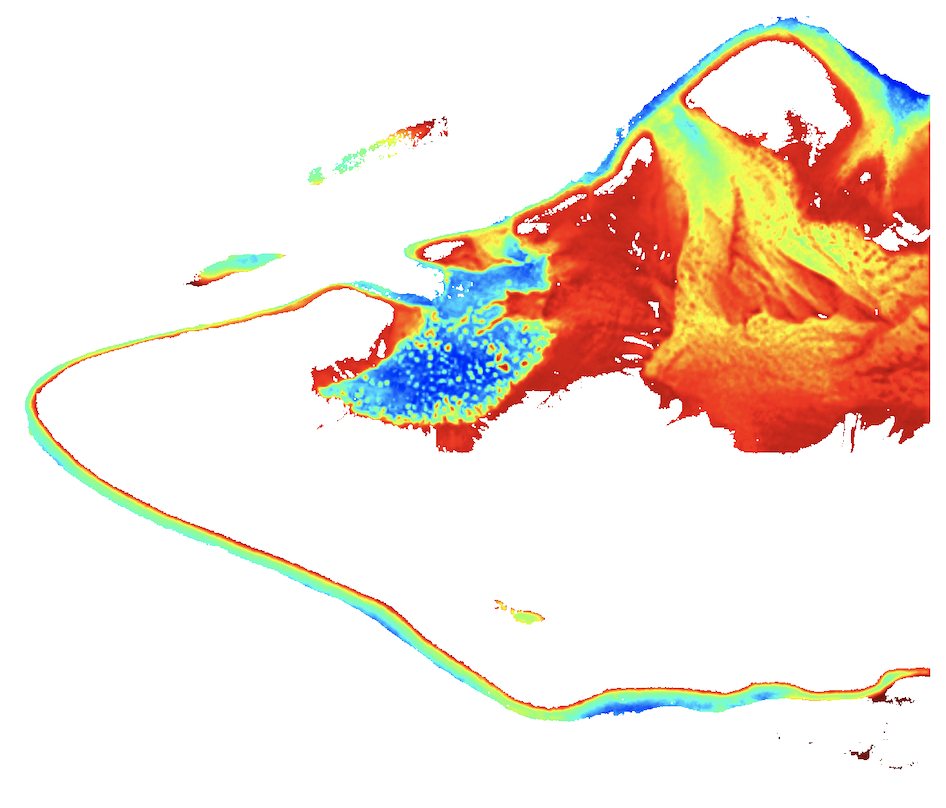} \\
\footnotesize ResNet50 & \footnotesize ResNet101 & \footnotesize Eff-B4 & \footnotesize ConvNeXt-L
\end{tabular}

\vspace{8pt}

\textbf{(b) Cartier Reef (Strategy 1)}\\
\vspace{2pt}

\begin{tabular}{cccc}
\includegraphics[width=0.23\textwidth]{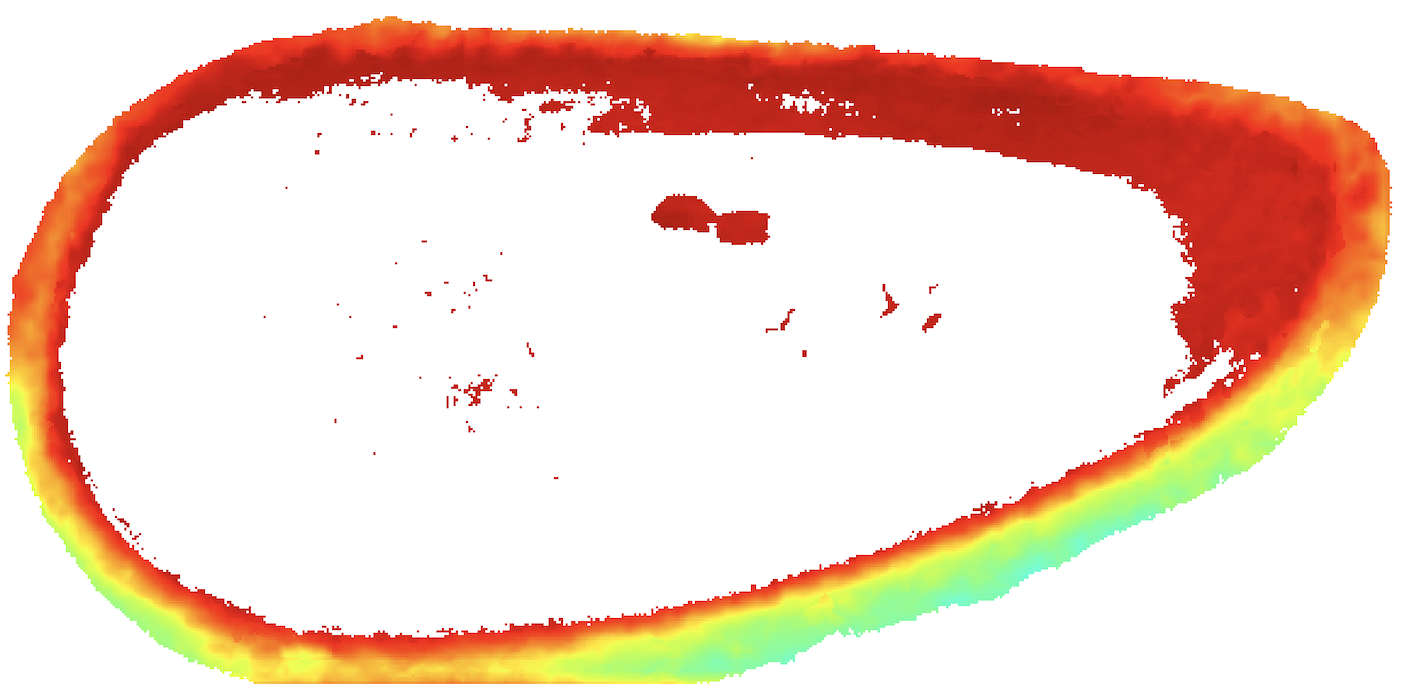} &
\includegraphics[width=0.23\textwidth]{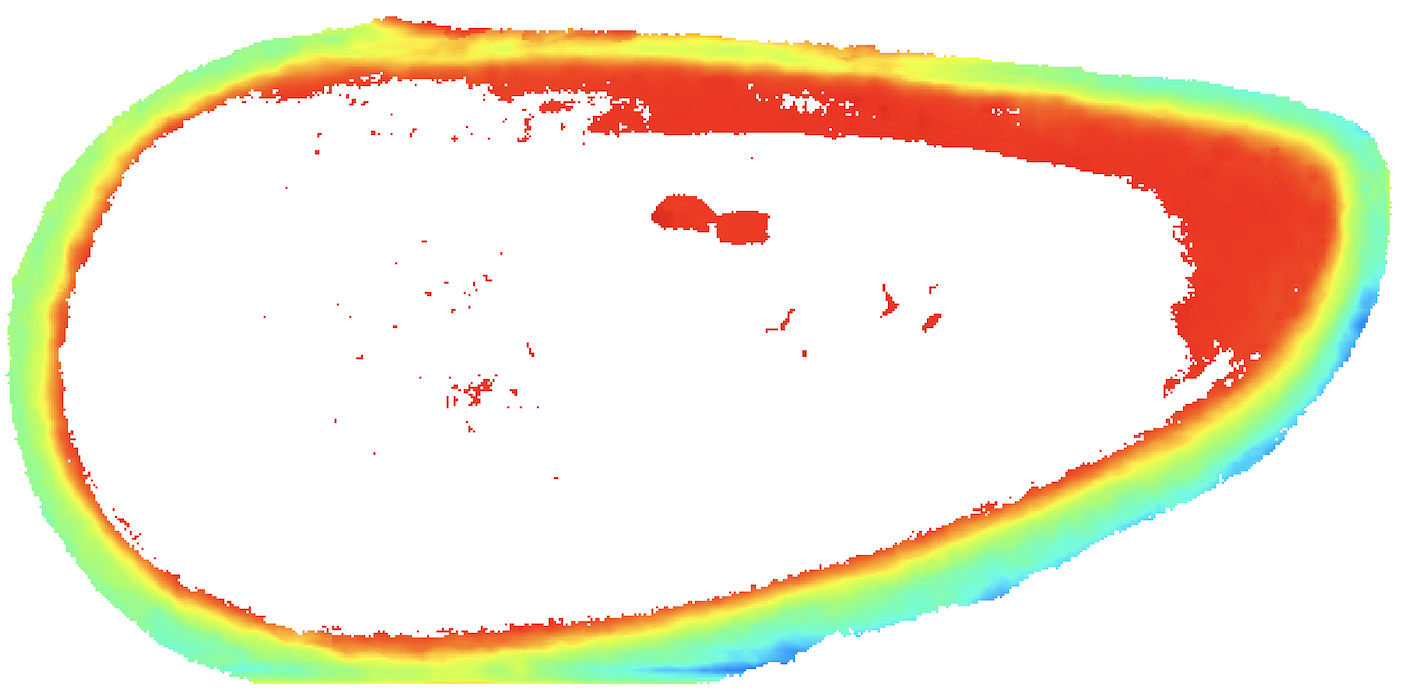} &
\includegraphics[width=0.23\textwidth]{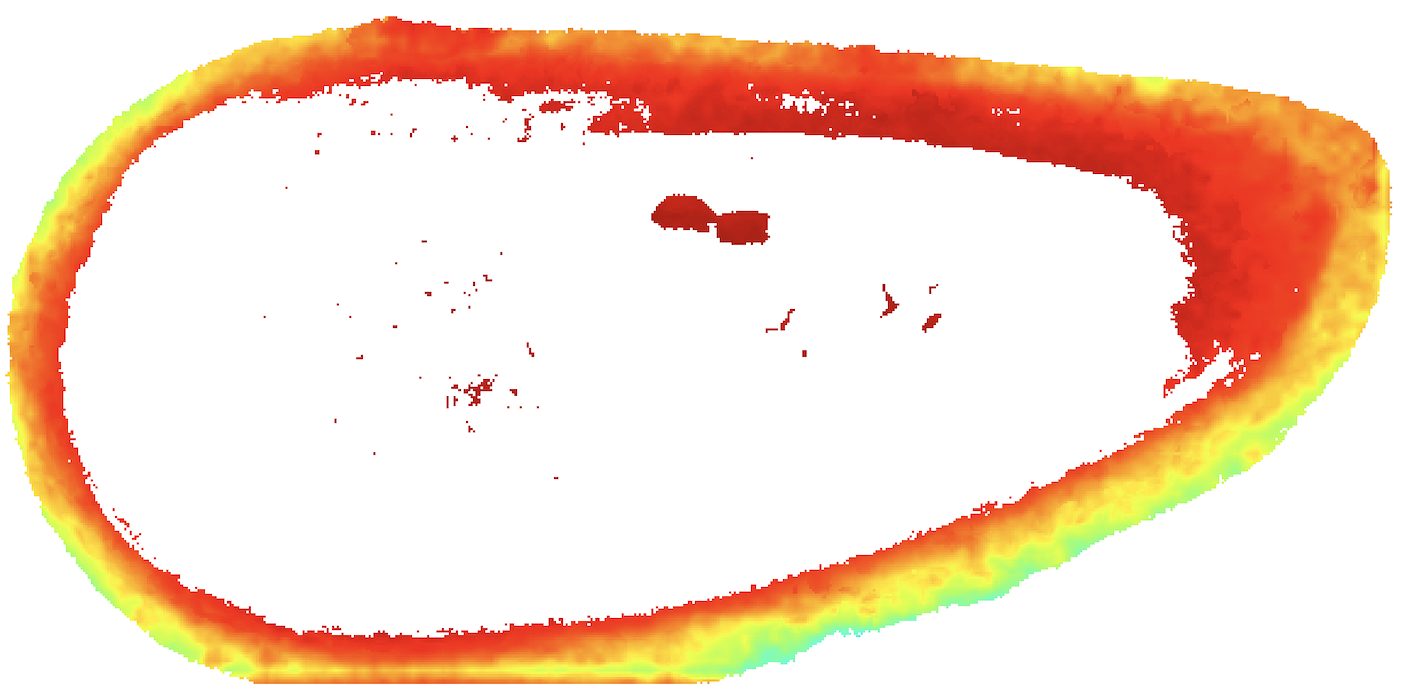} &
\includegraphics[width=0.23\textwidth]{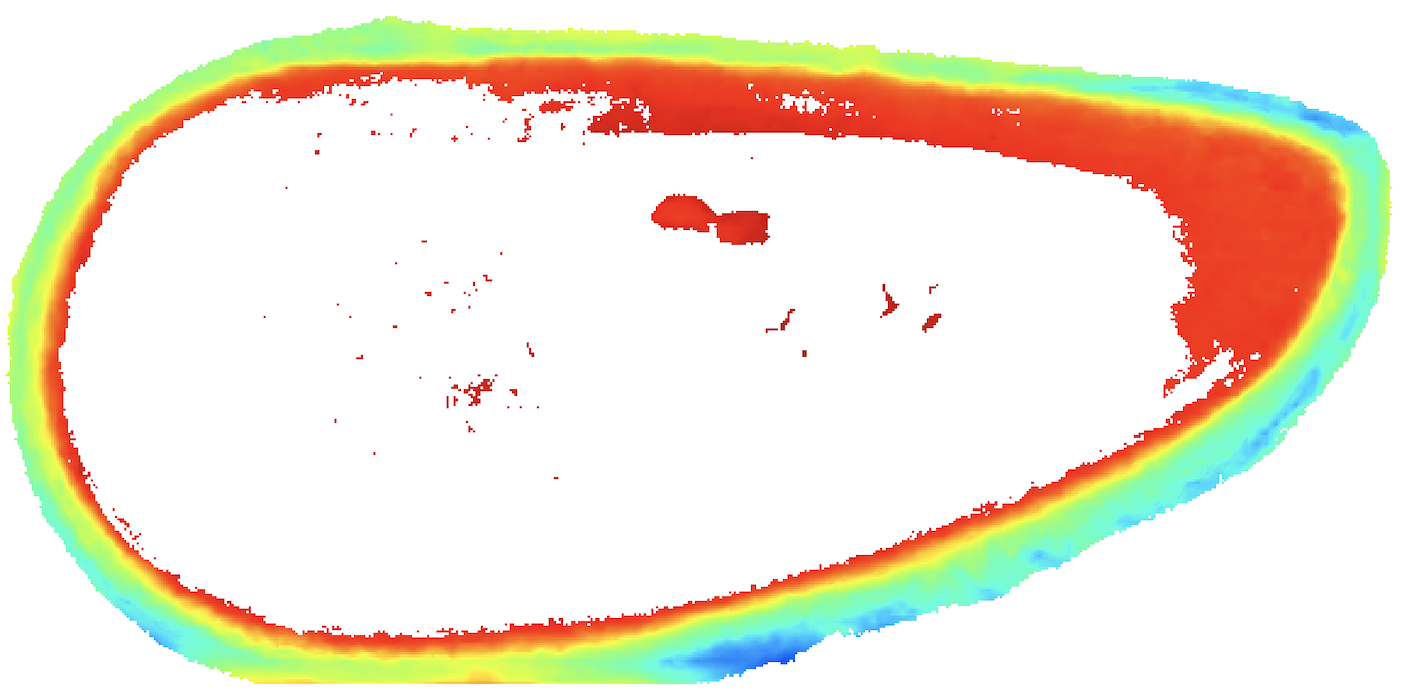} \\
\footnotesize ResNet50 & \footnotesize ResNet101 & \footnotesize Eff-B4 & \footnotesize ConvNeXt-L
\end{tabular}

\caption{Spatial prediction maps under Strategy~1 (random patch split). Row (a): Ashmore Reef; Row (b): Cartier Reef. Each column corresponds to a different deep learning architecture. Results highlight spatial inconsistencies, depth-dependent distortions, and reduced geomorphic coherence when spatial continuity is not preserved during training.}
\label{fig:strategy1_1x4}

\end{figure}

Additional controls on depth-distribution differences and cross-strategy generalization are provided in the Supplementary Information (Tables~S1--S2).

\section{Discussion}
\label{sec:discussion}

\subsection{Spectral Limits of Statistical Models}
\label{subsubsec:rf_results_discussion}
The Random Forest baseline illustrates the predictive capability achievable from pixel-level spectral information alone. Feature importance analysis confirms that performance is primarily driven by the visible spectrum, particularly the green band (B3), consistent with light-penetration theory. Within a single region, these spectral-depth relationships achieve high accuracy. However, RF models rely on local statistical associations and do not encode spatial structure, so performance degrades under cross-regional transfer. This establishes an important baseline: while statistical models effectively exploit within-region optical signals, their generalization is constrained by the physical and environmental variability of coastal systems.

\subsection{Spatial-Spectral Feature Learning in Deep Networks} 
\label{subsec:discussion_dl}

\subsubsection{Intra-Regional Performance} 

Intra-regional experiments reveal clear differences in how architectures learn spatial-spectral relationships. Although validation RMSE consistently exceeds training RMSE (Table~\ref{tab:train_val_test_performance}), predictions remain physically consistent across multiple reefs, suggesting that models learn transferable spectral-depth relationships. ResNet50 and ResNet101 achieve the strongest overall performance with shallow-water RMSE below 0.3~m. EfficientNet-B4 and ConvNeXt-Large show moderately higher errors in shallow regions, while Random Forest remains competitive numerically but lacks spatial detail due to its pixel-wise formulation.

Training dynamics highlight architectural differences: despite its larger parameter count, ConvNeXt-Large converges in a similar number of epochs as EfficientNet-B4, whereas ResNet models require longer optimization. Increasing patch size from $128 \times 128$ to $512 \times 512$ improves performance across all models, confirming that bathymetric prediction benefits from contextual information beyond local radiometric texture.

\subsubsection{Cross-Regional Performance}

Multi-temporal median aggregation consistently reduces RMSE relative to mean-based averaging across all models (Table~\ref{tab:transfer_metrics}), demonstrating the benefit of robust temporal fusion in cross-regional bathymetric prediction. This improvement reflects the reduced influence of acquisition-specific noise (e.g., clouds, cloud shadows, and atmospheric variability), which can otherwise introduce inconsistent depth estimates across individual Sentinel-2 scenes.


The joint analysis of RMSE, MAE, and $R^2$ in Table~\ref{tab:transfer_metrics} provides additional insight into error characteristics and explained variance under domain shift. For all models, RMSE is consistently higher than MAE, indicating the presence of larger deviations that disproportionately affect overall error magnitude. Meanwhile, $R^2$ values indicate that ResNet-based models and ConvNeXt-Large retain substantial explanatory power despite regional transfer, suggesting that a significant portion of bathymetric variability remains recoverable. This implies that cross-regional errors are not uniformly distributed but are influenced by occasional high-error predictions, particularly in optically complex or geomorphically heterogeneous areas.

This effect is more evident for ConvNeXt-Large (median RMSE = 2.98~m, median MAE = 1.56~m at Ashmore), where the gap between RMSE and MAE indicates improved robustness under median aggregation, while still retaining sensitivity to isolated large errors.

ResNet-based models and ConvNeXt-Large demonstrate relatively strong transferability across regions, whereas EfficientNet-B4 exhibits weaker generalization performance. Random Forest shows limited benefit from temporal aggregation and systematically higher errors in lagoon and outer-reef environments.

Across all models, performance degradation is most pronounced in geomorphically complex regions, suggesting that improved representation of high-relief reef structures and enhanced domain diversity could further improve cross-regional robustness.

Architectures that preserve hierarchical spatial context (ResNet and ConvNeXt) maintain more stable performance across shallow and intermediate depths. ConvNeXt-Large achieves its lowest cross-regional RMSE at intermediate depths, with minimum errors of approximately 0.97--1.01~m at 7--10~m depth at Cartier Reef and 0.65--0.85~m at 2--4~m depth at Ashmore Reef. EfficientNet-B4 and Random Forest show stronger depth-dependent degradation, with Random Forest exhibiting systematic overestimation across both sites.

Taken together, the cross-regional results demonstrate meaningful progress toward reef-to-reef transferability under clear-water conditions, while confirming that geographic transfer still incurs a substantial accuracy penalty relative to intra-regional performance. All study sites are outer-shelf or oceanic reefs with clear-water conditions; therefore, performance may differ in inshore or turbid environments.

\subsubsection{Benchmarking Against Reference Architectures}
\label{subsec:discussion_benchmark}


The benchmark experiment provides a comparative evaluation under consistent shallow-water preprocessing and evaluation conditions. Compared to the site-specific models reported in \cite{Agrafiotis2025}, the unified architectures achieve lower RMSE while requiring substantially fewer parameters than the transformer-based Swin-BathyUNet. These results suggest that modern convolutional backbones can provide strong representational capacity without the parameter overhead of large transformer designs, offering robust and parameter-efficient alternatives for shallow-water bathymetry estimation.

\subsection{Training Design Effects: Loss Functions and Spatial Continuity}
\label{subsec:training_design_discussion}

Beyond architectural design, training strategy plays a critical role in depth-dependent learning and cross-regional transferability. Both loss-function formulation and spatial organization of training data significantly influence model behavior within the 0--20~m depth range.

\subsubsection{Loss-Function Effects on Depth-Dependent Learning}

The SWF loss improves accuracy in optically sensitive shallow zones by up-weighting near-surface residuals during training, with the most pronounced benefit observed for higher-capacity architectures such as ConvNeXt-Large. RPE provides a complementary strategy by normalizing residuals relative to depth, yielding more balanced performance at greater depths. Standard RMSE tends to underperform when either shallow precision or depth-normalized stability is required.

Loss-function effectiveness appears to depend jointly on architectural capacity and depth-distribution mismatch between training and transfer domains. As shown in Table~\ref{tab:depth_distribution} and Fig.~\ref{fig:depth_bin_coverage_barplot}, Ashmore is strongly concentrated in shallow waters (0--4~m), whereas GBR contains a larger proportion of deeper pixels, resulting in broader depth coverage in the training set relative to the transfer domains. This distributional shift likely influences optimization behavior and learned spectral--depth relationships. ResNet101 maintains stable performance across all three loss functions, suggesting its learned representations are less sensitive to loss weighting. ConvNeXt-Large, by contrast, responds more strongly to loss design: SWF consistently yields the lowest depth-binned RMSE across shallow-to-intermediate depths, while RPE becomes advantageous beyond $\sim$15~m where shallow weighting is no longer beneficial. Overall, loss selection should be aligned with the target depth range and architectural capacity of the chosen model.

\subsubsection{Spatial Continuity and Geomorphic Context}

Models trained using spatially continuous patches consistently outperform those trained with randomly split patches, even when the latter contain more independent samples. Bathymetric depth is a continuous regression target---distinct from a per-pixel classification problem---with spatial structure operating across multiple scales: broad land-to-shallow-to-deep gradients are superimposed on local fine-scale variability such as coral blocks, reef ridges, and escarpments, and some geomorphic settings (e.g., reef fronts and fore-reef slopes) exhibit sharp local gradients rather than gradual transitions. Models trained on spatially continuous patches with sufficient receptive field can in principle learn both regimes; preserving spatial continuity enables networks to learn physically meaningful spectral--depth transitions, whereas random splitting disrupts these transitions and produces locally accurate but globally unstable mappings.

Cross-strategy experiments (Supplementary Section~S1) confirm this asymmetry: models trained with spatially continuous data generalize effectively to randomly distributed test tiles, whereas the reverse does not hold. Qualitative analysis (Fig.~\ref{fig:Pred_vs_GT_Hexbin}) shows tighter 1:1 alignment and reduced shallow bias under Strategy~2. Interestingly, ConvNeXt-Large exhibits stronger dependence on training volume: spatial continuity alone (Strategy~2 No Aug) does not improve cross-regional performance for this model, but the combination of preserved continuity and augmentation yields the strongest generalization, confirming the expectation that high-capacity architectures require more data to fully leverage geomorphic structure.

Additional visual comparisons of cross-regional predictions further illustrate these differences. Strategy~1 outputs (Fig.~\ref{fig:strategy1_1x4}) appear spatially blurred and exhibit reduced geomorphic definition, particularly in lagoon regions where deeper waters are frequently underestimated as shallow surfaces. In contrast, Strategy~2 outputs (Figs.~\ref{fig:ashmore_resultsfor5models} and \ref{fig:cartier_resultsfor5models}) preserve sharper reef-flat-to-slope transitions and more coherent spatial structure across both Ashmore and Cartier. These results suggest that preserving spatial continuity helps reduce spectral--depth ambiguity and enables the model to better learn physically consistent geomorphic depth transitions.

We acknowledge that the geomorphic-learning interpretation proposed here is a hypothesis supported by quantitative and qualitative evidence, rather than a directly verified mechanism. A fully rigorous attribution of these performance gains to geomorphic learning, as opposed to other forms of implicit spatial regularization, would require explicit interpretability analyses such as feature-map probing or activation-based class-activation mapping. A companion study by some of the present authors~\cite{Chowdhury2026} develops exactly such tools, including a regression-adapted Ablation-CAM (A-CAM-R) and a performance-retention test, applied to the Swin-BathyUNet model on the MagicBathyNet benchmark also used in this study, and reports findings consistent with the importance of wide spatial receptive fields for reliable Sentinel-2 depth retrieval. Extending such interpretability analyses to the architectures and Sentinel-2 datasets used in the present work is identified as an important direction for future work.

\subsection{Influence of Seafloor Type and Water-Column Variability on Transferability}
\label{subsec:seafloor_watercolumn}

Beyond architecture and training design, cross-regional accuracy loss is also influenced by environmental variability between reef systems, including differences in substrate composition, water-column optical properties, and surface effects such as sun-glint. These factors introduce spectral ambiguity and complicate the learning of transferable spectral--depth relationships across sites \cite{Liu2025,Santos2025}. More broadly, reduced transfer performance likely reflects both environmental differences between reef systems and insufficient representation of this variability within the training dataset, limiting the ability of the models to generalize across unseen conditions.


Substrate composition strongly influences the bottom-leaving reflectance signal that drives SDB. Carbonate sand produces high albedo across the visible spectrum, while coral framework, algal turf, and seagrass have progressively lower and spectrally distinct reflectances \cite{Stumpf2003,Hedley2018}. Dark substrates such as coral rubble or encrusting coralline algae can produce reflectance signals similar to those of much deeper clear-water columns, creating a fundamental depth--substrate ambiguity that is difficult to fully resolve using spectral information alone \cite{Lyzenga1978,Lyzenga2006}. If the substrate composition at a transfer site (e.g., Ashmore or Cartier) differs systematically from that at training sites (Pratas Island, GBR), the spectral-depth relationships learned during training will be partially mismatched at inference time, contributing directly to elevated cross-regional RMSE. Spatial continuity during training partially mitigates this by enabling models to learn substrate transitions in geomorphic context---a reef flat bounded by a fore-reef slope carries different albedo expectations than an isolated shallow patch---but cannot eliminate the ambiguity introduced by genuinely novel substrate assemblages.

Water-column optical properties further compound this effect. The inherent optical properties (IOPs) of reef waters---absorption and backscattering coefficients driven by colored dissolved organic matter (CDOM), phytoplankton, and non-algal particles---vary across sites and seasons, modulating how spectral reflectance attenuates with depth. Even under the clear-water conditions of all study sites in this work, inter-site differences in IOPs shift the effective attenuation coefficient, meaning that equivalent bottom reflectance values correspond to different depths at different locations \cite{Lyzenga2006}. Sun-glint further adds site- and date-specific spectral contamination that, while partially mitigated by multi-temporal median aggregation, is not fully removed without explicit correction \cite{Hedley2005}. 

An important caveat is that depth composition, substrate type, bottom albedo, water-column optical properties, turbidity, and reef morphology vary jointly between the training and transfer regions and cannot be independently controlled within the available dataset. The present study therefore does not quantitatively isolate the individual contributions of these factors to the observed cross-regional accuracy degradation. We identify the disentanglement of these drivers as an important direction for follow-on work, in particular through physics-informed architectures with dedicated output heads for substrate reflectance and water-column attenuation, as outlined in the Future Work section.

Future work could further improve cross-regional transferability by explicitly incorporating seafloor-type information and water-column optical characterization into the model framework (although this information is rarely available at comparable spatiotemporal resolutions), together with expanded training coverage across more diverse reef environments and optical conditions.

\subsection{Implications for Transferable Satellite-Derived Bathymetry}
\label{subsec:implications_transferable_sdb}

The results of this study provide broader insight into what enables transferable multispectral bathymetry. Successful cross-regional performance does not arise from architectural complexity alone, but from alignment between physical constraints, spatial structure, and optimization strategy.

First, optical physics defines the feasible solution space. Light attenuation and water-column scattering impose a practical depth ceiling near 20~m for multispectral Sentinel-2 imagery under typical reef conditions. Within this constrained domain, deep networks can approximate stable spectral-depth mappings, but cannot overcome the fundamental limits of bottom signal attenuation. Transferability therefore depends on learning relationships that remain valid within this physically interpretable range.

Second, spatial context is essential. Bathymetric variation follows geomorphic continuity rather than random spatial distribution. Training strategies that preserve contiguous reef structure enable networks to learn how reflectance evolves along slopes, flats, and lagoon transitions. This spatially structured learning reduces ambiguity between dark shallow substrates and deeper water, leading to improved cross-site generalization.

Third, depth-aware optimization refines model behavior within optically sensitive zones. By shaping the error surface through smooth depth-dependent weighting, SWF adjusts the balance between shallow precision and deeper stability without sacrificing global consistency. Loss-function design thus becomes a mechanism for encoding operational priorities into the training process.

Taken together, these findings suggest that scalable satellite-derived bathymetry requires three aligned components: (1) adherence to optical constraints, (2) preservation of geomorphic continuity during training, and (3) depth-aware optimization reflecting application requirements. When these elements are jointly satisfied, multispectral deep learning models can move beyond purely site-specific calibration toward improved reef-to-reef transferability under clear-water reef conditions. However, the persistent accuracy penalty under cross-regional transfer indicates that the current framework achieves partial, not yet operational, transferability. Sun-glint variability \cite{Hedley2005}, sensor geometry, and regional differences in water-column optical properties further contribute to this residual gap, and remain factors that are not fully captured by architecture or training design alone. Addressing these physical sources of domain shift, alongside explicit domain adaptation approaches \cite{Liu2025}, represents an important direction for closing this gap.

\subsubsection*{Reference-Data Uncertainty and Tidal Correction}

A persistent source of uncertainty remains in very shallow waters, where prediction accuracy is influenced not only by model behavior but also by preprocessing assumptions. In particular, tidal correction relies on global tide models such as DTU23, which operate at relatively coarse spatial resolution and use interpolated tidal constituents that may not fully resolve complex shoreline dynamics, wave effects, and local bathymetric controls. Consequently, residual tidal errors near coastlines may propagate into both training labels and model predictions, especially within the upper few meters where small vertical offsets correspond to large relative depth errors. Explicit characterization of tidal and reference-data uncertainty is therefore an important direction for future operational SDB systems, ensuring that model error can be distinguished from observational limitations and that predictions remain truly transferable across regions. Multi-temporal fusion is one way to reduce the impact of (partially) unknown temporal variations.

\section{Conclusions}
\label{sec:conclusions}

This study presents a transferable deep learning framework for shallow-water satellite-derived bathymetry (SDB) using multi-temporal multispectral Sentinel-2 imagery. 
Across multiple reef systems, deep architectures---particularly ResNet and ConvNeXt variants---demonstrate stable performance over the 0--20~m depth range, with highest accuracy in shallow to intermediate waters (approximately 0--12~m), and improved cross-regional robustness compared to classical machine learning approaches. Intra-regional RMSE ranges from 1.15 to 1.92~m, while cross-regional transfer yields moderate RMSE values of approximately 2.46--2.98~m, indicating partial but meaningful transferability across outer-shelf and oceanic reef systems under clear-water conditions. These results show that deep networks can learn spatial--spectral representations that generalize across reef systems, while also highlighting that cross-regional performance degradation remains a key unresolved challenge.

Performance gains arise not from network complexity alone, but from alignment between physical constraints, spatially structured training data, and depth-aware optimization. The Smooth Weight Function (SWF) enhances accuracy at shallow-to-intermediate depths, particularly for higher-capacity architectures such as ConvNeXt-Large, while preserving spatial continuity during training substantially improves transferability. Large contextual patches ($5.12 \times 5.12\ \mathrm{km}^2$) enable learning of coherent geomorphic transitions between reef flats, slopes, and lagoons, reducing depth drift and improving structural consistency.

The integration of multi-temporal imagery further strengthens robustness. During training, repeated observations under varying illumination, atmospheric, and water-column conditions act as implicit temporal augmentation, encouraging the learning of invariant physical relationships between reflectance and depth. During inference, aggregation across multiple dates enables statistical filtering that mitigates transient environmental noise, clouds and shadows, errors in tidal corrections, and other temporal nuisance factors.

In short, this study suggests the following practical guidelines for scalable SDB:
\begin{itemize}
    \item Incorporate depth-aware objectives such as SWF to improve shallow-to-intermediate depth accuracy;
    \item Preserve spatial continuity in training data rather than relying on random sampling;
    \item Employ large-context patches to capture geomorphic transitions; and
    \item Select architectures that maintain low-level spectral sensitivity while enabling hierarchical spatial refinement.
\end{itemize}

When following these guidelines, the general-purpose deep learning models presented in this work achieve competitive cross-regional accuracy, with ConvNeXt-Large reaching its lowest cross-regional RMSE at intermediate depths ($\approx$1~m at $\sim 5$ --10~m at Cartier Reef). 

We note that the elevated cross-regional RMSE observed here is, in part, a consequence of the evaluation setup: training jointly on Pratas Island and the Great Barrier Reef while reserving Ashmore and Cartier reefs as spatially independent hold-outs is a more rigorous transferability test than the within-region random splits often reported in prior SDB literature, and yields correspondingly larger but more honest cross-regional errors.

At the same time, closing the remaining gap toward globally scalable, survey-grade shallow-water bathymetry will require advances that go beyond simply swapping backbones or tuning hyperparameters. In this paper, we intentionally pushed general-purpose, pre-trained computer-vision models about as far as they can reasonably go for SDB. The next step is therefore a dedicated shallow-water bathymetry model explicitly designed around the physics, spatiotemporal nature, and uncertainty structure of the problem. In addition, depth-stratified sampling strategies represent a promising direction to address depth-distribution imbalance between training and transfer regions, and may further enhance cross-regional generalization performance.

\section{Future Work}
\label{sec:futurework}

Future work will focus on developing a more physically grounded shallow-water bathymetry (SWB) framework that extends beyond the lightweight physics-inspired components used in this study by more tightly integrating physical priors, multi-temporal learning, and uncertainty quantification. Key directions include scaling supervision across heterogeneous reference datasets; developing native multi-temporal architectures that directly learn from repeat imagery rather than relying on post hoc aggregation; incorporating explicit treatment of tidal variability and acquisition-dependent offsets; exploring physics-aligned objectives \cite{Sun2023} with calibrated confidence estimates; and extending predictions toward standardized map products with quality indicators.

Building on the present data-driven framework, we are developing in parallel a custom physics-informed neural network (PINN) architecture for shallow-water bathymetry, with dedicated output heads that explicitly model spatial variations in (`effective') substrate albedo and water-column attenuation, and that ingest multi-temporal Sentinel-2 image stacks directly rather than relying on post hoc median fusion. Such an architecture is specifically intended to disentangle the contributions of bathymetry, substrate reflectance, and water-column optics, addressing the limitations of the present study in which these factors covary across training and transfer regions and cannot be independently attributed.

More broadly, enforcing physics constraints from radiative transport is expected to act as a form of regularization: by restricting the network to solutions that respect physically plausible attenuation behaviour, it reduces the network's capacity to overfit residual noise or site-specific artefacts in the training data. As is typical with regularization, this may modestly raise the RMSE measured on the training set, since the network is no longer free to fit every idiosyncrasy of its training data. The intended payoff is a substantially smaller gap between training-set performance and performance on independent hold-out regions: the operationally meaningful measure of transferability is not the training RMSE in isolation, but the \emph{convergence} of training and cross-regional hold-out RMSE. A model whose training and independent hold-out errors are comparable is genuinely generalizable, whereas a model that achieves very low training RMSE only at the cost of substantially larger hold-out errors has simply overfit its training set. Expanding the training set to cover a broader range of reef morphologies, substrate types, depth regimes, and optical conditions has a similar effect, by enlarging the joint distribution of conditions over which the network is forced to remain consistent --- physics-informed regularization and richer training coverage are therefore complementary, rather than alternative, routes toward improved cross-regional transferability.

Complementary directions include explicit incorporation of seafloor-type priors through co-registered benthic habitat maps, semi-analytical correction approaches, and rigorous interpretability analyses such as the regression-adapted class-activation mapping framework recently developed by some of the present authors on a different architecture~\cite{Chowdhury2026}, all of which sit alongside the physics-aware modelling and richer training-coverage directions outlined above.

\vspace{6pt} 

\supplementary{This manuscript is accompanied by one supplementary PDF file, \textit{Supplementary Information}. The Supplementary Information provides: (i) a cross-strategy generalization analysis (Supplementary Section~S1; Supplementary Tables~S1--S2); (ii) a diagnostic experiment on patch size (Supplementary Section~S2; Supplementary Figs.~S1--S2); (iii) a depth-range diagnostic experiment (Supplementary Section~S3; Supplementary Fig.~S3); (iv) additional figures documenting training imagery and the Smooth Weight Function (Supplementary Section~S4; Supplementary Figs.~S4--S6); and (v) a depth-binned decomposition of cross-regional residuals into random-error and systematic-bias components against the IHO TVU at the 95\,\% confidence level (Supplementary Section~S5; Supplementary Figs.~S7--S8).}

\authorcontributions{Conceptualization, H.-J.H. and J.M.; 
methodology, H.-J.H.; 
software, H.-J.H.; 
validation, H.-J.H. and J.M.; 
formal analysis, H.-J.H.; 
investigation, H.-J.H.; 
resources, J.M.; 
data curation, H.-J.H.; 
writing---original draft preparation, H.-J.H.; 
writing---review and editing, H.-J.H. and J.M.; 
visualization, H.-J.H.; 
supervision, J.M.; 
project administration, J.M. 
All authors have read and agreed to the published version of the manuscript.}

\funding{This research received no external funding.}

\dataavailability{All codes and training weights are made freely available to the community on \url{https://github.com/buckai-observatory/DL_bathy} ($^\ast$upon acceptance of paper).}

\acknowledgments{During the preparation of this work the authors used ChatGPT in order to check for spelling and grammatical mistakes and to provide suggestions for the overall organization of sections. After using this tool/service, the authors reviewed and edited the content as needed and take full responsibility for the content of the published article.}

\conflictsofinterest{The authors declare that they have no known competing financial interests or personal relationships that could have appeared to influence the work reported in this paper.}

\begin{adjustwidth}{-\extralength}{0cm}

\PublishersNote{}
\end{adjustwidth}
\end{document}